\pdfoutput=1
\documentclass[11pt]{article}

\usepackage[final]{acl}

\usepackage{times}
\usepackage{latexsym}
\usepackage{tcolorbox}
\usepackage{booktabs}
\usepackage{algorithm}
\usepackage{algorithmic}
\usepackage{microtype}
\usepackage{hyperref}
\usepackage{multirow}
\usepackage{amsmath}
\usepackage{amssymb}
\usepackage{mathtools,caption,subcaption,svg,stfloats}
\usepackage{enumitem}
\usepackage{newfloat}
\usepackage{listings}
\usepackage{color, colortbl}
\usepackage{tabularx}
\usepackage[export]{adjustbox}
\usepackage{wrapfig}
\usepackage{xspace}

\usepackage[T1]{fontenc}
\usepackage[utf8]{inputenc}

\usepackage{microtype}

\usepackage{inconsolata}

\usepackage{graphicx}

\usepackage{hyperref}
\usepackage{url}
\usepackage{booktabs}

\usepackage{lineno}

\usepackage{amsmath}
\usepackage{dsfont}
\usepackage{subcaption}
\usepackage{multirow}
\usepackage{tabularx}
\usepackage{svg}
\usepackage{makecell}

\usepackage{threeparttable}

\usepackage{adjustbox}

\title{\includegraphics[scale=0.15]{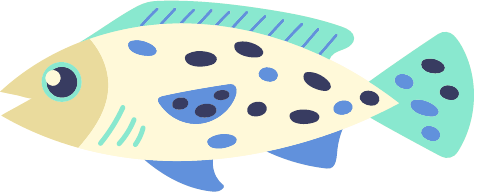} R\textsuperscript{2}-CoD: Understanding Text-Graph Complementarity in \underline{R}elational \underline{R}easoning via Knowledge \underline{Co}-\underline{D}istillation}

\author{
    Zhen Wu \quad
    Ritam Dutt \quad
    Luke M. Breitfeller \quad
    \\
    \textbf{Armineh Nourbakhsh}\thanks{\,Work done while the author was a student at Carnegie Mellon University.} \quad
    \textbf{Siddharth Parekh} \quad
    \textbf{Carolyn Rosé} \\
    Language Technologies Institute, Carnegie Mellon University \\
    \texttt{\{zhenwu, rdutt, mbreitfe, anourbak, spparekh, cprose\}@andrew.cmu.edu}
}

\begin{document}
\maketitle

\begin{abstract}
Relational reasoning lies at the core of many NLP tasks, drawing on complementary signals from text and graphs. While prior research has investigated how to leverage this dual complementarity, a detailed and systematic understanding of text-graph interplay and its effect on hybrid models remains underexplored. We take an analysis-driven approach to investigate text–graph representation complementarity via a unified architecture that supports knowledge co-distillation (CoD). We explore five tasks involving relational reasoning that differ in how text and graph structures encode the information needed to solve that task. By tracking how these dual representations evolve during training, we uncover interpretable patterns of alignment and divergence, and provide insights into when and why their integration is beneficial.
% \footnote{Our code is available at \url{https://github.com/zhenwu0831/R2_COD}}
\end{abstract}

\section{Introduction}
Incorporating modalities beyond the surface form of the text has shown promise for several challenging natural language processing (NLP) tasks. This is particularly true for \textbf{relational reasoning} based tasks where the objective is to understand or infer the semantic relationships within the input \citep{nastase15graphsurvey}. Examples of such tasks are relation extraction~\citep{zhang2018graphconvolutionpruneddependency, christopoulou2019connectingdotsdocumentlevelneural, guo2020attentionguidedgraphconvolutional}, knowledge base question answering (KBQA)~\citep{tian-etal-2024-augmenting,feng-he-2025-rgr,gao2025seendataimprovingkbqa}, and structured document interpretation or reasoning ~\citep{yao2018graphconvolutionalnetworkstext,wang2023graphicalapproachdocumentlayout, chen2025graphbaseddocumentstructureanalysis}.

A common and effective way to encode relational structure is through graphs~\citep{yao2018graphconvolutionalnetworkstext, lee-etal-2023-formnetv2, lin2025gt2veclargelanguagemodels, gururaja2023linguistic, dutt2022perkgqa}, where nodes represent textual units and edges encode relationships, like semantic links or ontological structure. This explicit representation of structured information enables models to leverage signals that are complementary to or explicitly absent from the text.

While many tasks utilize this text-graph representation to improve performance, how they complement each other remains underexplored. Some systematic reviews~\citep{stanton2021fidelity} observe that models fail to effectively integrate data from distinct modalities.
% may not synergize well when integrated.
This raises important open questions: How do text and graph representations relate to each other during learning? Do they converge toward similar representations, or diverge to encode distinct signals? And under what conditions is their integration most beneficial? 

To address these questions, we adopt an analysis-oriented approach and introduce a unified framework for characterizing the alignment and complementarity between text and graph representations across tasks. We inspect how these dual representations relate and evolve with knowledge co-distillation (CoD)~\citep{yao2024distilling}, an architectural framework that can generalize across a range of tasks where both text and graph inputs are available. We conduct this analysis across a diverse suite of five tasks involving relational reasoning spanning fine-grained, localized reasoning between entity pairs to multi-entity inference. To this end: 

\begin{itemize}[itemsep=-0.1em,leftmargin=1.05em,topsep=0em]
    \item We systematically analyze how text and graph representations complement each other under knowledge co-distillation (CoD) across five relational reasoning tasks.
    \item We identify consistent patterns ranging from complementarity to alignment and characterize how these patterns differ across tasks. 
    \item We provide practical insights to inform the effective use of CoD.
\end{itemize}

\section{Related work}
\paragraph{Text–graph integration in NLP:}
Graphs have long played an important role in NLP, traditionally used to capture structure in tasks such as syntactic parsing, information retrieval, text mining, and encode semantic representation through knowledge graphs, linguistic frameworks, and other semantic networks. Graph Neural Networks (GNNs)~\citep{4700287} and their variants such as Graph Convolutional Neural Networks (GCNs) ~\citep{bruna2014spectralnetworkslocallyconnected} and Graph Attention (GAT) layers ~\citep{veličković2018graphattentionnetworks} have become the de-facto way to integrate text and graph representations across a variety of tasks. 

In text classification, graphs have been used to jointly model word and document relations~\citep{yao2018graphconvolutionalnetworkstext} and to enhance transformers with structured information~\citep{lin-etal-2021-bertgcn}. Knowledge graphs provide support for reasoning and information retrieval for QA ~\citep{sun2018opendomainquestionanswering, yasunaga2022qagnnreasoninglanguagemodels, lin2025gt2veclargelanguagemodels}. For document understanding, graph-based methods have been applied to paragraph recognition~\citep{9706840, liu2022unifiedlineparagraphdetection}, information extraction~\citep{lee-etal-2023-formnetv2}, and layout or structure analysis~\citep{wang2023graphicalapproachdocumentlayout, chen2025graphbaseddocumentstructureanalysis}. More recently, such approaches have also been used to detect AI-generated content~\citep{valdez-gomez-adorno-2025-text}.

% Integrating text and graph representations enables models to capture long-range dependencies, incorporate structured external knowledge, and support multi-hop reasoning over entities and relations that are not explicitly connected in the text alone.

While text–graph integration has been widely used for performance gains, little is known about how their representations relate during learning. We analyze this relationship and how it is shaped by task characteristics and learning objectives.

\paragraph{Knowledge distillation (KD):}  One of the earliest works in this space was of \citet{bucilua06compression}, i.e. a kind of model compression to facilitate efficient ensembling of complex classifiers. \citet{hinton2015distilling} refined it to distill knowledge from one model to another. Later, this type of directed, teacher-student knowledge distillation (KD) has seen usage in several NLP tasks~\citep{sanh2019distilbert, sun2019patient, liang2020mixkd, liu2022multi}.
% Models such as DistilBERT \citep{sanh2019distilbert}, BERT-PKD \citep{sun2019patient}, MixKD \citep{liang2020mixkd}, MGSKD \citep{liu2022multi} apply this type of directed, teacher-student knowledge distillation (KD) to the natural language space. 
As opposed to distilling information from one model to another, \citet{zhang2018deep} proposed the idea of mutual learning where information is shared between models. Finally, \citet{tian2019crd} introduced contrastive representational distillation, which later works~\citep{sun2020contrastive, fu2021lrc} showed is effective at refining KD-loss for shared representational spaces.

% \rd{Not super happy about this part of the related work section tbh}.

Though KD is widely prevalent in NLP, its effectiveness in successfully compressing complex tasks remains unclear. \citet{stanton2021fidelity} argues that a gap exists in our current understanding of KD, evident in the difficulty in obtaining model fidelity for certain types of teachers. Though it is known that KD's efficacy varies across models, the reason remains unknown.

\paragraph{Representation analysis:}

Representation analysis examines the internal representations learned by models to better understand how they encode and process information. Subsequently, a variety of tools have been developed for this purpose. These range from traditional methods such as Principal Component Analysis (PCA) \citep{10.3389/frobt.2019.00153} and Canonical Correlation Analysis (CCA) for dimensionality reduction and visualization,
%PCA, for instance, helps uncover dominant axes of variation in embedding spaces and creates human-interpretable distributed representations~\citep{10.3389/frobt.2019.00153}. 
to more targeted approaches such as classifier probes to test whether specific linguistic properties are encoded in model representations~\citep{belinkov2021probingclassifierspromisesshortcomings, gupta-etal-2015-distributional}. Recently, sparse autoencoders \citep{gao2024scalingevaluatingsparseautoencoders, cunningham2023sparseautoencodershighlyinterpretable, ng2011sparse} have also been deployed for extracting interpretable features from model representations.

To support our goal of analyzing how text and graph representations are related during learning, we require lightweight, task-agnostic tools to enable consistent and interpretable comparisons across tasks. We thus adopt PCA and leverage distance-based metrics to answer our questions.

\section{Task suite and formulations} \label{sec:task_suite}
% We evaluate text–graph interaction across five relational reasoning tasks. They span from fine-grained, localized reasoning between entity pairs to multi-entity inference. 

\begin{figure}[t]
    \centering
    \includegraphics[width=\linewidth]{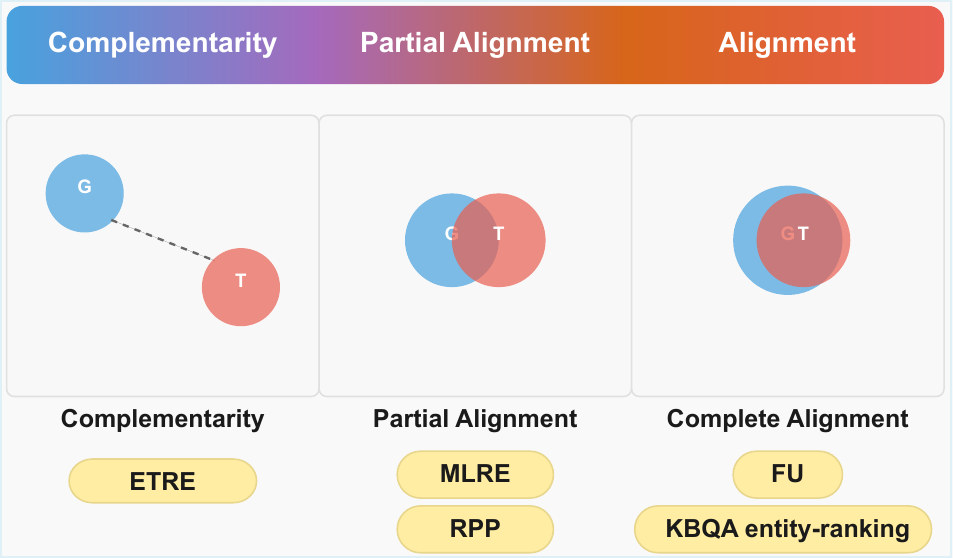}
    % \vspace{\baselineskip}
    \caption{Task spectrum of representation relationships. Left: they remain distinct and complementary. Middle: they show some similarity but do not fully align. Right: they converge toward aligned representations. This spectrum motivates our task selection for analysis.}
    \label{fig:spectrum}
\end{figure}

We propose a spectrum of how the relationship between the text and the graph representations can vary as visualized in \textbf{Figure~\ref{fig:spectrum}}. This spectrum ranges from cases where text and graph encode largely complementary information and preserve distinct representations  (left), to cases where they tend to converge and form aligned representations (right). In between, partial alignment refers to the case where the representations become more similar but do not fully converge. To cover this spectrum, we select five relational reasoning tasks with diverse characteristics, such as 1) how explicitly the graph models the relation or structure that the task seeks to predict, 2) whether nodes have direct correspondence to textual spans, and 3) the scope of reasoning (e.g., local mention pairs versus global graph structure). This diversity enables us to examine how these variations shape text-graph representation relations. %We provide a summary table to outline the nature and diversity of the tasks in Table~\ref{tab:task_details}.
We outline the goal, input and output, an illustrated example, graph construction method, and the knowledge type of each task in Table~\ref{tab:task_details}.

\begin{table*}
\centering
\resizebox{\linewidth}{!}{
\begin{tabular}{p{1.2cm} p{2.5cm} p{2cm} p{2.3cm} p{8.0cm} p{1.3cm}}
\toprule
\textbf{Task} & \textbf{Goal} & \textbf{Input} & \textbf{Output} & \textbf{Example} &  \textbf{K-Type} \\
\midrule
ETRE & Predict temporal relation between two events & Text passage + Syntactic graph and Time-aware graph & Relation label (e.g., BEFORE/AFTER) & In: Atlanta nineteen ninety-six. A bomb <E1> blast </E1> shocks the Olympic games. One person is killed. January nineteen ninety-seven. Atlanta again. This time a bomb at an abortion clinic. More people are <E2> hurt </E2>.
Out: Event E1 took place BEFORE Event E2. & Episodic \\
\midrule
MLRE & Predict semantic relation between entities & Text passage + Dependency graph & Relation label (e.g. sibling) & In: The <E1> wood </E1> is used as fuel and to make posts for <E2> fences </E2>. Out: The relation between E1 and E2: material used &  Episodic \\
\midrule
FU & Predict token relationships in scanned forms & OCR tokens with layout info & Label over token pairs  & We present an example in Figure~\ref{fig:fu_example} & Episodic \\
\midrule
RPP & Predict reasoning path over the KG for a question. & Question + KG subgraph & Reasoning Path & 
\multirow{2}{*}{%
\parbox[t]{8.0cm}{\raggedright
In: Question: What was Elie Wiesel's father's name?\\
KG: \texttt{ Elie Wiesel <E1> | <E1> book.author.book\_editions\_published <E2> | <E3> people.person.gender <E4> ...}\\
Out: Reasoning Pattern Type: T2 — The answer is located a single-hop away from the two constraints.\\
Entities ranked: \texttt{ <E6>, <E4>, ...}
}}
& Static \\
KBQA entity-ranking
& Extract answers from a KG for a question
&
& Ranked list of candidate entities
& & \\
% \multirow{2}{*}
% \parbox[t]{5.8cm}{\raggedright
% {In: Question: What was Elie Wiesel's father's name?\\
% \texttt{KG: Elie Wiesel <E1> | <E1> book.author.book\_editions\_published <E2> | <E3> people.person.gender <E4> ...RPP}\\
% Output: Reasoning Pattern Type: T2 — The answer is located a single-hop away from the two constraints. Entities ranked: <E6>, <E4>, ...}} &
% \multirow{2}{*}{Extracted from external KB (Freebase) for each question} & \multirow{2}{*}{Static} \\
% KBQA entity-ranking & Reason over KG to extract answers for given question &  & Ranked list of candidate entities &  &  &  \\
\bottomrule

\end{tabular}}
\caption{For each task, we state the goal, the input/output format, an illustrative example, and the graph construction method. We also distinguish between tasks grounded in \textit{episodic} knowledge (context-dependent and document-specific), and those involving \textit{static} knowledge (holds independently of context) in the Knowledge(K)-Type column.}
\label{tab:task_details}

\end{table*}

\paragraph{Event temporal relation extraction (ETRE)}

The objective of ETRE is to predict the temporal relationship $y_{ij}$ between a pair of event mentions $(i,j)$ within a short passage $q$ using a fixed relation label set (e.g., \textit{before}, \textit{after}, \textit{simultaneous}, \textit{vague}). Because distinct layers of text encode time cues for short- and long-distance mention pairs, the model represents the text using linear transformers alongside associated graph $G(V,E)$. The graph contains nodes $V$ for event mentions and time expressions, and edges $E$ encoding structural relations. It is derived from the $q$ using part-of-speech labeling and by applying temporal logic to limited date-time associations which can be extracted from $q$. Thus, while the graph does not explicitly encode the temporal relation being predicted or have direct correspondence with text spans, it reflects long-distance structural dependencies not captured by linear transforms. We follow~\citet{yao2024distilling} and use three benchmark datasets: TimeBank-Dense (TB-Dense)~\citep{tbdense}, TDDiscourse-Auto (TDDAuto) and TDDiscourse-Manual (TDDMan)~\citep{tdd}.
% , which between them capture both short- and long-range event interactions from short news articles.

\paragraph{Multilingual relation extraction (MLRE)}
In a similar vein, the task of MLRE involves identifying the semantic relation between a pair of entity mentions within a given sentence(s) $q$ for a particular language. Each text input has its corresponding graph $G(V,E)$ generated by an off-the-shelf dependency parser \citep{qi2020stanza} where $V$ represents the words in the sentence(s) and the $E$ represents the syntactic dependencies between the words. We initialize the nodes(words) in the graph by pooling across its constituent token embeddings, and further augment it with the structural information obtained from the graph's topology via Walklets \citep{walklets}. We emphasize that the dependency relations capture the explicit linguistic signals between words but do not encode the relation being predicted. We provide an example in the Appendix~\ref{appendix:dependency}. We use the RED\textsuperscript{fm}~\citep{huguet-cabot-etal-2023-red} dataset which covers five languages.

\paragraph{Reasoning pattern prediction (RPP)}
% The isomorphism prediction task focuses on structural fidelity of reasoning in knowledge base question answering (KBQA). 
Given a question $q$ and its associated subgraph $G=(V,E)$ from the knowledge base, the goal is to infer the reasoning pattern or RP of the question $q$. 
Each pattern corresponds to a particular reasoning path, composed of single/multiple hops and single/multiple constraints. We provide detailed descriptions in Appendix~\ref{data_processing}. The text input includes $q$ and a linearized serialization of the subgraph $G^{\text{linear}}$, while the graph input uses the same question paired with the explicit graph structure $G$. Thus, both text and graph encode the same information but in structurally distinct forms. We initialize the nodes in each subgraph with Walklets embeddings following the same procedure as \citet{dutt2022perkgqa}. RP prediction requires reasoning over the entire graph with respect to the question, rather than individual tokens or nodes. We use WebQSP dataset for our task~\citep{WebQSP,xie-etal-2022-unifiedskg}.

\paragraph{KBQA entity-ranking}
We formulate extracting the correct answer(s) for a given question from its associated subgraph as a ranking problem.  The model operates over a shared set of candidate entities and assigns a relevance score to each entity based on its likelihood of being the correct answer. The input is the same as in the reasoning pattern prediction task setting. 
% we WebQSP~\citep{WebQSP,xie-etal-2022-unifiedskg} as our representative dataset, where each question $q$ is paired with a subgraph $G(V,E)$ for the graph side, and its linearized form $G^{\text{linear}}(V,E)$ for the text side. 
To enable entity-level predictions from the text model, we extract an embedding for each candidate entity by identifying its corresponding span in the text and aggregating the token representations produced by the text encoder. Each entity thus has a one-to-one correspondence: it appears as a node $v_i \in V$ in the graph and as a token span $s_i \subseteq q$ in the text.

\paragraph{Form understanding (FU):} This task involves identifying key–value relationships between textual spans extracted from scanned forms, such as ``Date: 2024-12-01''. Each input document is processed by OCR to yield textual tokens with bounding-box coordinates. The corresponding graph $G(V,E)$ encodes the visual layout of the document:  $V$ represents OCR tokens and $E$ captures spatial relations between the tokens such as alignment, proximity, and reading order. Such a framework encodes positional cues central to the task objective, and establishes a one-to-one correspondence between the nodes and the OCR tokens. We adopt the experimental setup of~\citet{nourbakhsh-etal-2024-aligatr} and include three multimodal datasets, i.e. SROIE~\citep{sroie}, FUNSD~\citep{jaume2019funsd}, and CORD~\citep{park2019cord}.

\section{Unified framework for analysis} \label{section:framework}
\begin{figure*}[h]
    \centering
    \includegraphics[width=0.9\linewidth]{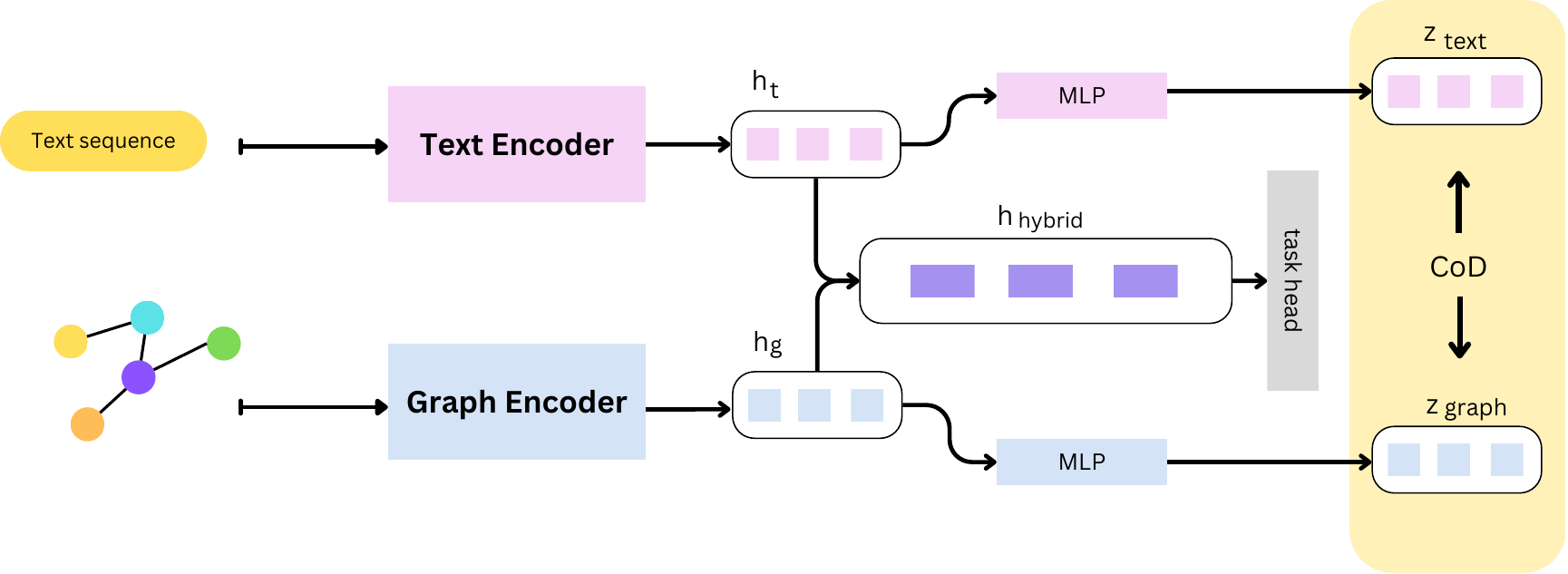}
    % \vspace{\baselineskip}
    \caption{Our unified framework for analyzing how text and graph representations complement each other. A text sequence and its corresponding graph are processed by separate encoders. Their outputs are used in two ways: (1) combined as hybrid inputs for task prediction, and (2) projected into a shared space where a contrastive co-distillation (CoD) objective encourages mutual learning and enables representation-level analysis.}
    \label{fig:framework}
\end{figure*}

We propose a unified, task-agnostic framework, henceforth called 
R\textsuperscript{2}-CoD (\textbf{Figure \ref{fig:framework}}), to understand how text and graph representations relate during learning. We choose a framework that is generalizable to observe how information from text and graph are represented and integrated.

Across tasks, each instance corresponds to a text-graph pair, as defined in Section~\ref{sec:task_suite}. These are encoded using modality-specific encoders: $h_{\text{t}}=f_{\text{t}}(q)$ and $h_{\text{g}}=f_{\text{g}}(G)$. We then create a hybrid representation $h_{\text{hybrid}}$ through concatenation or residual connection to perform task-specific prediction and compute the task loss:
\begin{align}
    h_{\text{hybrid}} &= f_{\text{fuse}}(h_\text{t}, h_\text{g}) \\ 
    \mathcal{L}_{\text{task}} &= \mathcal{L}(h_\text{hybrid}, y)
\end{align}
where $y$ denotes the gold supervision and $\mathcal{L}(\cdot,\cdot)$ is the task-specific loss function. We present model configurations, loss function, and evaluation metrics used for each task in Table~\ref{tab:task_exp_details} in the Appendix. 

To analyze text and graph representations, we require a shared space where they can be directly compared. Thus, we apply modality-specific MLP projection heads that learn to map each representation into a shared latent space during training:

\begin{align}
z_{\text{text}}=\text{MLP}_{\text{t}}(h_{\text{t}}), z_{\text{graph}}=\text{MLP}_{\text{g}}(h_{\text{g}}).    
\end{align}

\subsection{Contrastive co-distillation}
While learning a shared space enables comparison, it cannot solely influence how text and graph will complement one another. We thus apply a contrastive knowledge co-distillation (CoD) objective~\citep{yao2024distilling} which combines a contrastive loss with a stop-gradient operation~\citep{Chen_2021_CVPR} to explicitly encourage bidirectional knowledge transfer. Such a formulation allows us to observe how the information encoded in one modality influences the other during mutual learning. 
% We hypothesize that CoD could leverage and accentuate the inherent complementarity between the two modalities.

% \begin{table*}[t]
% \centering
% \small
% \begin{tabularx}{0.8\linewidth}{l l *{4}{>{\centering\arraybackslash}X}}
% \toprule
% \textbf{Task} & \textbf{Dataset} & \textbf{Text only} & \textbf{Graph only} & \textbf{Hybrid + CoD} & \textbf{Hybrid + no CoD} \\
% \midrule
% \multirow{1}{*}{ETRE} 
%   % & TB-Dense & 61.9 & -- & \textbf{85.6} & 76.4 \\
%   & TDDAuto & 61.6 & 34.6 & \textbf{77.1} & \underline{68.9} \\
%   % & TDDMan & 37.1 & -- & \textbf{55.1} & 44.5 \\
% \midrule
% \multirow{1}{*}{Form understanding\protect\footnotemark} 
%   & FUNSD & 33 & 22 & \textbf{38} & \underline{35} \\
% \midrule
% \multirow{1}{*}{MLRE}
%   & RED\textsuperscript{fm} & \underline{79.1} & 76.3 & \underline{79.1} & \textbf{80.0} \\
% \midrule
% Reasoning pattern prediction & WebQSP & 62.4 & 63.2 & \textbf{65.9} & \underline{65.6} \\
% \midrule
% KBQA entity-ranking & WebQSP & 80.7 & 52.2 & \textbf{83.8} & \underline{83.5} \\
% \bottomrule
% \end{tabularx}
% \caption{Task performance (averaged across multiple seeds) for text-only, graph-only, hybrid with CoD, and hybrid without CoD.\protect\footnotemark Best performance in \textbf{bold}, second-best \underline{underlined}.}
% \label{tab:task_performance}
% \end{table*}

\begin{figure*}
  \centering
    \resizebox{0.9\textwidth}{!}{  % Resize to 90% of the page width
    \begin{minipage}{\textwidth}
      \centering
  % Row 1: PCA plots - 3 across full width
  \begin{subfigure}[t]{0.32\textwidth}
    \centering
    \includegraphics[trim=38 25 70 70, clip, width=\linewidth]{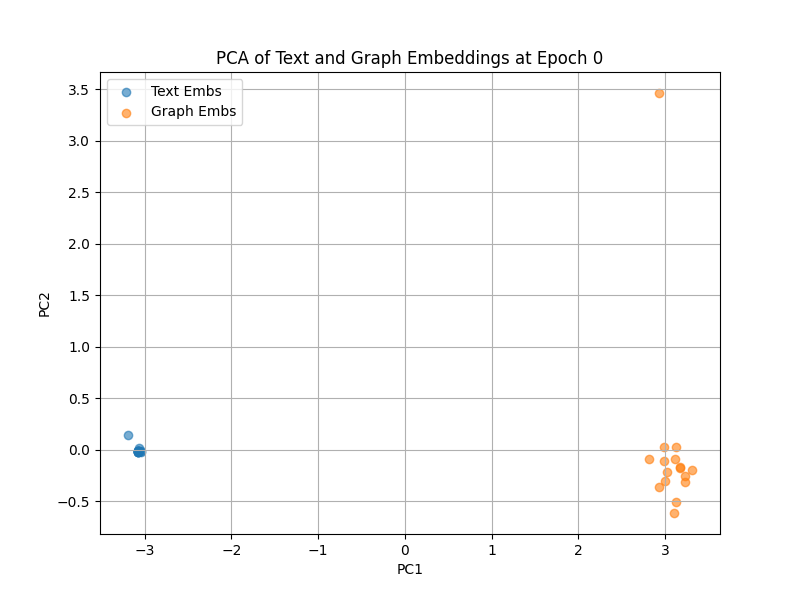}
    \caption{Initial epoch}
  \end{subfigure}
  \hfill
  \begin{subfigure}[t]{0.32\textwidth}
    \centering
    \includegraphics[trim=38 25 70 70, clip, width=\linewidth]{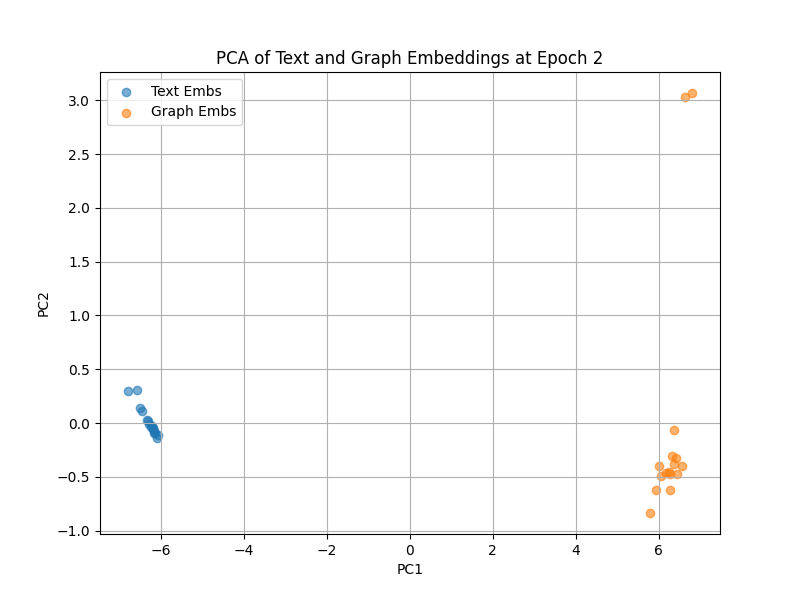}
    \caption{Intermediate epoch}
  \end{subfigure}
  \hfill
  \begin{subfigure}[t]{0.32\textwidth}
    \centering
    \includegraphics[trim=38 25 70 70, clip, width=\linewidth]{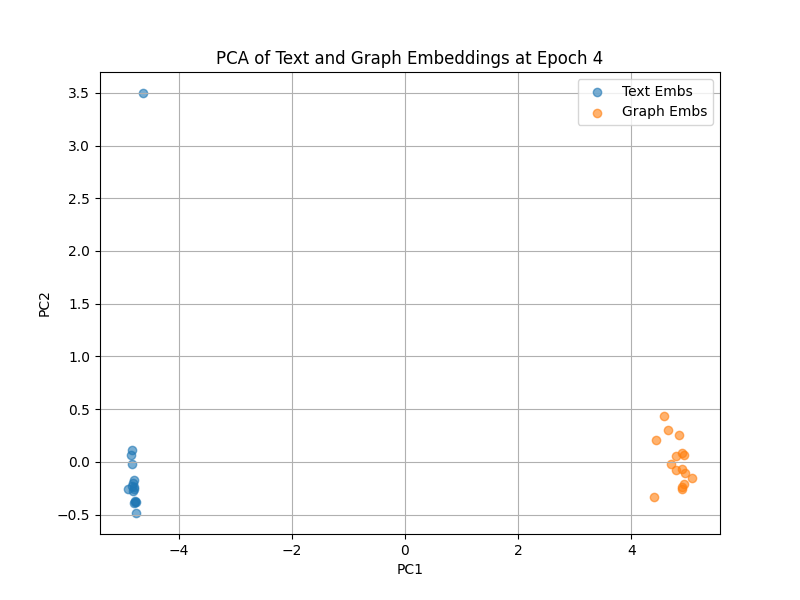}
    \caption{Final epoch}
  \end{subfigure}

  % \vspace{0.2em}

  % Row 2: Distance-related plots in 2x2 grid, trimmed more aggressively
  \begin{subfigure}[t]{0.49\textwidth}
    \centering
    \includegraphics[trim=0 10 0 160, clip, width=0.8\linewidth]{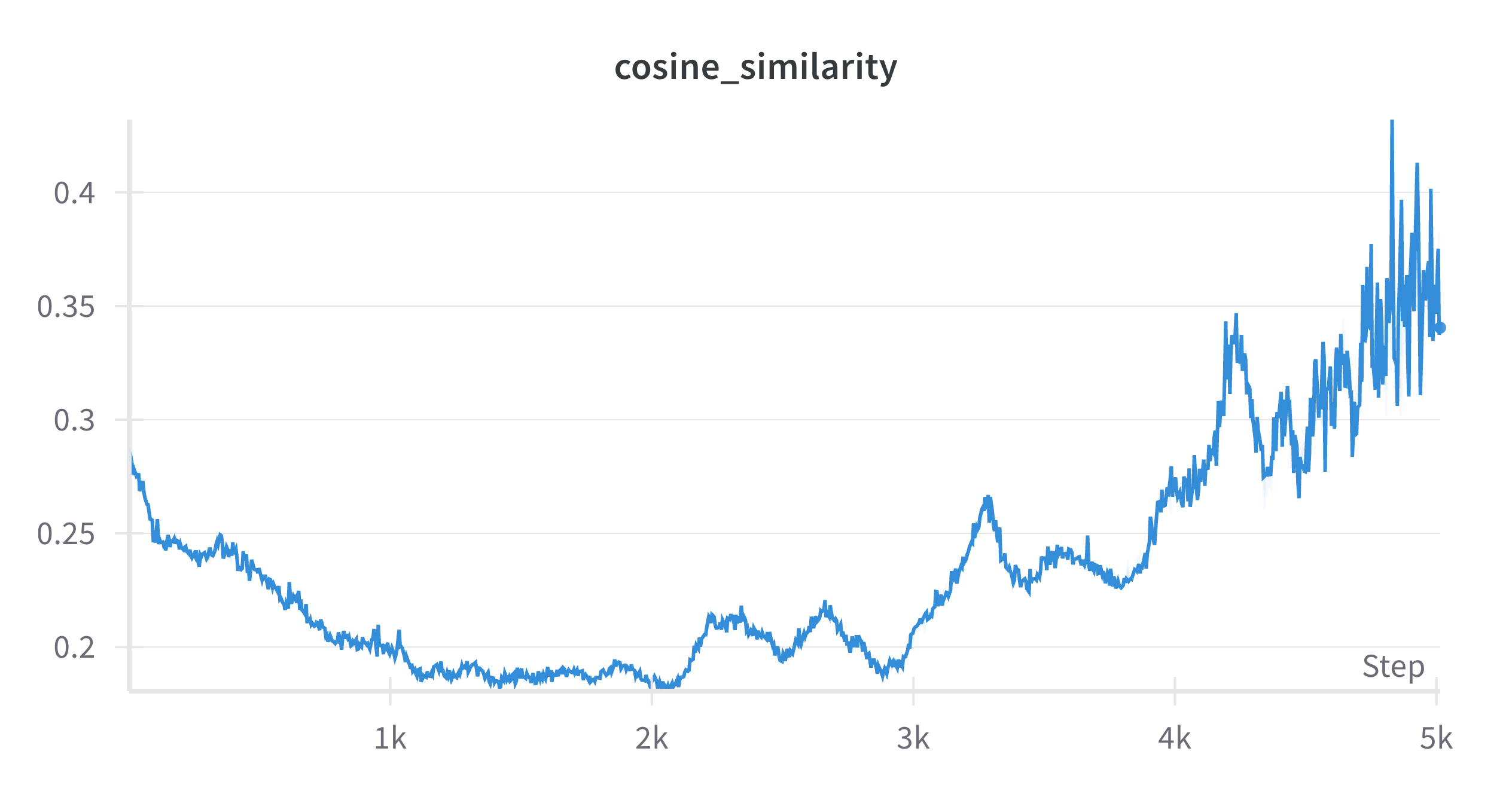}
    \caption{Cosine similarity}
  \end{subfigure}
  \hfill
  \begin{subfigure}[t]{0.49\textwidth}
    \centering
    \includegraphics[trim=0 10 0 160, clip, width=0.8\linewidth]{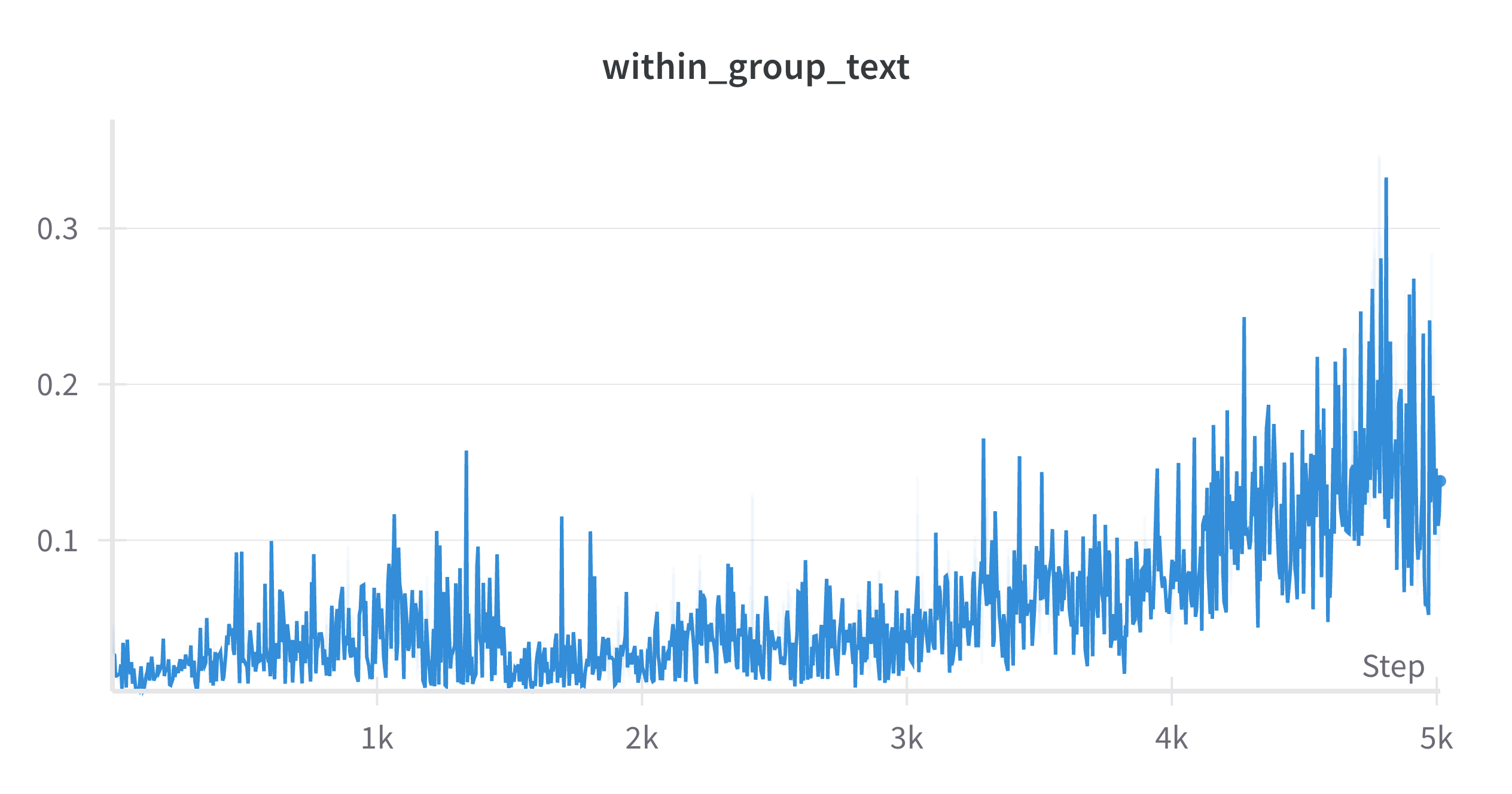}
    \caption{Distance within text}
  \end{subfigure}

  % \vspace{0.2em}

  \begin{subfigure}[t]{0.49\textwidth}
    \centering
    \includegraphics[trim=0 10 0 160, clip, width=0.8\linewidth]{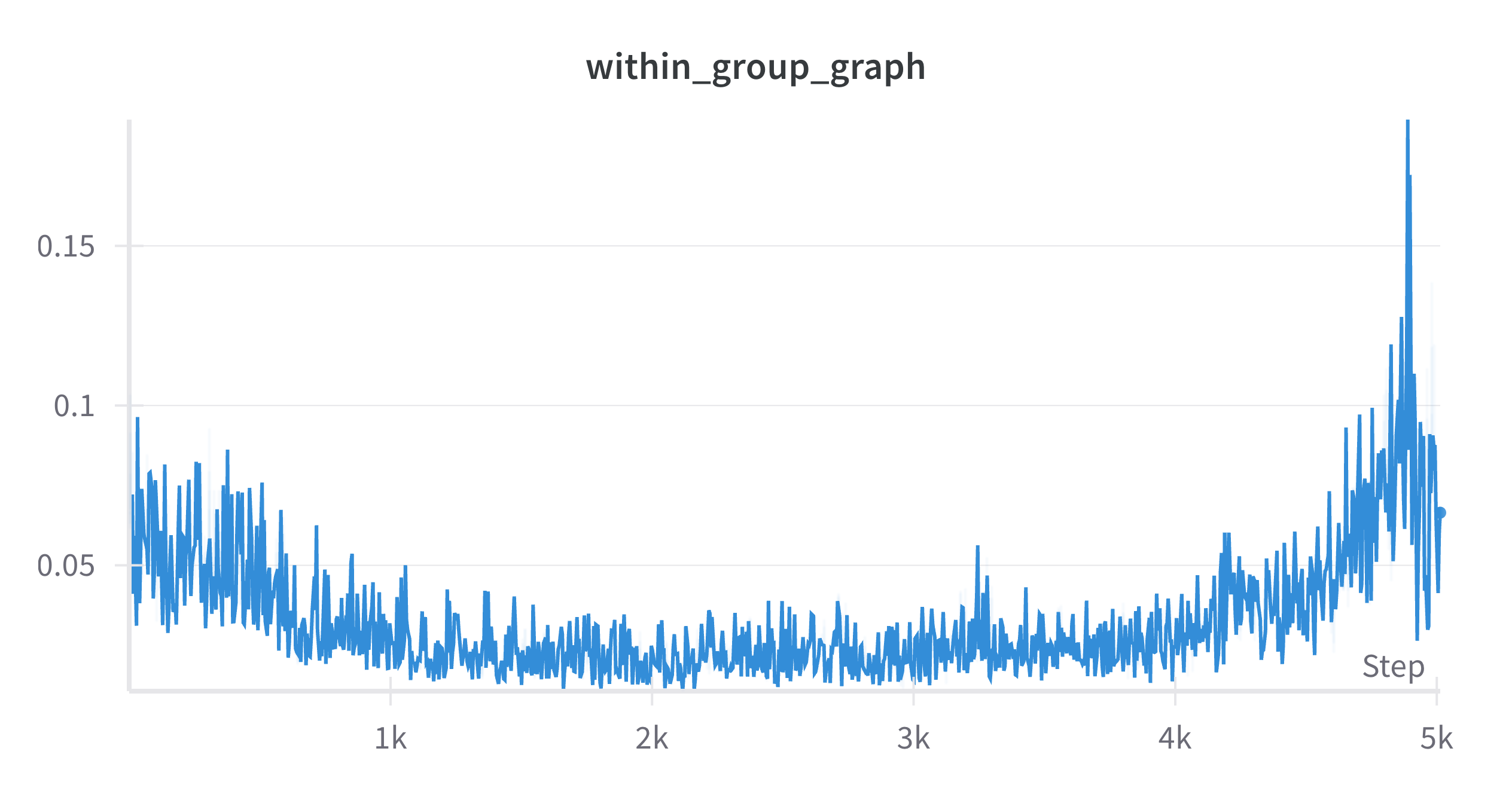}
    \caption{Distance within graph}
  \end{subfigure}
  \hfill
  \begin{subfigure}[t]{0.49\textwidth}
    \centering
    \includegraphics[trim=0 10 0 160, clip, width=0.8\linewidth]{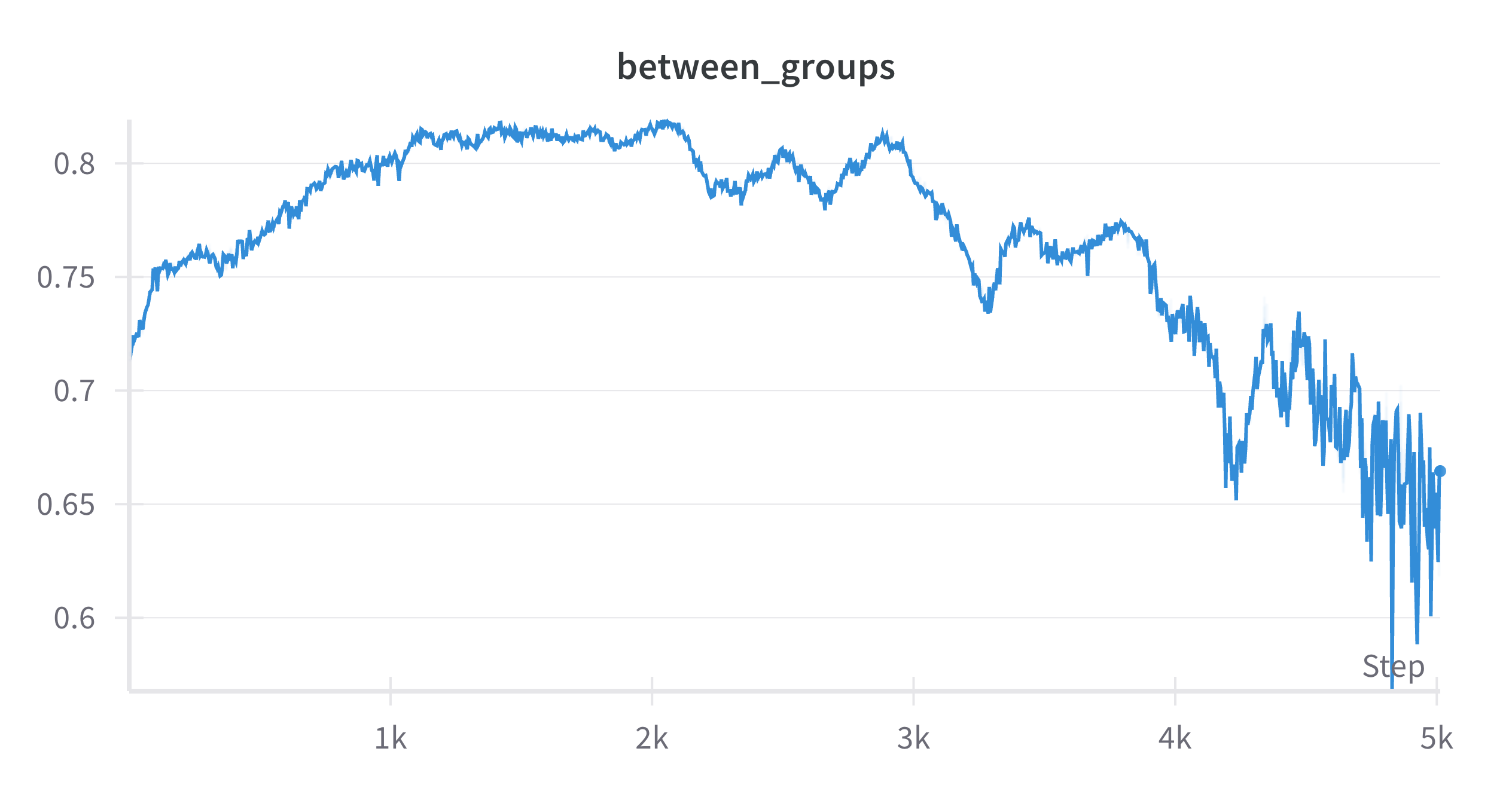}
    \caption{Distance between text and graph}
  \end{subfigure}

  \end{minipage}
  }

  \caption{
    Results for ETRE on the TDDMan dataset. PCA visualizations (top) across training stages, and corresponding distance-based metrics (bottom). The text and graph representations remain well-separated, and the between-group distance remains consistently higher than the within-group distances. 
  }
  \label{fig:etre_tddman_results}
\end{figure*}

Formally, the contrastive loss $l_{cl}$ between the teacher $t$ and the student $s$ representations is:
\begin{equation} \label{cl_loss}
    l_{\text{cl}}(t, s) = -\log\dfrac{e^{sim(t, s)/\tau}}{\sum_{u}\mathds{1}_{[u \neq t]}\,e^{sim(t, u)/\tau}}
\end{equation}
where $u$ indicates representations from the training data other than $t$ and $s$, $sim(.,.)$ is cosine similarity, $\tau$ is the temperature scaling parameter~\citep{tian2022contrastiverepresentationdistillation}. Note that the notions of ``teacher'' and ``student'' are interchangeable and fully symmetric: one scenario treats the text projection behaves as the teacher supervising the graph projection, while in another the graph projection supervises the text projection. This bidirectional design ensures that either modality can act as teacher or student at each step, thus mutually distilling knowledge from each other. Hence, the full CoD loss is computed as 
\begin{equation} \label{cod_loss}
    \mathcal{L}_{\text{CoD}} = \dfrac{1}{2}\sum_{i}[l_{\text{cl}}(z_i^{\text{text}}, \hat{z_i}^{\text{graph}}) + l_{\text{cl}}({z_i}^{\text{graph}}, \hat{z_i}^{\text{text}})]
\end{equation}
where $\hat{.}$ is the stop gradient operator~\citep{Chen_2021_CVPR} that sets the input variable to a constant. Finally, we combine this with the task loss to enable end-to-end model optimization:
\begin{equation} \label{total_loss}
\mathcal{L}_{\text{total}} = \mathcal{L}_{\text{task}} + \lambda\mathcal{L}_{\text{CoD}}
\end{equation}
where $\lambda$ controls the weight of the CoD signal. CoD serves as a task-agnostic framework to facilitates learning and analysis over dual modalities.

\subsection{Measuring representation relations}
To evaluate how text and graph relate during learning, we need tools that can surface both alignment and divergence in the shared space. Our goal is to characterize the degree to which text and graph converge, remain distinct, or shift in their relationship throughout training under CoD.

To support visual interpretation, we apply PCA, which reduces the projected embeddings $(z_{\text{text}}, z_{\text{graph}})$ into a two dimensional space and reveals their spatial arrangement at various stages of training, i.e. whether there is clustering, separation, or overlap between modalities.

For a more precise measurement, we also compute batch-level cosine similarity between paired representations, along with average within- and between-modality distances based on cosine distance. Formally, given two representation vectors $\mathbf{x}$ and $\mathbf{y}$, the cosine similarity is defined as
\[
\operatorname{cos\_sim}(\mathbf{x}, \mathbf{y}) = \frac{\mathbf{x}^\top \mathbf{y}}{\|\mathbf{x}\|\|\mathbf{y}\|},
\]
and the corresponding cosine distance is given by
\[
\operatorname{cos\_dist}(\mathbf{x}, \mathbf{y}) = 1 - \operatorname{cos\_sim}(\mathbf{x}, \mathbf{y}).
\]
Together, these measures capture both the directional and spatial properties of different modalities in the learned representation space.

\section{Analysis and discussions}
\label{section:discussion}

\subsection{RQ1: Does combining text and graph representations improve performance?}
We examine whether integrating textual and graph-based representations improves task performance, and whether CoD facilitates more effective integration. We compare four model configurations: (1) text-only, (2) graph-only, (3) hybrid without CoD, and (4) hybrid with CoD. We present the main results in Table \ref{tab:task_performance}. The task suite statistics and training times in Table~\ref{tab:task_stats} illustrate that CoD introduces minimal additional cost. Additional results for different model combinations in Table~\ref{tab:additional_backbones} further demonstrate the generalizability of our CoD framework across different text and graph models.\footnote{For consistency, we report results either from our own experiments or from existing work when the same architecture is adopted.}

\begin{table}[t]
\centering
\small
\begin{tabular}{lccccc}    
% \begin{tabularx}{0.8\linewidth}{l l *{4}{>{\centering\arraybackslash}X}}
\toprule
\textbf{Task} & \textbf{Dataset} & \textbf{Text} & \textbf{Graph} & \textbf{CoD} & \textbf{T+G} \\
\midrule
\multirow{1}{*}{ETRE} 
  % & TB-Dense & 61.9 & -- & \textbf{85.6} & 76.4 \\
  & TDDAuto & 61.6 & 34.6 & \textbf{77.1} & \underline{68.9} \\
  % & TDDMan & 37.1 & -- & \textbf{55.1} & 44.5 \\
\midrule
\multirow{1}{*}{FU} 
  & FUNSD & 33 & 22 & \textbf{38} & \underline{35} \\
\midrule
\multirow{1}{*}{MLRE}
  & RED\textsuperscript{fm} & \textbf{79.7} & 48.6 & 78.6 & \underline{79.5} \\
\midrule
RPP & WebQSP & 62.4 & 63.2 & \textbf{65.9} & \underline{65.6} \\
\midrule
KBQA & WebQSP & 80.7 & 52.2 & \textbf{83.8} & \underline{83.5} \\
\bottomrule
% \end{tabularx}
\end{tabular}
\caption{Task performance (averaged across three seeds) for text-only (Text), graph-only (Graph), hybrid with CoD (CoD), and hybrid without CoD, i.e. with only the text and graph representations (T+G). Best performance in \textbf{bold}, second-best \underline{underlined}. We present results for one representative dataset per task due to resource constraints. Similar trends hold for other datasets. For FU, the model was pretrained on a 1,000-example subset of its original pretraining corpus.}
% \textsuperscript{1}Model was pretrained on a 1,000-example subset of its original pretraining corpus.
% \textsuperscript{2}We present results for one representative dataset per task due to resource constraints. Similar trends hold for other datasets.
\label{tab:task_performance}
\end{table}

\begin{figure*}[h]
  \centering
    \resizebox{\textwidth}{!}{  % Resize to 90% of the page width
    \begin{minipage}{\textwidth}
      \centering
  
  % Row 1: Three PCA plots
  \begin{subfigure}[t]{0.32\textwidth}
    \centering
    \includegraphics[trim=38 25 70 70, clip, width=\linewidth]{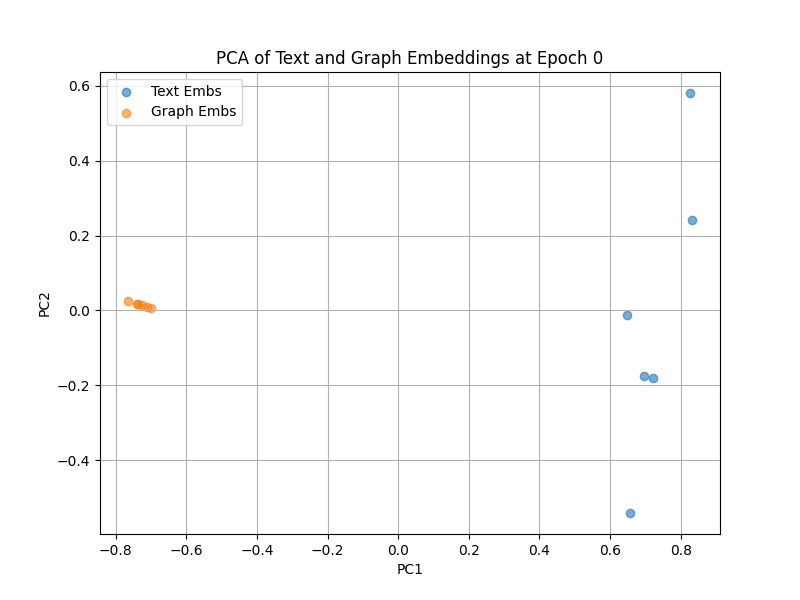}
    \caption{Initial epoch}
  \end{subfigure}
  \hfill
  \begin{subfigure}[t]{0.32\textwidth}
    \centering
    \includegraphics[trim=38 25 70 70, clip, width=\linewidth]{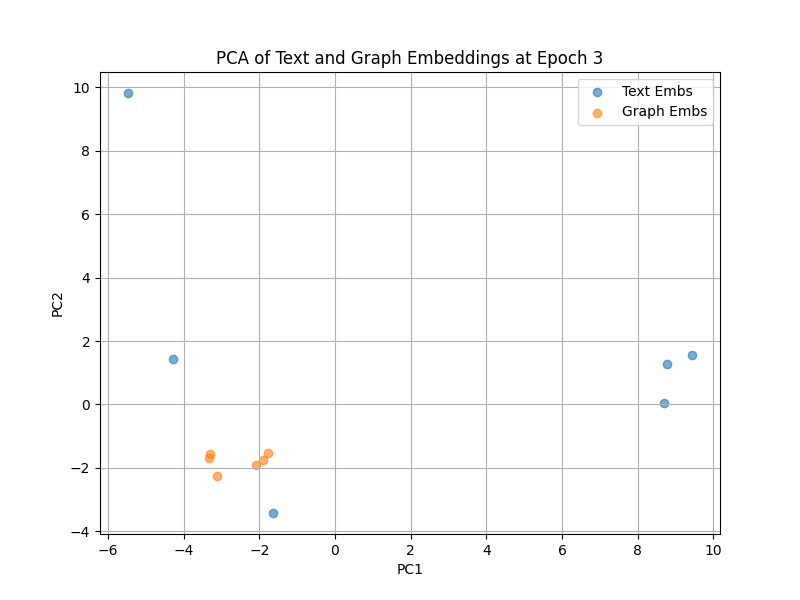}
    \caption{Intermediate epoch}
  \end{subfigure}
  \hfill
  \begin{subfigure}[t]{0.32\textwidth}
    \centering
    \includegraphics[trim=38 25 70 70, clip, width=\linewidth]{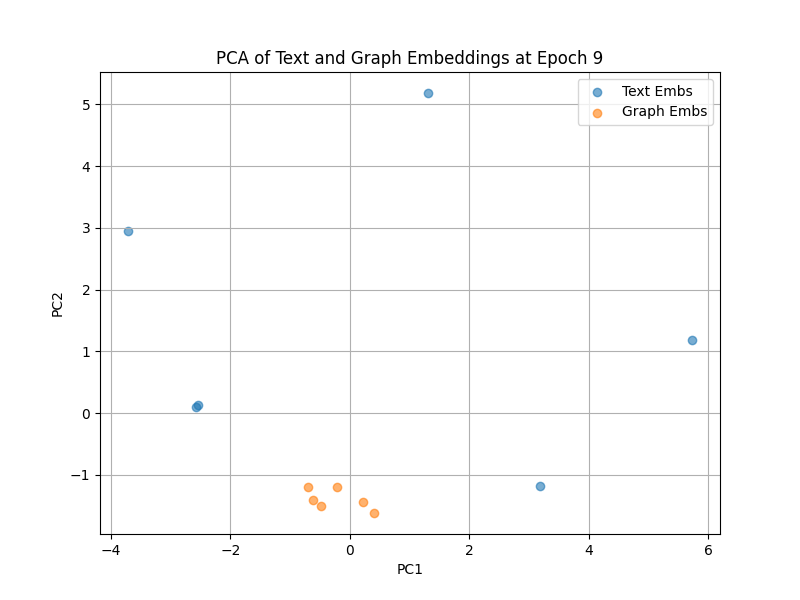}
    \caption{Final epoch}
  \end{subfigure}
  
  \vspace{1em}
  
  \begin{subfigure}[t]{0.48\textwidth}
    \centering
    \includegraphics[trim=0 10 0 440, clip, width=0.8\linewidth]{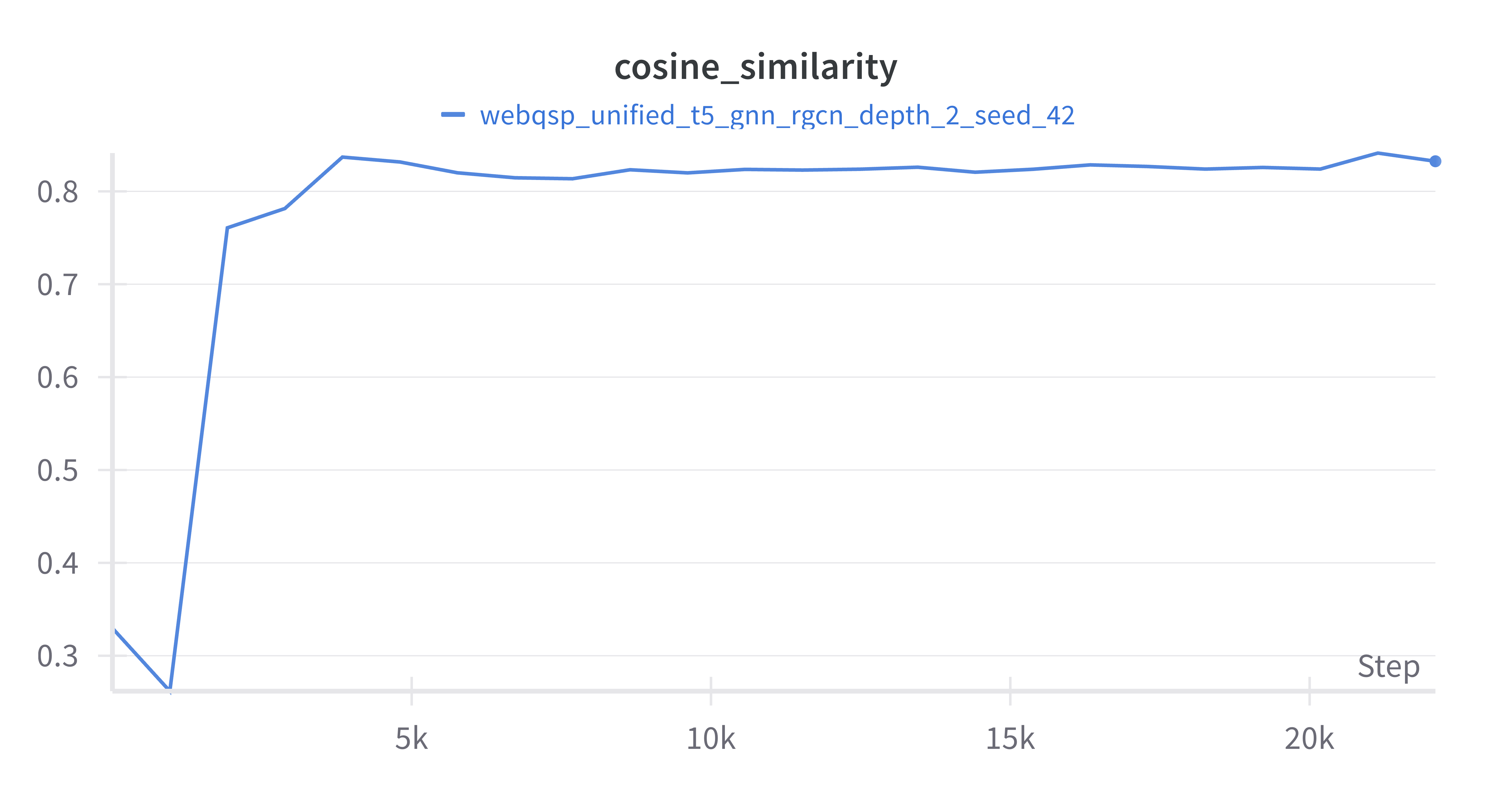}
    \caption{Cosine similarity}
  \end{subfigure}
  \hfill
  \begin{subfigure}[t]{0.48\textwidth}
    \centering
    \includegraphics[trim=0 10 0 440, clip, width=0.8\linewidth]{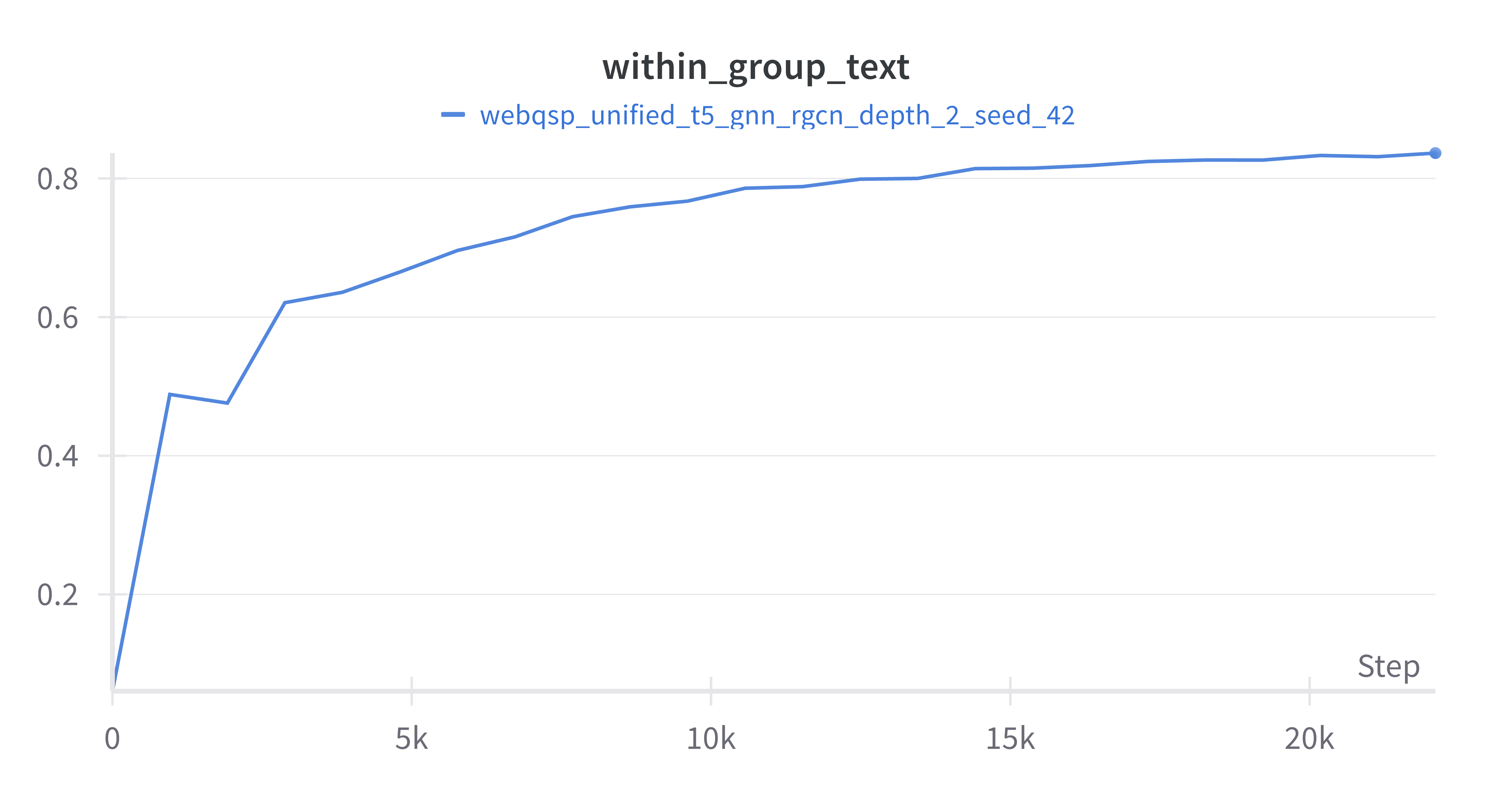}
    \caption{Distance within text}
  \end{subfigure}

  \vspace{1em}

  \begin{subfigure}[t]{0.48\textwidth}
    \centering
    \includegraphics[trim=0 10 0 440, clip, width=0.8\linewidth]{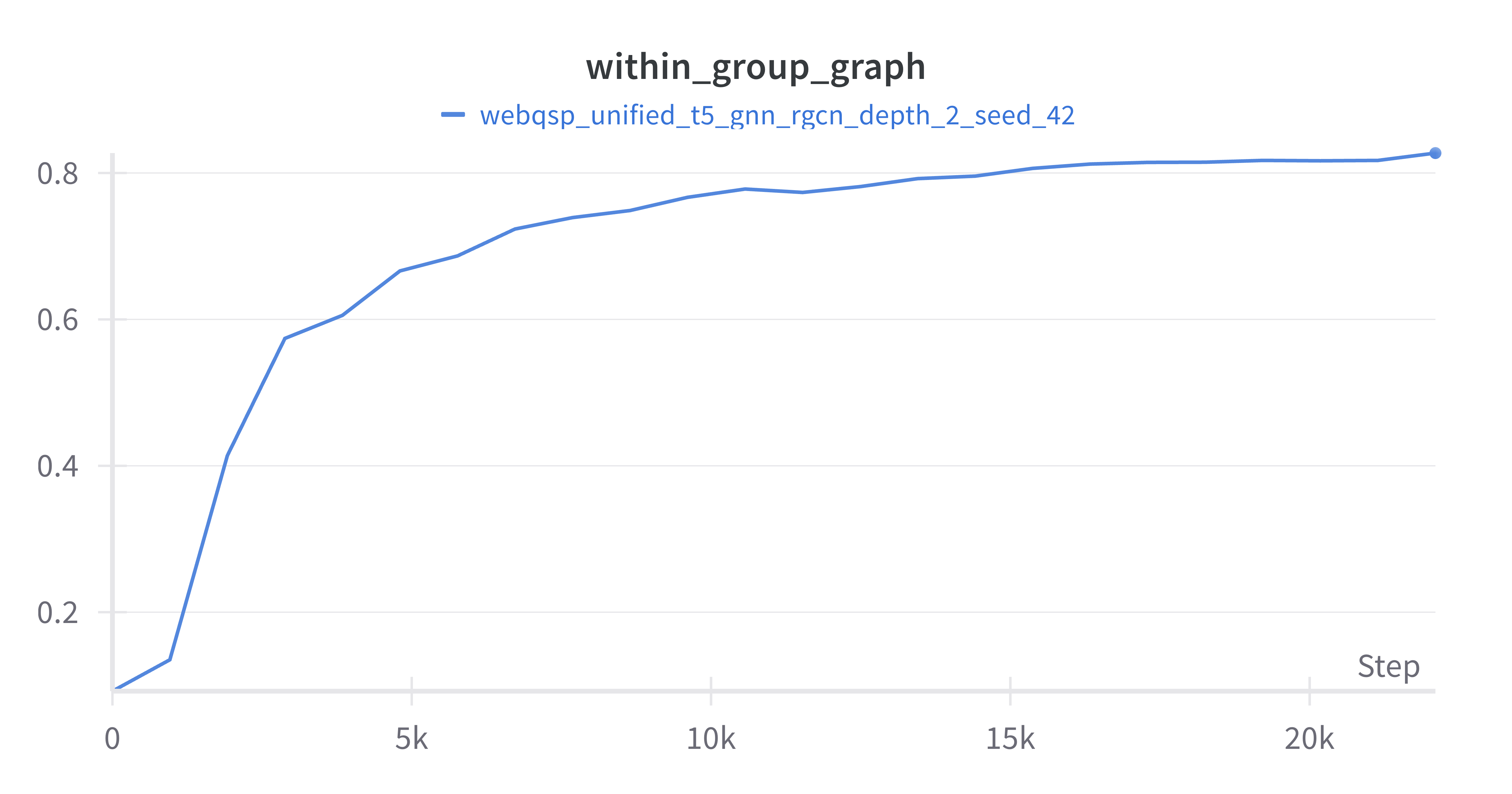}
    \caption{Distance within graph}
  \end{subfigure}
  \hfill
  \begin{subfigure}[t]{0.48\textwidth}
    \centering
    \includegraphics[trim=0 10 0 440, clip, width=0.8\linewidth]{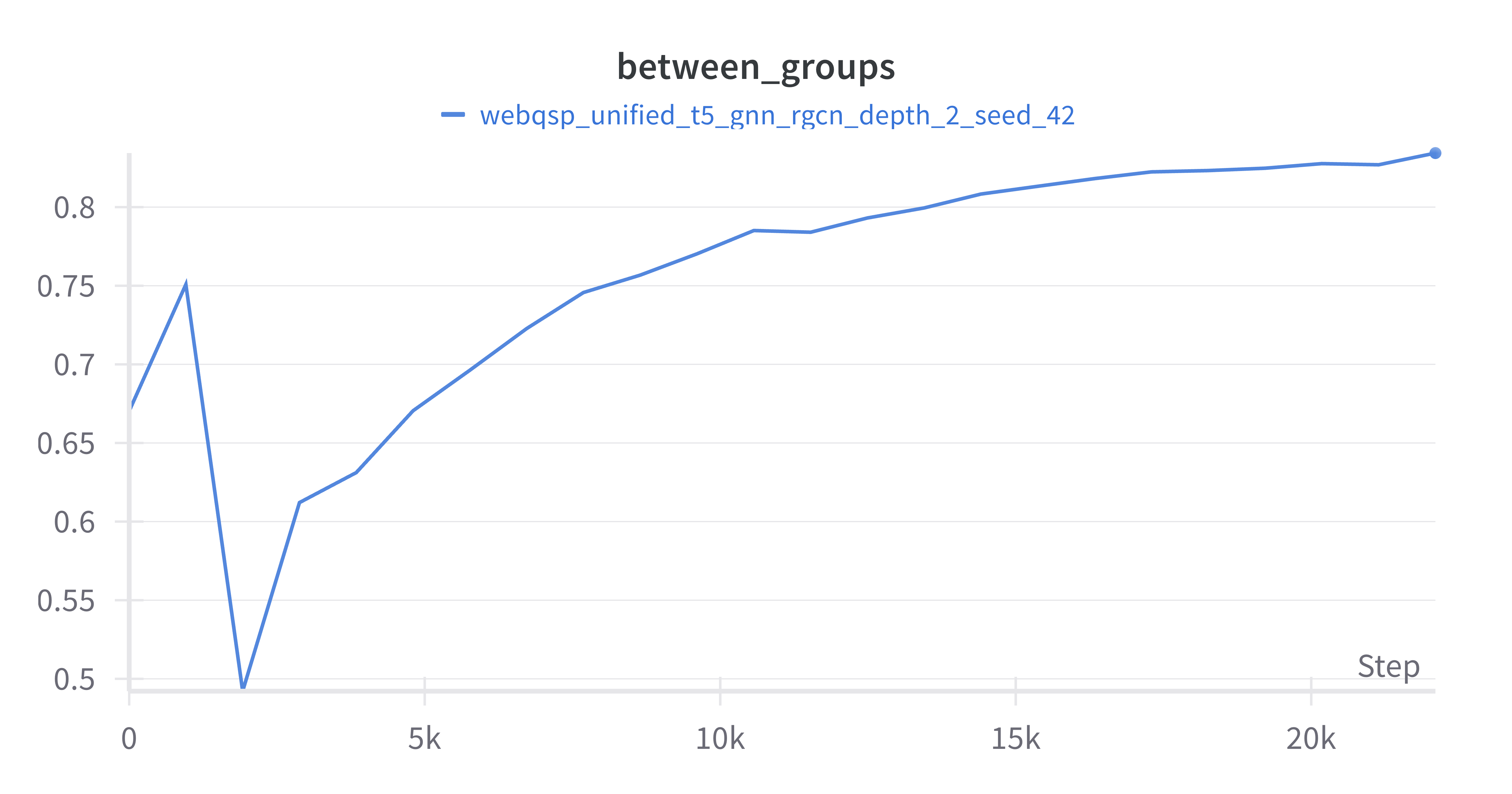}
    \caption{Distance between text and graph}
  \end{subfigure}

  \end{minipage}
  }
  % \vspace{\baselineskip}
  \caption{
    Results for reasoning pattern prediction on the WebQSP dataset. The text and graph representations move closer but stay largely separable. The between-group distance increases during training.}
  \label{fig:iso_results}
\end{figure*}

Across the tasks, we observe that hybrid models consistently outperform the text-only and graph-only baselines, and that incorporating the CoD loss leads to further gains. The only exception is for MLRE where the hybrid approaches achieve performance comparable to the text-based baseline, possibly because the graph representations fail to capture any complementary signals. Prior work has demonstrated how large-scale pretraining enables transformer models to encode syntactic information within their parameters \citep{starace-etal-2023-probing,liu-etal-2024-fantastic} and thus employing off-the-shelf parsers to capture dependency information shows little promise \citep{sachan2021syntax}. 
% This contrasts with the ETRE task where the graph explicitly encodes temporal information not present in the text. 

For the KBQA tasks, where text and graph inputs aim to encode the same information albeit coming from two different formats, i.e. the linearized format (for the text) versus the topological structure (for the graph), CoD offers only marginal gains over the default hybrid setting. In contrast, tasks like FU where text and graph encode different information (form content versus layout structure from OCR), CoD shows more improvement. 

% \rd{In contrast, tasks like ETRE where the modalities encode different types of signals (semantic content versus syntactic relations), CoD shows much more significant improvement.}

% \footnotetext{Model was pretrained on a 1,000-example subset of its original pretraining corpus.}

% \footnotetext{We present results for one representative dataset per task due to resource constraints. Similar trends hold for other datasets.}

\subsection{RQ2: How do text and graph representations relate during learning?}

Using the representation analysis framework detailed in Section~\ref{section:framework}, we observe three qualitatively distinct trends in the spatial relationship between text and graph representations that aligns with the task spectrum proposed in Figure~\ref{fig:spectrum}: complementarity, partial alignment, and complete alignment.

% find consistent patterns of how text and graph representations complement each other that aligns with the task spectrum proposed in Figure~\ref{fig:spectrum}, ranging from representational divergence to convergence. We provide extended results for all tasks in Appendix~\ref{appendix:all_results}.

\paragraph{Complementarity (ETRE):} The text and graph representations remain well-separated throughout training, signifying that they contribute distinct, complementary signals rather than converging towards a shared embedding space.

The PCA visualization confirms this complementarity. In Figure~\ref{fig:etre_tddman_results}, text and graph representations consistently occupy distinct regions. 
% The cosine similarity between them starts relatively low and remains bounded to 0.4. Moreover, the between-group distance remains higher than the within-group distances across training steps. 
% This divergence is even more significant for the MLRE task (Figure~\ref{fig:multiling_results}), where the between-group distance increases over time.
We attribute this separation to distinctiveness in how text and graph encode task-relevant information.
% In both tasks, there is no direct correspondence between tokens and nodes. 
In ETRE, the text representation provides local semantic cues around event mentions, while the graph encodes structural information in an attempt to quantify semantic temporal and discourse relations. 
% In MLRE, not only do the modalities encode different types of signals, but they also originate from different sources: the graph is derived from an external dependency parser rather than from the text directly. 
These structural and semantic divergences could lead text and graph representations to retain independent representation space.

\begin{figure*}[hbtp]
  \centering
    \resizebox{\textwidth}{!}{  % Resize to 90% of the page width
    \begin{minipage}{\textwidth}
      \centering
  
  % Row 1: Three PCA plots
  \begin{subfigure}[t]{0.32\textwidth}
    \centering
    \includegraphics[trim=38 25 70 70, clip, width=0.9\linewidth]{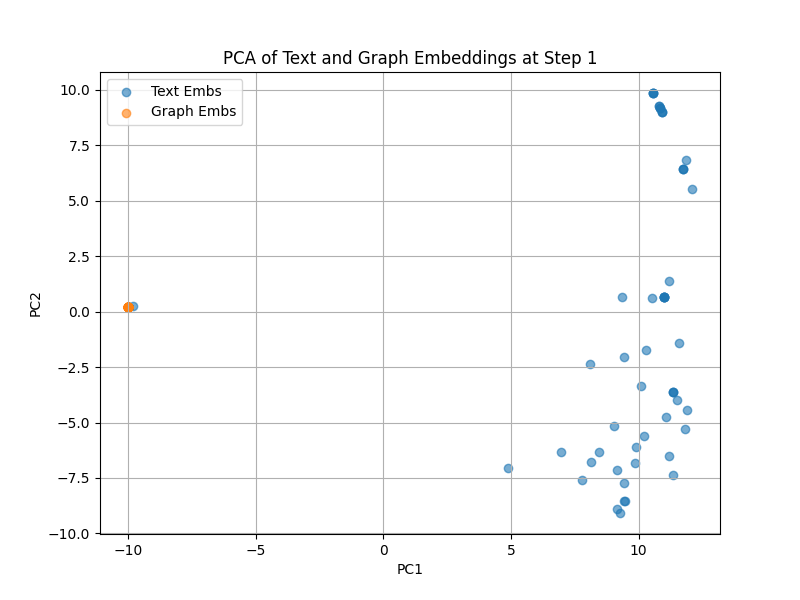}
    \caption{Initial epoch}
  \end{subfigure}
  \hfill
  \begin{subfigure}[t]{0.32\textwidth}
    \centering
    \includegraphics[trim=38 25 70 70, clip, width=0.9\linewidth]{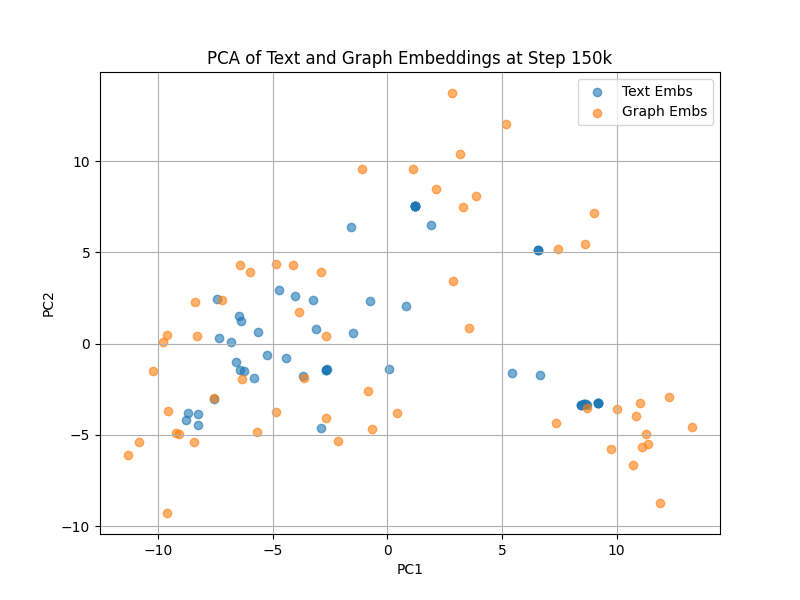}
    \caption{Intermediate epoch}
  \end{subfigure}
  \hfill
  \begin{subfigure}[t]{0.32\textwidth}
    \centering
    \includegraphics[trim=38 25 70 70, clip, width=0.9\linewidth]{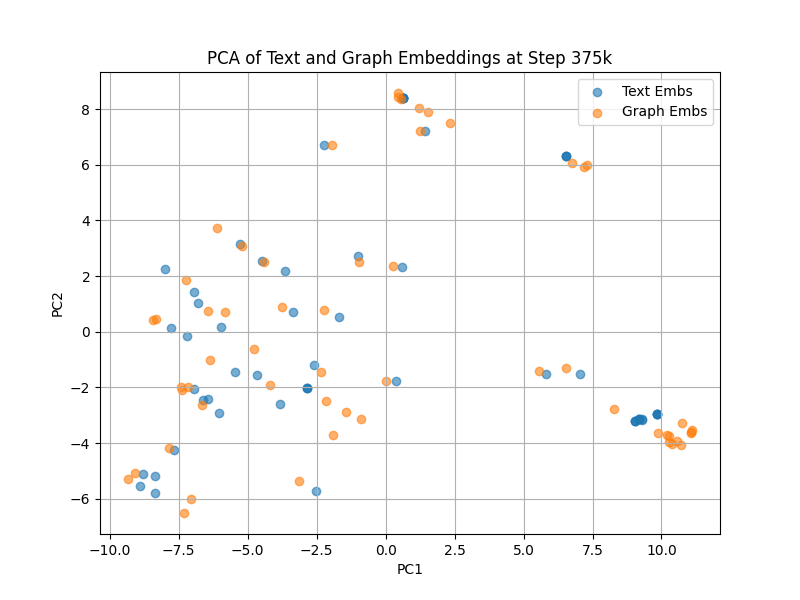}
    \caption{Final epoch}
  \end{subfigure}
  
  \vspace{1em}
  
  % Row 2: Four distance-related plots
  % \begin{subfigure}[t]{0.22\textwidth}
  %   \centering
  %   \includegraphics[width=\linewidth]{figures/for}
  %   \caption{Cosine similarity}
  % \end{subfigure}
  % \hfill
  \begin{subfigure}[t]{0.32\textwidth}
    \centering
    \includegraphics[trim=0 20 0 20, clip, width=0.9\linewidth]{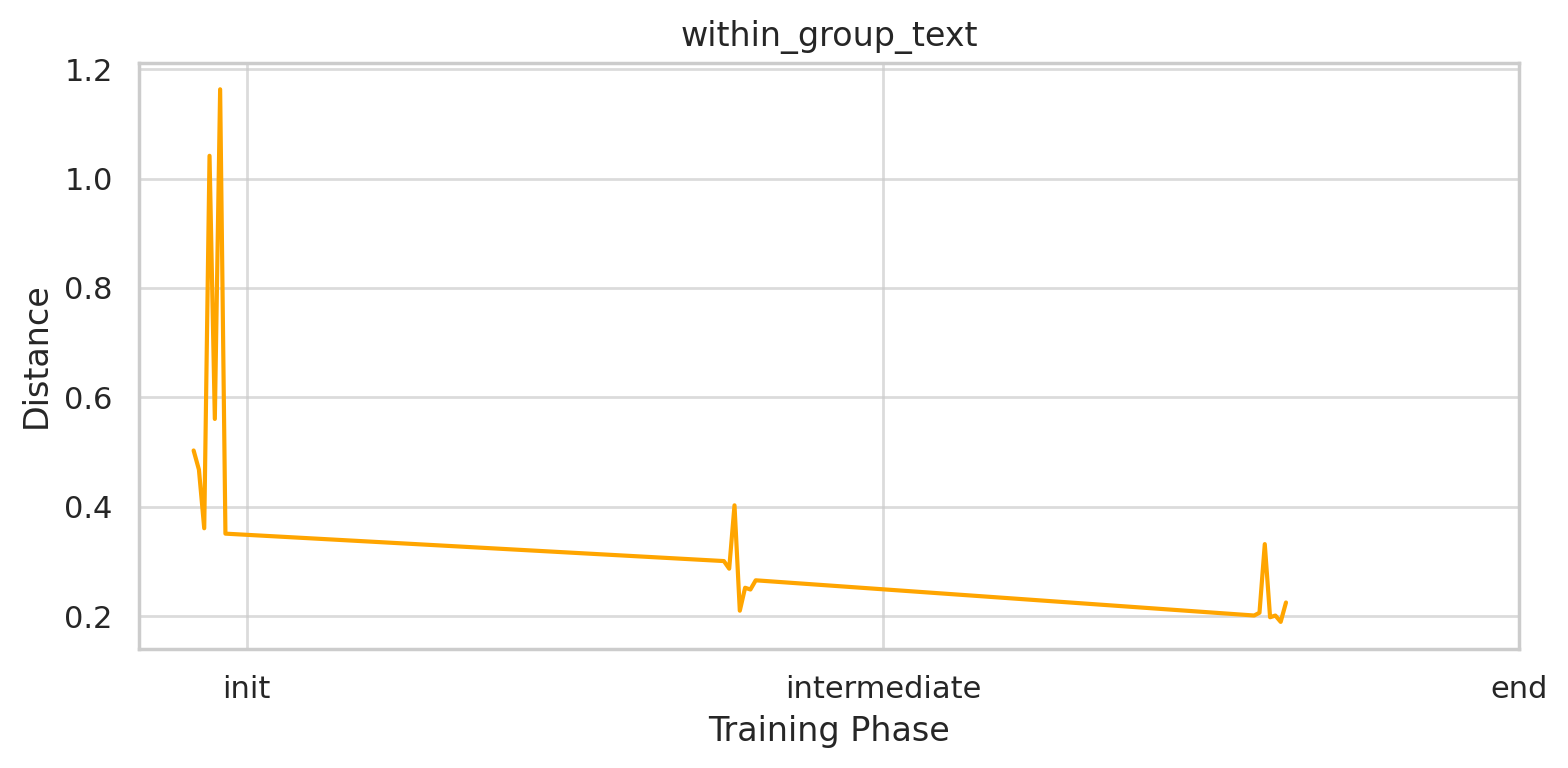}
    \caption{Distance within text}
  \end{subfigure}
  \hfill
  \begin{subfigure}[t]{0.32\textwidth}
    \centering
    \includegraphics[trim=0 20 0 20, clip, width=0.9\linewidth]{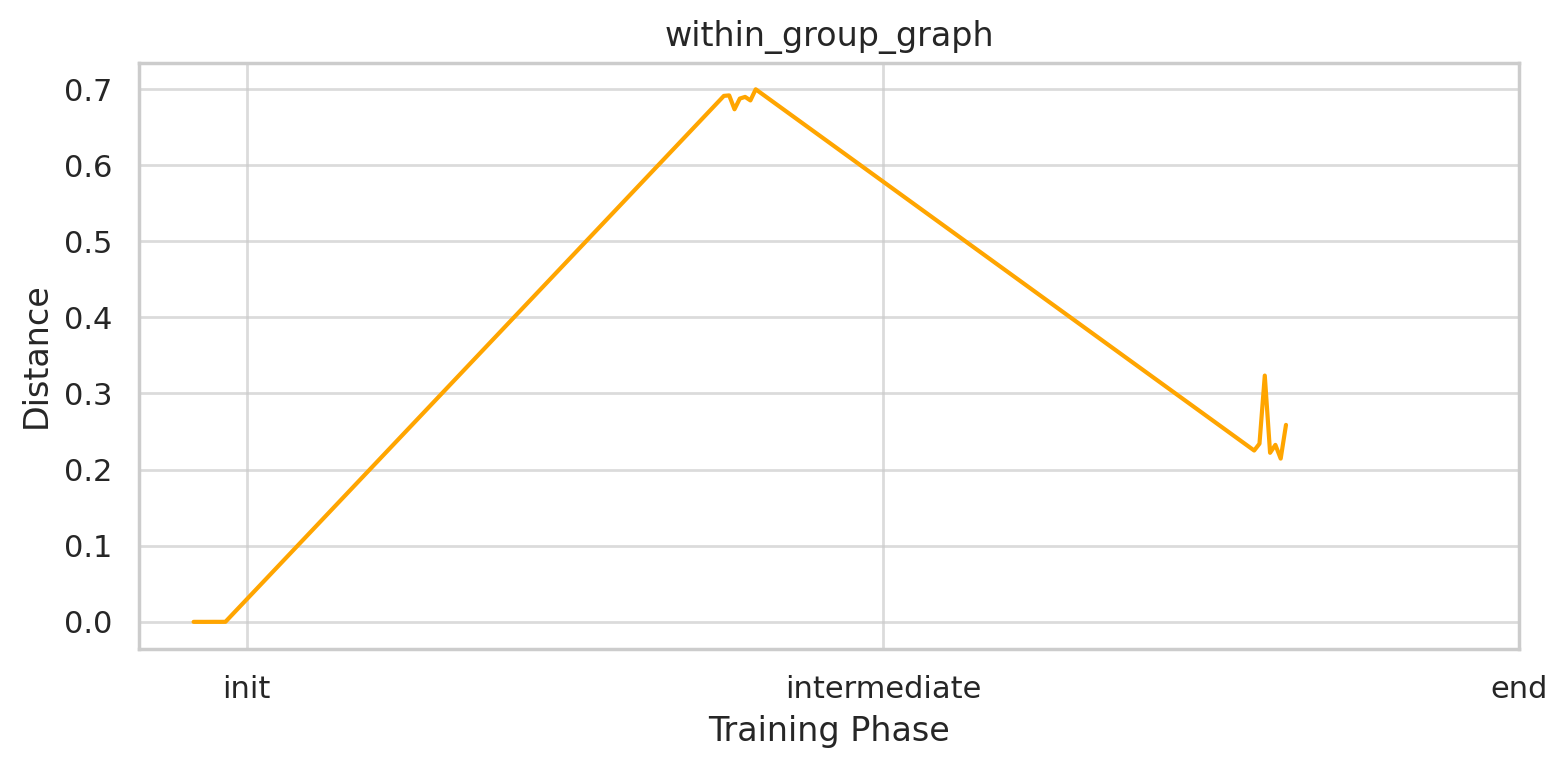}
    \caption{Distance within graph}
  \end{subfigure}
  \hfill
  \begin{subfigure}[t]{0.32\textwidth}
    \centering
    \includegraphics[trim=0 20 0 20, clip, width=0.9\linewidth]{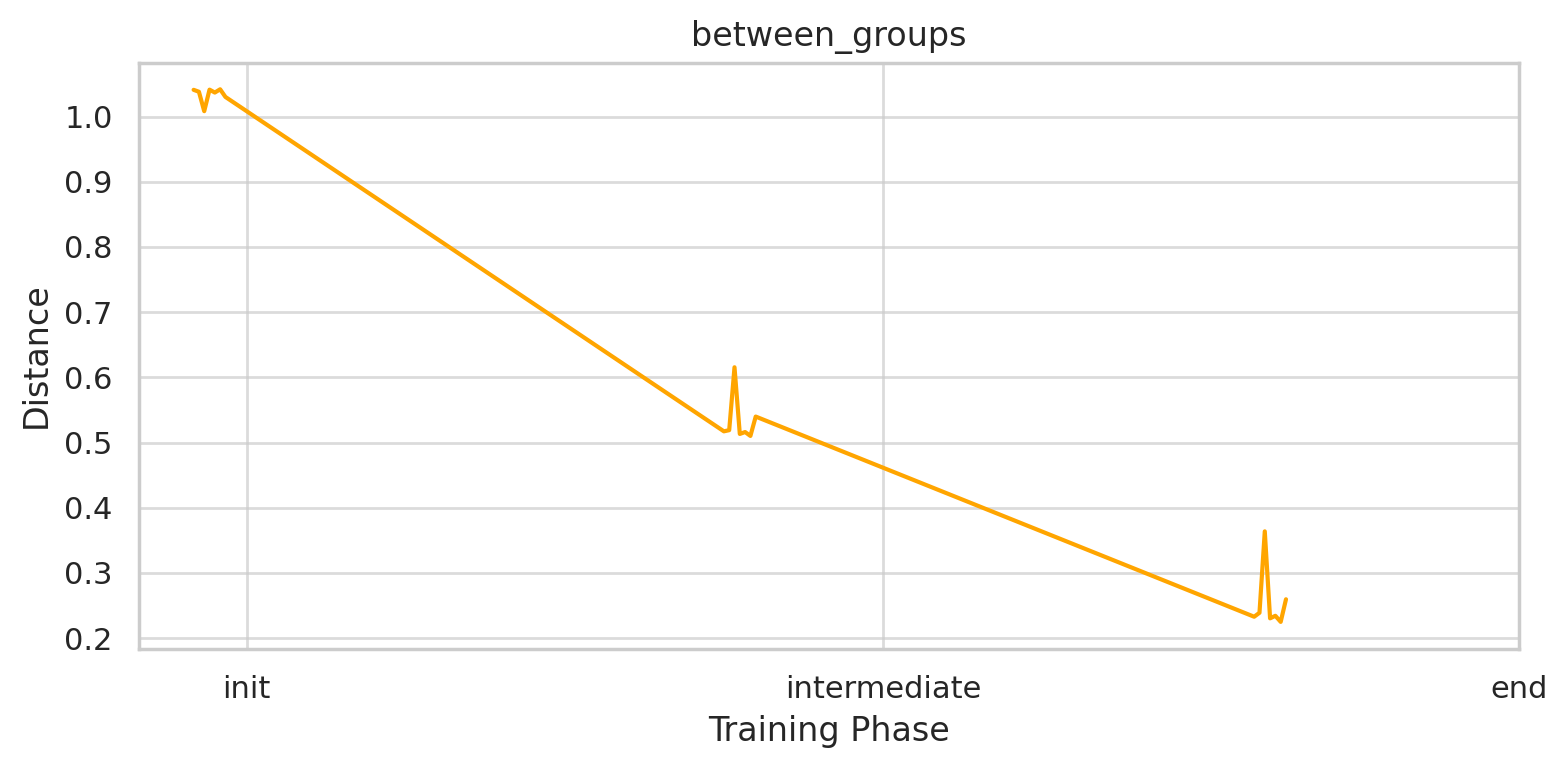}
    \caption{Distance between text and graph}
  \end{subfigure}
  % \vspace{\baselineskip}
  \end{minipage}
  }
  \caption{Results for form understanding on the CORD dataset.\protect\footnotemark  The text and graph representations draw closer and form overlapping clusters during training, and the between-group distance decreases and eventually approaches the within-group distances.}
  \label{fig:form_results}
\end{figure*}

\paragraph{Partial alignment (MLRE and RPP):} We observe that MLRE and RPP exhibit moderate convergence between text and graph. PCA visualizations (Figure~\ref{fig:REDFM_PCA_results} and~\ref{fig:iso_results}) show that the text and graph representations move closer in the shared space during training, yet remain largely separable. 
% For RPP, cosine similarity increases steadily, and between-group distances decrease relative to within-group. 
This suggests that text and graph are aligning but do not collapse into a single unified cluster. This behavior aligns with the task objective: while the inputs encode equivalent information, the objective is to classify the reasoning path traversed in the graph, not specific tokens or nodes. Thus, the text and graph representations can evolve in parallel without needing to fully align. This allows each of them to retain its inductive biases while adapting to shared learning signals through CoD.

% We see a similar story for MLRE. An interesting phenomena is that both the cosine similarity and the different distance metrics increase as training progresses; the former is attributed to the CoD loss and the latter due to the way the nodes are initialized with the token embeddings.

\paragraph{Complete alignment (FU and KBQA):} FU and KBQA appear near the alignment end of our spectrum. In both tasks, text and graph representations show strong convergence. E.g., PCA visualizations in Figure~\ref{fig:form_results} show a clear alignment trajectory: initial representations are moderately separated in the shared space, but progressively draw closer during training. By the final epochs, the paired embeddings often form overlapping clusters. 
% Also the between-group distance consistently decreases and eventually approximates or falls below the within-group distances. 

We explain this finding by establishing the fine-grained correspondence in input structure between text and graph for both tasks. Each graph node $v_i \in V$ has a clear textual counterpart as a token span $s_i \subseteq q$. In FU, OCR tokens are linked to spatially grounded nodes, while in entity ranking, candidate answer entities are matched between graph nodes and text tokens. This one-to-one correspondence likely encourages representations to align.
\footnotetext{We present distance metrics for three representative training phases due to resource constraints.}

To complement the PCA-based categorization, we further analyze cosine similarity and distance metrics to offer quantitative insight into the degree of alignment.

\paragraph{Cosine similarity increases consistently across tasks} due to the CoD objective, which encourages directional agreement between text and graph representations. However, in contrast to other tasks, tasks characterized by complementarity like ETRE exhibits a weaker increase, whose cosine similarity remains bounded to 0.4. 

\paragraph{Alignment is indicated by how closely between-group distances match within-group ones.} 
% When the between-group distance of the representations approaches that of the within-group distances, the embeddings display alignment rather than complementarity. 
In ETRE, the between-group distance remains consistently higher than the within-group distances, which reflects strong complementarity. 
% In both partial and complete alignment tasks, within-group distances increase over training.
Otherwise, we observe patterns on the alignment side.
However, the between-group distance trends between partial and complete alignment tasks diverge: in partial alignment tasks like MLRE and RPP, the between-group distance increases, whereas in complete alignment tasks such as FU and KBQA, the between-group distance steadily decreases and eventually approaches the within-group distances.

% This pattern is evident in tasks like FU and KBQA, where the between-group distance steadily decreases and eventually matches or falls below the within-group distances. In contrast, ETRE maintains a higher between-group distance throughout training. 

\subsection{RQ3: How do task characteristics shape the effects of CoD?}

Building on the representation patterns observed in RQ2, we now examine what task-specific characteristics may shape how CoD influences learning. We only consider the cases where adding in CoD consistently brought about performance gains.  

\paragraph{Same input, different task objectives}
Although RPP and KBQA share the identical input, they differ in task objectives: the former identifies reasoning patterns in the subgraph on a global level, whereas the latter scores individual entities at a local level. Despite this shared input, the learned representations behave differently under CoD. RPP demonstrates partial convergence, while KBQA shows strong alignment. This contrast suggests that the level at which reasoning is required, in this case global vs.\ local, can shape how the representations align in the representation space.

\paragraph{Same reasoning scope, different graph construction:}

ETRE and FU both involve localized reasoning between pairs: event mentions in ETRE and field spans in forms. However, their graph constructions differ in how directly they support the task. In FU, edges explicitly capture spatial layout relations that closely match the key–value associations being predicted. In ETRE, the graph encodes distinct layers of linguistic cues (e.g., syntax, discourse), which support but do not directly define the target temporal relation. Under CoD, FU shows complete alignment while ETRE demonstrates complementarity. This indicates that how well the graph structure reflects the task objective can influence whether CoD promotes complementarity or alignment.

\paragraph{With or without token-node correspondence}
In FU and KBQA entity-ranking, there is a strong one-to-one correspondence between graph nodes and text token spans. This provides a scaffold that supports representational convergence, which is reinforced through CoD. This is in contrast to complementary information encoded in ETRE, where representations remain more distinct, and CoD preserves separation. This highlights that explicit token–node correspondence could act as a structural prior that facilitates CoD towards alignment.

\section{Conclusion}
We analyze how text and graph representations complement each other during learning within a unified, task-agnostic framework using contrastive co-distillation (CoD) as a lens. We select five diverse relational reasoning tasks and observe a spectrum of representational behaviors from alignment to complementarity shaped by differences in task structure, such as whether the graph encodes the prediction target explicitly, whether nodes correspond directly to textual spans, and whether reasoning operates at a local or global level. These findings improve our understanding of text-graph representation relations and offer practical insights into applying CoD in structured NLP tasks.

\section{Limitations}
\paragraph{Task coverage} While our selected five relational reasoning tasks covers a broad spectrum of complementarity and alignment patterns, extending the framework to other tasks may reveal additional representational behaviors.

\paragraph{Analysis metrics} We rely on PCA visualizations and cosine/distance-based metrics for representation analysis. These methods provide interpretable trends but may not capture all fine-grained or non-linear interactions between text and graph, which could be explored with advanced probing or disentanglement techniques.

\section{Ethical considerations}
\paragraph{Bias propagation} Our framework builds on pretrained text and graph encoders, which may inherit and amplify biases present in the underlying data sources.

% \section{Acknowledgements}
% This work is supported by NSF DRK12-2405615 and NSF ITEST-2241670.

\bibliography{acl_latex}

\begin{thebibliography}{65}
\providecommand{\natexlab}[1]{#1}

\bibitem[{Belinkov(2021)}]{belinkov2021probingclassifierspromisesshortcomings}
Yonatan Belinkov. 2021.
\newblock \href {https://arxiv.org/abs/2102.12452} {Probing classifiers: Promises, shortcomings, and advances}.
\newblock \emph{Preprint}, arXiv:2102.12452.

\bibitem[{Bruna et~al.(2014)Bruna, Zaremba, Szlam, and LeCun}]{bruna2014spectralnetworkslocallyconnected}
Joan Bruna, Wojciech Zaremba, Arthur Szlam, and Yann LeCun. 2014.
\newblock \href {https://arxiv.org/abs/1312.6203} {Spectral networks and locally connected networks on graphs}.
\newblock \emph{Preprint}, arXiv:1312.6203.

\bibitem[{Buciluǎ et~al.(2006)Buciluǎ, Caruana, and Niculescu-Misil}]{bucilua06compression}
Cristian Buciluǎ, Rich Caruana, and Alexandru Niculescu-Misil. 2006.
\newblock Model compression.
\newblock In \emph{In Proceedings of the 12th ACM SIGKDD international conference on Knowledge discovery and data mining}, pages 535--541.

\bibitem[{Busbridge et~al.(2019)Busbridge, Sherburn, Cavallo, and Hammerla}]{busbridge2019relationalgraphattentionnetworks}
Dan Busbridge, Dane Sherburn, Pietro Cavallo, and Nils~Y. Hammerla. 2019.
\newblock \href {https://arxiv.org/abs/1904.05811} {Relational graph attention networks}.
\newblock \emph{Preprint}, arXiv:1904.05811.

\bibitem[{Cassidy et~al.(2014)Cassidy, McDowell, Chambers, and Bethard}]{tbdense}
Taylor Cassidy, Bill McDowell, Nathanael Chambers, and Steven Bethard. 2014.
\newblock An annotation framework for dense event ordering.
\newblock In \emph{Proceedings of the 52nd Annual Meeting of the Association for Computational Linguistics (Volume 2: Short Papers)}, pages 501--506.

\bibitem[{Chen and He(2021)}]{Chen_2021_CVPR}
Xinlei Chen and Kaiming He. 2021.
\newblock Exploring simple siamese representation learning.
\newblock In \emph{Proceedings of the IEEE/CVF Conference on Computer Vision and Pattern Recognition (CVPR)}, pages 15750--15758.

\bibitem[{Chen et~al.(2025)Chen, Liu, Zheng, Wen, Peng, Zhang, and Stiefelhagen}]{chen2025graphbaseddocumentstructureanalysis}
Yufan Chen, Ruiping Liu, Junwei Zheng, Di~Wen, Kunyu Peng, Jiaming Zhang, and Rainer Stiefelhagen. 2025.
\newblock \href {https://arxiv.org/abs/2502.02501} {Graph-based document structure analysis}.
\newblock \emph{Preprint}, arXiv:2502.02501.

\bibitem[{Christopoulou et~al.(2019)Christopoulou, Miwa, and Ananiadou}]{christopoulou2019connectingdotsdocumentlevelneural}
Fenia Christopoulou, Makoto Miwa, and Sophia Ananiadou. 2019.
\newblock \href {https://arxiv.org/abs/1909.00228} {Connecting the dots: Document-level neural relation extraction with edge-oriented graphs}.
\newblock \emph{Preprint}, arXiv:1909.00228.

\bibitem[{Cunningham et~al.(2023)Cunningham, Ewart, Riggs, Huben, and Sharkey}]{cunningham2023sparseautoencodershighlyinterpretable}
Hoagy Cunningham, Aidan Ewart, Logan Riggs, Robert Huben, and Lee Sharkey. 2023.
\newblock \href {https://arxiv.org/abs/2309.08600} {Sparse autoencoders find highly interpretable features in language models}.
\newblock \emph{Preprint}, arXiv:2309.08600.

\bibitem[{Devlin et~al.(2019)Devlin, Chang, Lee, and Toutanova}]{devlin2019bertpretrainingdeepbidirectional}
Jacob Devlin, Ming-Wei Chang, Kenton Lee, and Kristina Toutanova. 2019.
\newblock \href {https://arxiv.org/abs/1810.04805} {Bert: Pre-training of deep bidirectional transformers for language understanding}.
\newblock \emph{Preprint}, arXiv:1810.04805.

\bibitem[{Dutt et~al.(2022)Dutt, Bhattacharjee, Gangadharaiah, Roth, and Rose}]{dutt2022perkgqa}
Ritam Dutt, Kasturi Bhattacharjee, Rashmi Gangadharaiah, Dan Roth, and Carolyn Rose. 2022.
\newblock Perkgqa: Question answering over personalized knowledge graphs.
\newblock In \emph{Findings of the Association for Computational Linguistics: NAACL 2022}, pages 253--268.

\bibitem[{Dutt et~al.(2023)Dutt, Khosla, Bannihatti~Kumar, and Gangadharaiah}]{grailqapp}
Ritam Dutt, Sopan Khosla, Vinayshekhar Bannihatti~Kumar, and Rashmi Gangadharaiah. 2023.
\newblock \href {https://doi.org/10.18653/v1/2023.ijcnlp-main.58} {{G}rail{QA}++: A challenging zero-shot benchmark for knowledge base question answering}.
\newblock In \emph{Proceedings of the 13th International Joint Conference on Natural Language Processing and the 3rd Conference of the Asia-Pacific Chapter of the Association for Computational Linguistics (Volume 1: Long Papers)}, pages 897--909, Nusa Dua, Bali. Association for Computational Linguistics.

\bibitem[{Feng and He(2025)}]{feng-he-2025-rgr}
Tengfei Feng and Liang He. 2025.
\newblock \href {https://aclanthology.org/2025.coling-main.205/} {{RGR}-{KBQA}: Generating logical forms for question answering using knowledge-graph-enhanced large language model}.
\newblock In \emph{Proceedings of the 31st International Conference on Computational Linguistics}, pages 3057--3070, Abu Dhabi, UAE. Association for Computational Linguistics.

\bibitem[{Ferrone and Zanzotto(2020)}]{10.3389/frobt.2019.00153}
Lorenzo Ferrone and Fabio~Massimo Zanzotto. 2020.
\newblock \href {https://doi.org/10.3389/frobt.2019.00153} {Symbolic, distributed, and distributional representations for natural language processing in the era of deep learning: A survey}.
\newblock \emph{Frontiers in Robotics and AI}, 6.

\bibitem[{Fu et~al.(2021)Fu, Zhou, Yang, Tang, Liu, Liu, and Li}]{fu2021lrc}
Hao Fu, Shaojun Zhou, Qihong Yang, Junjie Tang, Guiquan Liu, Kaikui Liu, and Xiaolong Li. 2021.
\newblock Lrc-bert: latent-representation contrastive knowledge distillation for natural language understanding.
\newblock In \emph{Proceedings of the AAAI Conference on Artificial Intelligence}, volume~35, pages 12830--12838.

\bibitem[{Gao et~al.(2024)Gao, la~Tour, Tillman, Goh, Troll, Radford, Sutskever, Leike, and Wu}]{gao2024scalingevaluatingsparseautoencoders}
Leo Gao, Tom~Dupré la~Tour, Henk Tillman, Gabriel Goh, Rajan Troll, Alec Radford, Ilya Sutskever, Jan Leike, and Jeffrey Wu. 2024.
\newblock \href {https://arxiv.org/abs/2406.04093} {Scaling and evaluating sparse autoencoders}.
\newblock \emph{Preprint}, arXiv:2406.04093.

\bibitem[{Gao et~al.(2025)Gao, Lau, and Qi}]{gao2025seendataimprovingkbqa}
Shengxiang Gao, Jey~Han Lau, and Jianzhong Qi. 2025.
\newblock \href {https://arxiv.org/abs/2502.12737} {Beyond seen data: Improving kbqa generalization through schema-guided logical form generation}.
\newblock \emph{Preprint}, arXiv:2502.12737.

\bibitem[{Gu et~al.(2021)Gu, Kase, Vanni, Sadler, Liang, Yan, and Su}]{gu2021beyond}
Yu~Gu, Sue Kase, Michelle Vanni, Brian Sadler, Percy Liang, Xifeng Yan, and Yu~Su. 2021.
\newblock Beyond iid: three levels of generalization for question answering on knowledge bases.
\newblock In \emph{Proceedings of the Web Conference 2021}, pages 3477--3488.

\bibitem[{Guo et~al.(2020)Guo, Zhang, and Lu}]{guo2020attentionguidedgraphconvolutional}
Zhijiang Guo, Yan Zhang, and Wei Lu. 2020.
\newblock \href {https://arxiv.org/abs/1906.07510} {Attention guided graph convolutional networks for relation extraction}.
\newblock \emph{Preprint}, arXiv:1906.07510.

\bibitem[{Gupta et~al.(2015)Gupta, Boleda, Baroni, and Pad{\'o}}]{gupta-etal-2015-distributional}
Abhijeet Gupta, Gemma Boleda, Marco Baroni, and Sebastian Pad{\'o}. 2015.
\newblock \href {https://doi.org/10.18653/v1/D15-1002} {Distributional vectors encode referential attributes}.
\newblock In \emph{Proceedings of the 2015 Conference on Empirical Methods in Natural Language Processing}, pages 12--21, Lisbon, Portugal. Association for Computational Linguistics.

\bibitem[{Gururaja et~al.(2023)Gururaja, Dutt, Liao, and Rose}]{gururaja2023linguistic}
Sireesh Gururaja, Ritam Dutt, Tinglong Liao, and Carolyn Rose. 2023.
\newblock Linguistic representations for fewer-shot relation extraction across domains.
\newblock In \emph{Proceedings of the 61st Annual Meeting of the Association for Computational Linguistics (Volume 1: Long Papers)}, pages 7502--7514.

\bibitem[{Hinton et~al.(2015)Hinton, Vinyals, Dean et~al.}]{hinton2015distilling}
Geoffrey Hinton, Oriol Vinyals, Jeff Dean, and 1 others. 2015.
\newblock Distilling the knowledge in a neural network.
\newblock \emph{arXiv preprint arXiv:1503.02531}, 2(7).

\bibitem[{Huang et~al.(2019)Huang, Chen, He, Bai, Karatzas, Lu, and Jawahar}]{sroie}
Zheng Huang, Kai Chen, Jianhua He, Xiang Bai, Dimosthenis Karatzas, Shijian Lu, and CV~Jawahar. 2019.
\newblock Icdar2019 competition on scanned receipt ocr and information extraction.
\newblock In \emph{2019 International Conference on Document Analysis and Recognition (ICDAR)}, pages 1516--1520. IEEE.

\bibitem[{Huguet~Cabot et~al.(2023)Huguet~Cabot, Tedeschi, Ngonga~Ngomo, and Navigli}]{huguet-cabot-etal-2023-red}
Pere-Llu{\'i}s Huguet~Cabot, Simone Tedeschi, Axel-Cyrille Ngonga~Ngomo, and Roberto Navigli. 2023.
\newblock \href {https://doi.org/10.18653/v1/2023.acl-long.237} {{RED}$^{\textrm{fm}}$: a filtered and multilingual relation extraction dataset}.
\newblock In \emph{Proceedings of the 61st Annual Meeting of the Association for Computational Linguistics (Volume 1: Long Papers)}, pages 4326--4343, Toronto, Canada. Association for Computational Linguistics.

\bibitem[{Jaume et~al.(2019)Jaume, Ekenel, and Thiran}]{jaume2019funsd}
Guillaume Jaume, Hazim~Kemal Ekenel, and Jean-Philippe Thiran. 2019.
\newblock Funsd: A dataset for form understanding in noisy scanned documents.
\newblock In \emph{2019 International Conference on Document Analysis and Recognition Workshops (ICDARW)}, volume~2, pages 1--6. IEEE.

\bibitem[{Jiang and Usbeck(2022)}]{jiang2022knowledge}
Longquan Jiang and Ricardo Usbeck. 2022.
\newblock Knowledge graph question answering datasets and their generalizability: Are they enough for future research?
\newblock In \emph{Proceedings of the 45th International ACM SIGIR Conference on Research and Development in Information Retrieval}, pages 3209--3218.

\bibitem[{Lee et~al.(2023)Lee, Li, Zhang, Dozat, Perot, Su, Zhang, Sohn, Glushnev, Wang, Ainslie, Long, Qin, Fujii, Hua, and Pfister}]{lee-etal-2023-formnetv2}
Chen-Yu Lee, Chun-Liang Li, Hao Zhang, Timothy Dozat, Vincent Perot, Guolong Su, Xiang Zhang, Kihyuk Sohn, Nikolay Glushnev, Renshen Wang, Joshua Ainslie, Shangbang Long, Siyang Qin, Yasuhisa Fujii, Nan Hua, and Tomas Pfister. 2023.
\newblock \href {https://doi.org/10.18653/v1/2023.acl-long.501} {{F}orm{N}et{V}2: Multimodal graph contrastive learning for form document information extraction}.
\newblock In \emph{Proceedings of the 61st Annual Meeting of the Association for Computational Linguistics (Volume 1: Long Papers)}, pages 9011--9026, Toronto, Canada. Association for Computational Linguistics.

\bibitem[{Liang et~al.(2020)Liang, Hao, Shen, Zhou, Chen, Chen, and Carin}]{liang2020mixkd}
Kevin~J Liang, Weituo Hao, Dinghan Shen, Yufan Zhou, Weizhu Chen, Changyou Chen, and Lawrence Carin. 2020.
\newblock Mixkd: Towards efficient distillation of large-scale language models.
\newblock In \emph{International Conference on Learning Representations}.

\bibitem[{Lin et~al.(2025)Lin, Qian, Han, Choudhary, Wei, Wang, Genc, Huang, Wang, Subbian, Koutra, and Sun}]{lin2025gt2veclargelanguagemodels}
Jiacheng Lin, Kun Qian, Haoyu Han, Nurendra Choudhary, Tianxin Wei, Zhongruo Wang, Sahika Genc, Edward~W Huang, Sheng Wang, Karthik Subbian, Danai Koutra, and Jimeng Sun. 2025.
\newblock \href {https://arxiv.org/abs/2410.11235} {Gt2vec: Large language models as multi-modal encoders for text and graph-structured data}.
\newblock \emph{Preprint}, arXiv:2410.11235.

\bibitem[{Lin et~al.(2021)Lin, Meng, Sun, Han, Kuang, Li, and Wu}]{lin-etal-2021-bertgcn}
Yuxiao Lin, Yuxian Meng, Xiaofei Sun, Qinghong Han, Kun Kuang, Jiwei Li, and Fei Wu. 2021.
\newblock \href {https://doi.org/10.18653/v1/2021.findings-acl.126} {{B}ert{GCN}: Transductive text classification by combining {GNN} and {BERT}}.
\newblock In \emph{Findings of the Association for Computational Linguistics: ACL-IJCNLP 2021}, pages 1456--1462, Online. Association for Computational Linguistics.

\bibitem[{Liu et~al.(2022{\natexlab{a}})Liu, Tao, Feng, and Zhao}]{liu2022multi}
Chang Liu, Chongyang Tao, Jiazhan Feng, and Dongyan Zhao. 2022{\natexlab{a}}.
\newblock Multi-granularity structural knowledge distillation for language model compression.
\newblock In \emph{Proceedings of the 60th Annual Meeting of the Association for Computational Linguistics (Volume 1: Long Papers)}, pages 1001--1011.

\bibitem[{Liu et~al.(2022{\natexlab{b}})Liu, Wang, Raptis, and Fujii}]{liu2022unifiedlineparagraphdetection}
Shuang Liu, Renshen Wang, Michalis Raptis, and Yasuhisa Fujii. 2022{\natexlab{b}}.
\newblock \href {https://arxiv.org/abs/2203.09638} {Unified line and paragraph detection by graph convolutional networks}.
\newblock \emph{Preprint}, arXiv:2203.09638.

\bibitem[{Liu et~al.(2019)Liu, Ott, Goyal, Du, Joshi, Chen, Levy, Lewis, Zettlemoyer, and Stoyanov}]{liu2019robertarobustlyoptimizedbert}
Yinhan Liu, Myle Ott, Naman Goyal, Jingfei Du, Mandar Joshi, Danqi Chen, Omer Levy, Mike Lewis, Luke Zettlemoyer, and Veselin Stoyanov. 2019.
\newblock \href {https://arxiv.org/abs/1907.11692} {Roberta: A robustly optimized bert pretraining approach}.
\newblock \emph{Preprint}, arXiv:1907.11692.

\bibitem[{Liu et~al.(2024)Liu, Kong, Liu, and Sun}]{liu-etal-2024-fantastic}
Zhu Liu, Cunliang Kong, Ying Liu, and Maosong Sun. 2024.
\newblock \href {https://doi.org/10.18653/v1/2024.findings-acl.866} {Fantastic semantics and where to find them: Investigating which layers of generative {LLM}s reflect lexical semantics}.
\newblock In \emph{Findings of the Association for Computational Linguistics: ACL 2024}, pages 14551--14558, Bangkok, Thailand. Association for Computational Linguistics.

\bibitem[{Naik et~al.(2019)Naik, Breitfeller, and Rose}]{tdd}
Aakanksha Naik, Luke Breitfeller, and Carolyn Rose. 2019.
\newblock Tddiscourse: A dataset for discourse-level temporal ordering of events.
\newblock In \emph{Proceedings of the 20th Annual SIGdial Meeting on Discourse and Dialogue}, pages 239--249.

\bibitem[{Nastase et~al.(2015)Nastase, Mihalcea, and Radav}]{nastase15graphsurvey}
Vivi Nastase, Rada Mihalcea, and Dragomir~R. Radav. 2015.
\newblock A survey of graphs in natural language processing.
\newblock \emph{Natural Language Engineering}, 5:665--698.

\bibitem[{Ng et~al.(2011)}]{ng2011sparse}
Andrew Ng and 1 others. 2011.
\newblock Sparse autoencoder.
\newblock \emph{CS294A Lecture notes}, 72(2011):1--19.

\bibitem[{Nourbakhsh et~al.(2024)Nourbakhsh, Jin, Parekh, Shah, and Rose}]{nourbakhsh-etal-2024-aligatr}
Armineh Nourbakhsh, Zhao Jin, Siddharth Parekh, Sameena Shah, and Carolyn Rose. 2024.
\newblock \href {https://doi.org/10.18653/v1/2024.findings-emnlp.778} {{A}li{GAT}r: Graph-based layout generation for form understanding}.
\newblock In \emph{Findings of the Association for Computational Linguistics: EMNLP 2024}, pages 13309--13328, Miami, Florida, USA. Association for Computational Linguistics.

\bibitem[{Park et~al.(2019)Park, Shin, Lee, Lee, Surh, Seo, and Lee}]{park2019cord}
Seunghyun Park, Seung Shin, Bado Lee, Junyeop Lee, Jaeheung Surh, Minjoon Seo, and Hwalsuk Lee. 2019.
\newblock Cord: a consolidated receipt dataset for post-ocr parsing.
\newblock In \emph{Workshop on Document Intelligence at NeurIPS 2019}.

\bibitem[{Perozzi et~al.(2017)Perozzi, Kulkarni, Chen, and Skiena}]{walklets}
Bryan Perozzi, Vivek Kulkarni, Haochen Chen, and Steven Skiena. 2017.
\newblock Don't walk, skip! online learning of multi-scale network embeddings.
\newblock In \emph{Proceedings of the 2017 IEEE/ACM International Conference on Advances in Social Networks Analysis and Mining 2017}, pages 258--265.

\bibitem[{Qi et~al.(2020)Qi, Zhang, Zhang, Bolton, and Manning}]{qi2020stanza}
Peng Qi, Yuhao Zhang, Yuhui Zhang, Jason Bolton, and Christopher~D Manning. 2020.
\newblock Stanza: A python natural language processing toolkit for many human languages.
\newblock In \emph{Proceedings of the 58th Annual Meeting of the Association for Computational Linguistics: System Demonstrations}, pages 101--108.

\bibitem[{Raffel et~al.(2023)Raffel, Shazeer, Roberts, Lee, Narang, Matena, Zhou, Li, and Liu}]{raffel2023exploringlimitstransferlearning}
Colin Raffel, Noam Shazeer, Adam Roberts, Katherine Lee, Sharan Narang, Michael Matena, Yanqi Zhou, Wei Li, and Peter~J. Liu. 2023.
\newblock \href {https://arxiv.org/abs/1910.10683} {Exploring the limits of transfer learning with a unified text-to-text transformer}.
\newblock \emph{Preprint}, arXiv:1910.10683.

\bibitem[{Sachan et~al.(2021)Sachan, Zhang, Qi, and Hamilton}]{sachan2021syntax}
Devendra Sachan, Yuhao Zhang, Peng Qi, and William~L Hamilton. 2021.
\newblock Do syntax trees help pre-trained transformers extract information?
\newblock In \emph{Proceedings of the 16th Conference of the European Chapter of the Association for Computational Linguistics: Main Volume}, pages 2647--2661.

\bibitem[{Sanh et~al.(2019)Sanh, Debut, Chaumond, and Wolf}]{sanh2019distilbert}
Victor Sanh, Lysandre Debut, Julien Chaumond, and Thomas Wolf. 2019.
\newblock Distilbert, a distilled version of bert: smaller, faster, cheaper and lighter.
\newblock \emph{arXiv preprint arXiv:1910.01108}.

\bibitem[{Scarselli et~al.(2009)Scarselli, Gori, Tsoi, Hagenbuchner, and Monfardini}]{4700287}
Franco Scarselli, Marco Gori, Ah~Chung Tsoi, Markus Hagenbuchner, and Gabriele Monfardini. 2009.
\newblock \href {https://doi.org/10.1109/TNN.2008.2005605} {The graph neural network model}.
\newblock \emph{IEEE Transactions on Neural Networks}, 20(1):61--80.

\bibitem[{Schlichtkrull et~al.(2017)Schlichtkrull, Kipf, Bloem, van~den Berg, Titov, and Welling}]{schlichtkrull2017modelingrelationaldatagraph}
Michael Schlichtkrull, Thomas~N. Kipf, Peter Bloem, Rianne van~den Berg, Ivan Titov, and Max Welling. 2017.
\newblock \href {https://arxiv.org/abs/1703.06103} {Modeling relational data with graph convolutional networks}.
\newblock \emph{Preprint}, arXiv:1703.06103.

\bibitem[{Stanton et~al.(2021)Stanton, Izmailov, Kirichenko, Alemi, and Wilson}]{stanton2021fidelity}
Samuel Stanton, Pavel Izmailov, Polina Kirichenko, Alexander~A. Alemi, and Andrew~G. Wilson. 2021.
\newblock Does knowledge distillation really work?
\newblock \emph{Advances in neural information processing systems}, 34:6906--6919.

\bibitem[{Starace et~al.(2023)Starace, Papakostas, Choenni, Panagiotopoulos, Rosati, Leidinger, and Shutova}]{starace-etal-2023-probing}
Giulio Starace, Konstantinos Papakostas, Rochelle Choenni, Apostolos Panagiotopoulos, Matteo Rosati, Alina Leidinger, and Ekaterina Shutova. 2023.
\newblock \href {https://doi.org/10.18653/v1/2023.findings-emnlp.476} {Probing {LLM}s for joint encoding of linguistic categories}.
\newblock In \emph{Findings of the Association for Computational Linguistics: EMNLP 2023}, pages 7158--7179, Singapore. Association for Computational Linguistics.

\bibitem[{Sun et~al.(2018)Sun, Dhingra, Zaheer, Mazaitis, Salakhutdinov, and Cohen}]{sun2018opendomainquestionanswering}
Haitian Sun, Bhuwan Dhingra, Manzil Zaheer, Kathryn Mazaitis, Ruslan Salakhutdinov, and William~W. Cohen. 2018.
\newblock \href {https://arxiv.org/abs/1809.00782} {Open domain question answering using early fusion of knowledge bases and text}.
\newblock \emph{Preprint}, arXiv:1809.00782.

\bibitem[{Sun et~al.(2019)Sun, Cheng, Gan, and Liu}]{sun2019patient}
Siqi Sun, Yu~Cheng, Zhe Gan, and Jingjing Liu. 2019.
\newblock Patient knowledge distillation for bert model compression.
\newblock In \emph{Proceedings of the 2019 Conference on Empirical Methods in Natural Language Processing and the 9th International Joint Conference on Natural Language Processing (EMNLP-IJCNLP)}, pages 4323--4332.

\bibitem[{Sun et~al.(2020)Sun, Gan, Fang, Cheng, Wang, and Liu}]{sun2020contrastive}
Siqi Sun, Zhe Gan, Yuwei Fang, Yu~Cheng, Shuohang Wang, and Jingjing Liu. 2020.
\newblock Contrastive distillation on intermediate representations for language model compression.
\newblock In \emph{Proceedings of the 2020 Conference on Empirical Methods in Natural Language Processing (EMNLP)}, pages 498--508.

\bibitem[{Tian et~al.(2020)Tian, Krishnan, and Isola}]{tian2019crd}
Yonglong Tian, Dilip Krishnan, and Phillip Isola. 2020.
\newblock Contrastive representation distillation.
\newblock In \emph{International Conference on Learning Representations}.

\bibitem[{Tian et~al.(2022)Tian, Krishnan, and Isola}]{tian2022contrastiverepresentationdistillation}
Yonglong Tian, Dilip Krishnan, and Phillip Isola. 2022.
\newblock \href {https://arxiv.org/abs/1910.10699} {Contrastive representation distillation}.
\newblock \emph{Preprint}, arXiv:1910.10699.

\bibitem[{Tian et~al.(2024)Tian, Song, Wu, Zhou, Wang, Yang, Xu, Cao, and Wang}]{tian-etal-2024-augmenting}
Yuhang Tian, Dandan Song, Zhijing Wu, Changzhi Zhou, Hao Wang, Jun Yang, Jing Xu, Ruanmin Cao, and HaoYu Wang. 2024.
\newblock \href {https://doi.org/10.18653/v1/2024.findings-emnlp.699} {Augmenting reasoning capabilities of {LLM}s with graph structures in knowledge base question answering}.
\newblock In \emph{Findings of the Association for Computational Linguistics: EMNLP 2024}, pages 11967--11977, Miami, Florida, USA. Association for Computational Linguistics.

\bibitem[{Valdez-Valenzuela et~al.(2025)Valdez-Valenzuela, G{\'o}mez-Adorno, and Montes-y G{\'o}mez}]{valdez-gomez-adorno-2025-text}
Andric Valdez-Valenzuela, Helena G{\'o}mez-Adorno, and Manuel Montes-y G{\'o}mez. 2025.
\newblock \href {https://aclanthology.org/2025.genaidetect-1.10/} {Text graph neural networks for detecting {AI}-generated content}.
\newblock In \emph{Proceedings of the 1stWorkshop on GenAI Content Detection (GenAIDetect)}, pages 134--139, Abu Dhabi, UAE. International Conference on Computational Linguistics.

\bibitem[{Veličković et~al.(2018)Veličković, Cucurull, Casanova, Romero, Liò, and Bengio}]{veličković2018graphattentionnetworks}
Petar Veličković, Guillem Cucurull, Arantxa Casanova, Adriana Romero, Pietro Liò, and Yoshua Bengio. 2018.
\newblock \href {https://arxiv.org/abs/1710.10903} {Graph attention networks}.
\newblock \emph{Preprint}, arXiv:1710.10903.

\bibitem[{Wang et~al.(2023)Wang, Krumdick, Tong, Halim, Sokolov, Barda, Vendryes, and Tanner}]{wang2023graphicalapproachdocumentlayout}
Jilin Wang, Michael Krumdick, Baojia Tong, Hamima Halim, Maxim Sokolov, Vadym Barda, Delphine Vendryes, and Chris Tanner. 2023.
\newblock \href {https://arxiv.org/abs/2308.02051} {A graphical approach to document layout analysis}.
\newblock \emph{Preprint}, arXiv:2308.02051.

\bibitem[{Wang et~al.(2022)Wang, Fujii, and Popat}]{9706840}
Renshen Wang, Yasuhisa Fujii, and Ashok~C. Popat. 2022.
\newblock \href {https://doi.org/10.1109/WACV51458.2022.00259} {Post-ocr paragraph recognition by graph convolutional networks}.
\newblock In \emph{2022 IEEE/CVF Winter Conference on Applications of Computer Vision (WACV)}, pages 2533--2542.

\bibitem[{Xie et~al.(2022)Xie, Wu, Shi, Zhong, Scholak, Yasunaga, Wu, Zhong, Yin, Wang, Zhong, Wang, Li, Boyle, Ni, Yao, Radev, Xiong, Kong, Zhang, Smith, Zettlemoyer, and Yu}]{xie-etal-2022-unifiedskg}
Tianbao Xie, Chen~Henry Wu, Peng Shi, Ruiqi Zhong, Torsten Scholak, Michihiro Yasunaga, Chien-Sheng Wu, Ming Zhong, Pengcheng Yin, Sida~I. Wang, Victor Zhong, Bailin Wang, Chengzu Li, Connor Boyle, Ansong Ni, Ziyu Yao, Dragomir Radev, Caiming Xiong, Lingpeng Kong, and 4 others. 2022.
\newblock \href {https://doi.org/10.18653/v1/2022.emnlp-main.39} {{U}nified{SKG}: Unifying and multi-tasking structured knowledge grounding with text-to-text language models}.
\newblock In \emph{Proceedings of the 2022 Conference on Empirical Methods in Natural Language Processing}, pages 602--631, Abu Dhabi, United Arab Emirates. Association for Computational Linguistics.

\bibitem[{Yao et~al.(2024)Yao, Breitfeller, Naik, Zhou, and Rose}]{yao2024distilling}
Hao-Ren Yao, Luke Breitfeller, Aakanksha Naik, Chunxiao Zhou, and Carolyn Rose. 2024.
\newblock Distilling multi-scale knowledge for event temporal relation extraction.
\newblock In \emph{Proceedings of the 33rd ACM International Conference on Information and Knowledge Management}, pages 2971--2980.

\bibitem[{Yao et~al.(2018)Yao, Mao, and Luo}]{yao2018graphconvolutionalnetworkstext}
Liang Yao, Chengsheng Mao, and Yuan Luo. 2018.
\newblock \href {https://arxiv.org/abs/1809.05679} {Graph convolutional networks for text classification}.
\newblock \emph{Preprint}, arXiv:1809.05679.

\bibitem[{Yasunaga et~al.(2022)Yasunaga, Ren, Bosselut, Liang, and Leskovec}]{yasunaga2022qagnnreasoninglanguagemodels}
Michihiro Yasunaga, Hongyu Ren, Antoine Bosselut, Percy Liang, and Jure Leskovec. 2022.
\newblock \href {https://arxiv.org/abs/2104.06378} {Qa-gnn: Reasoning with language models and knowledge graphs for question answering}.
\newblock \emph{Preprint}, arXiv:2104.06378.

\bibitem[{Yih et~al.(2016)Yih, Richardson, Meek, Chang, and Suh}]{WebQSP}
Wen-tau Yih, Matthew Richardson, Chris Meek, Ming-Wei Chang, and Jina Suh. 2016.
\newblock \href {https://doi.org/10.18653/v1/P16-2033} {The value of semantic parse labeling for knowledge base question answering}.
\newblock In \emph{Proceedings of the 54th Annual Meeting of the Association for Computational Linguistics (Volume 2: Short Papers)}, pages 201--206, Berlin, Germany. Association for Computational Linguistics.

\bibitem[{Zhang et~al.(2018{\natexlab{a}})Zhang, Xiang, Hospedales, and Lu}]{zhang2018deep}
Ying Zhang, Tao Xiang, Timothy~M Hospedales, and Huchuan Lu. 2018{\natexlab{a}}.
\newblock Deep mutual learning.
\newblock In \emph{Proceedings of the IEEE conference on computer vision and pattern recognition}, pages 4320--4328.

\bibitem[{Zhang et~al.(2018{\natexlab{b}})Zhang, Qi, and Manning}]{zhang2018graphconvolutionpruneddependency}
Yuhao Zhang, Peng Qi, and Christopher~D. Manning. 2018{\natexlab{b}}.
\newblock \href {https://arxiv.org/abs/1809.10185} {Graph convolution over pruned dependency trees improves relation extraction}.
\newblock \emph{Preprint}, arXiv:1809.10185.

\end{thebibliography}

\newpage
\clearpage

\appendix
% \section{Appendix}
\renewcommand{\thesection}{\Alph{section}}

\begin{table*}[ht]
    \centering
    \renewcommand{\arraystretch}{1.3} % Adjust line spacing
    \setlength{\tabcolsep}{5pt} % Adjust column spacing
    \resizebox{\textwidth}{!}{ % Ensures table fits within page width
    \begin{tabular}{ccp{5.0cm}p{3.5cm}p{6.0cm}}
    \toprule
    \textbf{RP} & \textbf{Illustration} & \textbf{Definition} & \textbf{Example Question} & \textbf{S-expression} \\
    \midrule
    T-0 & \raisebox{-\height}{\includegraphics[width=2.0cm]{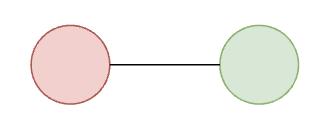}} 
    & A single-hop path from the constraint to the answer.
    & What is the name of money in Brazil? 
    & (JOIN (R location.country.currency\_used) m.015fr) \\
    \midrule
    T-1 & \raisebox{-\height}{\includegraphics[width=2.8cm]{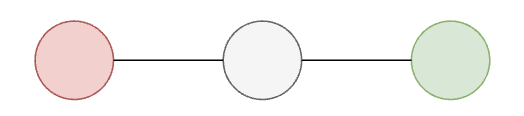}} 
    & A two-hop path from the constraint to the answer.
    & Where does the Queen of Denmark live? 
    & (JOIN (R people.place\_lived.location) (JOIN (R people.person.places\_lived) m.0g2kv)) \\
    \midrule
    T-2 & \raisebox{-\height}{\includegraphics[width=2.8cm]{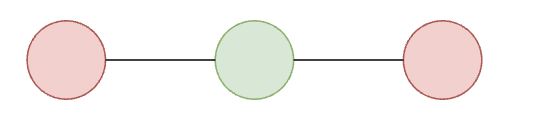}} 
    & Two single-hop paths arising from two different constraints and converging to the same answer. 
    & What was Elie Wiesel's father's name? 
    &  (AND (JOIN people.person.gender m.05zppz) (JOIN (R people.person.parents) m.02vsp)) \\
    \midrule
    T-3 & \raisebox{-\height}{\includegraphics[width=2.8cm]{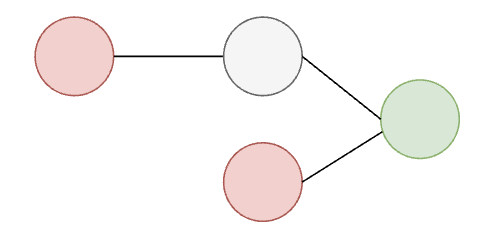}} 
    & Two paths (one single-hop and another two-hop) arising from two different constraints and converging to the same answer. 
    & Where did Joe Namath attend college? 
    &  (AND (JOIN common.topic.notable\_types m.01y2hnl) (JOIN (R education.education.institution) (JOIN (R people.person.education) m.01p\_3k))) \\
    \midrule
    T-4 & \raisebox{-\height}{\includegraphics[width=2.8cm]{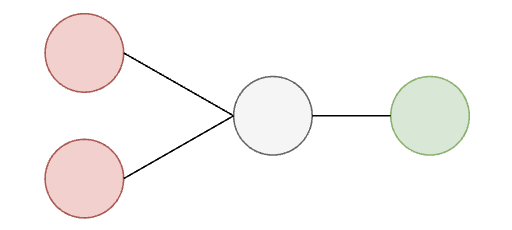}} 
    & Two two-hop paths arising from two different constraints and converging to an intermediate common node before reaching the answer. 
    & Who does Zach Galifianakis play in The Hangover? 
    &  (JOIN (R film.performance.character) (AND (JOIN film.performance.film m.0n3xxpd) (JOIN (R film.actor.film) m.02\_0d2))) \\
    \bottomrule
    \end{tabular}
    }
    \caption{Reasoning patterns with their corresponding definitions, example questions, and S-expressions.}
    \label{tab:isomorphism_types}
\end{table*}

\begin{table}[h]
\centering
\resizebox{\linewidth}{!}{
\begin{tabular}{cccccc}
\toprule
RP &  Illustration &  i.i.d. & Comp & Z.S. & Total \\ \midrule

T-0 & \begin{minipage}{.1\textwidth}\includegraphics[scale=0.1]{imgs/iso-0.png}\end{minipage}&  50.3 & 0.0 &  49.7 &  54.5 

\\ \midrule
T-1 & \begin{minipage}{.1\textwidth}\includegraphics[scale=0.09]{imgs/iso-1.png}\end{minipage}&  37.3 & 44.3 & 18.4 & 23.5

\\ \midrule

T-2 & \begin{minipage}{.1\textwidth}\includegraphics[scale=0.09]{imgs/iso-2.png}\end{minipage}& 17.1 & 47.1 & 35.7 & 5.2 
\\ \midrule

T-3 & \begin{minipage}{.1\textwidth}\includegraphics[scale=0.09]{imgs/iso-3.png}\end{minipage}&  83.3 & 6.7 & 10.0 & 2.2

\\ \midrule
T-4 & \begin{minipage}{.1\textwidth}\includegraphics[scale=0.09]{imgs/iso-4.png}\end{minipage}&  12.8 & 81.5 & 5.6 & 14.5 
\\ \midrule

\multicolumn{2}{c}{ALL} & 40.8 & 24.9 & 34.3 & 100.0
\\

\bottomrule
\end{tabular}}
\caption{Distribution of reasoning patterns over the generalization splits (i.i.d., compositional (Comp), zero-shot (Z.S.)) of our modified WebQSP dataset.}

\label{tab:isomorphisms-distribution}
\end{table}

\begin{figure*}
    \centering
    \includegraphics[width=1.0\linewidth]{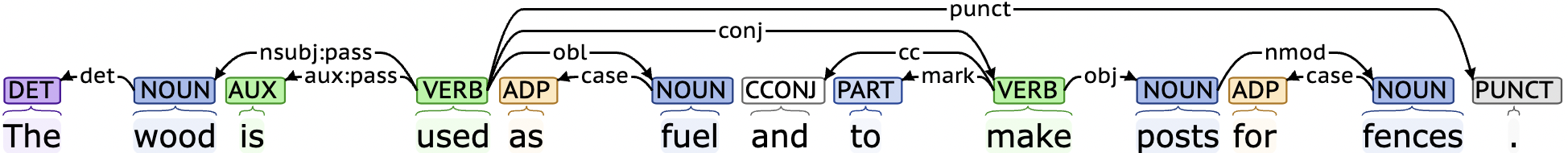}
    \caption{Example depicting the supplemental information provided by the \textit{dependency tree}. The entities of interest are \textcolor{red}{wood} and \textcolor{red}{fences}, having the relationship \textcolor{green}{material\_used}. The path \textit{wood} $\leftarrow$ \textit{used} $\rightarrow$ \textit{make} $\rightarrow$ \textit{posts} $\rightarrow$ \textit{fences} elicits this relationship.}
    \label{fig:dependency}
\end{figure*}

\begin{table*}[]
\centering
\small
\begin{tabular}{@{}lp{2.0cm}p{2.3cm}p{2.3cm}p{2cm}@{}}
\toprule
\textbf{Task} & \textbf{Text model} & \textbf{Graph model} & \textbf{Loss function} & \textbf{Metric} \\
\midrule
ETRE & RoBERTa\textsuperscript{1} & \{1,2,3\}-layer RGAT\textsuperscript{4} & cross-entropy (CE) & weighted F1 \\
Form understanding & RoBERTa\textsuperscript{1} & 2-layer RGAT\textsuperscript{4} & binary CE & F1 \\
MLRE & mBERT-base\textsuperscript{2} & 2-layer RGCN \textsuperscript{5} & CE & macro F1 \\
Reasoning pattern prediction & T5-base\textsuperscript{3} & 2-layer RGCN\textsuperscript{5} & CE & macro F1 \\
KBQA entity-ranking & T5-base\textsuperscript{3} & 2-layer RGCN\textsuperscript{5} & binary CE & Hits@K\textsuperscript{6} \\
\bottomrule
\end{tabular}
\caption{Model configurations, training objectives, and evaluation metrics for each task. The text and graph model backbones listed in this table are used for the primary results in Table~\ref{tab:task_performance}.}
\label{tab:task_exp_details}
\end{table*}

\begin{table*}[t]
\centering
\small
\begin{tabular}{lccccccc}
\toprule
\textbf{Task} & \makecell{\textbf{LR}} & \makecell{\textbf{Batch} \\ \textbf{size}} & \makecell{\textbf{Drop} \\ \textbf{out}} & \makecell{\textbf{Temp.}} & \makecell{\textbf{Max} \\ \textbf{input len}} & \makecell{\textbf{GNN} \\ \textbf{layers}} & \makecell{\textbf{GNN} \\ \textbf{hidden dim}} \\
\midrule
ETRE (TDDMan)        & 1e-5  & 16 & 0.1 & 0.1  & --    & 2 & 256 \\
ETRE (TDDAuto)       & 1e-5  & 32 & 0.1 & 0.04 & --    & 3 & 256 \\
ETRE (TB-Dense)      & 1e-5  & 32 & 0.1 & 0.9  & --    & 1 & 256 \\
MLRE                 & 1e-5  & 16 & 0.2 & 0.1  & 512   & 2 & 768 \\
Reasoning pattern prediction    & 5e-5  & 6  & 0.2 & 0.1  & 512   & 2 & 768 \\
KBQA entity-ranking    & 5e-5  & 4  & 0.2 & 0.1  & 1024  & 2 & 768 \\
Form understanding   & \multicolumn{7}{c}{\textit{Same settings as in}~\citet{nourbakhsh-etal-2024-aligatr}} \\
\bottomrule
\end{tabular}
% \vspace{4pt}
\caption{Hyperparameters used across tasks. Temperature refers to $\tau$ in CoD. All experiments use a shared space dimension of 2048.}
\label{tab:hyperparams}
\end{table*}

\begin{table*}[t]
\centering
\small
\renewcommand{\arraystretch}{1.2}
\begin{tabularx}{\textwidth}{
  @{} l l r r X X @{}
}
\toprule
\textbf{Task} & \textbf{Dataset} & \textbf{Train} & \textbf{Test} & \textbf{Number of labels} & \textbf{Training time} \\
\midrule
\multirow{3}{*}{\textbf{ETRE}} 
  & TDDMan   & 4,000   & 1,500   & 5  & 28 min  \\
  & TDDAuto  & 32,609  & 4,258   & 5  & 3h 40min \\
  & TB-Dense & 4,032   & 1,427   & 6  & 26 min  \\
\midrule
\multirow{5}{*}{\textbf{MLRE}} 
  & REDFM (en) & 8,504   & 1,235   & 32 & 6h 7min  \\
  & REDFM (es) & 5,194   &   733   & 32 & 2h 30min \\
  & REDFM (fr) & 5,452   &   975   & 32 & 3h 14min \\
  & REDFM (de) & 5,909   &   811   & 32 & 2h 46min \\
  & REDFM (it) & 4,597   & 1,086   & 32 & 2h 38min \\
\midrule
\textbf{Reasoning pattern prediction} & WebQSP       & 3,014     & 1,343     & 5  & 1h  \\
\textbf{KBQA answer-ranking}    & WebQSP  & 3,014 & 1,343 & Number of gold answers & 3h  \\
\midrule
\multirow{3}{*}{\textbf{Form understanding}} 
  & SROIE   & 626   & 347   & 4  & 10h       \\
  & FUNSD   & 149   & 50    & 4  & 4h 36min  \\
  & CORD    & 800   & 100   & 30 & 17h 47min \\
\bottomrule
\end{tabularx}
\caption{Task suite statistics and training times. We train for 1000 epochs for form understanding.}
\label{tab:task_stats}
\end{table*}

\section{Task suite details}
\subsection{Data processing for reasoning pattern prediction and KBQA entity-ranking} \label{data_processing}

We use the WebQSP dataset~\citep{WebQSP} for our two KBQA related experiments, i.e. reasoning pattern prediction and entity-ranking. An exploratory analysis of WebQSP highlighted a significant overlap of relations and classes across the train and test splits. Subsequently, we employed the approach of~\citet{jiang2022knowledge} to obtain development and test splits that characterize different generalization levels in equal proportion. The three generalization levels for KBQA tasks include  i.i.d, compositional, and zero-shot.

The i.i.d. case implies that the questions observed during inference follow similar logical templates to those during training; for example the questions ``Who was the author of Oliver Twist?'' and  ``Who wrote Pride and Prejudice?'' follow similar logical templates. We contrast this with the compositional case, where questions in the test split operate over the same set of relations that were present in the training set (such as the ``written-by'' relation), but different logical templates. For example, the questions ``Who wrote Pride and Prejudice?'' and ``Who wrote both The Talisman and It?'' require reasoning over the same relation ``written-by'' but follows different reasoning paths, since the former involves only one constraint or entity, whereas the latter involves two. Finally, questions in the zero-shot split operate over new or unseen relations that were not present in the training dataset. For example, the questions ``Who wrote Pride and Prejudice?'' and ``Who directed Pride and Prejudice in 2005?'' involves different relations, i.e. ``written-by'' and ``directed-by'' respectively. We defer the readers to past work \citep{gu2021beyond, jiang2022knowledge, grailqapp} for a more thorough description of the different generalization splits. 

% We leverage the idea of isomorphism from \citet{grailqapp} to characterize the complexity of the reasoning path required to be traversed to answer a given KBQA question.

We characterize the complexity of the reasoning pattern to answer a given KBQA question based on \citet{grailqapp}.
Given the modified version of WebQSP dataset, we identify the following five reasoning patterns that accounted for $\geq$ 97\% of the dataset across all splits. We describe the different reasoning patterns in Table \ref{tab:isomorphism_types} and outline their distribution in the our modified WebQSP dataset in Table \ref{tab:isomorphisms-distribution}. 

To accommodate the input length constraints of models like T5, we simplify the representation of knowledge base entities in the linearized graph input. Instead of using full entity identifiers (e.g., \texttt{m.02896}), we assign short, unique placeholder tokens (e.g., \texttt{<E1>}, \texttt{<E2>}) to each entity as a part of the tokenizer vocabulary. This helps reduce the input sequence length and avoids unwanted subword tokenization. In addition, we ensure that these placeholder tokens are assigned consistently across modalities: the same entity is represented as node $v_i$ in the graph and as token \texttt{<Ei>} in the linearized text.

\begin{table*}[]
\centering
% \small

% ---------- Subtable 1 ----------
\begin{tabular}{@{}llccc@{}}
\toprule
\multicolumn{5}{c}{\textbf{(a) Reasoning pattern prediction}} \\
\midrule
Text encoder & Graph encoder & Hybrid (CoD) & Text only & Graph only \\
\midrule
T5     & RGCN & \textbf{0.6190} & 0.5700 & 0.5840 \\
T5     & RGAT & \textbf{0.6120} & 0.5700 & 0.4966 \\
BERT   & RGCN & \textbf{0.5999} & 0.5835 & 0.5840 \\
BERT   & RGAT & \textbf{0.5956} & 0.5835 & 0.4966 \\
GPT-2  & RGCN & \textbf{0.6022} & 0.5614 & 0.5840 \\
GPT-2  & RGAT & \textbf{0.6049} & 0.5614 & 0.4966 \\
\bottomrule
\end{tabular}

\vspace{0.8em}

% ---------- Subtable 2 ----------
\begin{threeparttable}
\begin{tabular}{@{}llcc@{}}
\toprule
\multicolumn{4}{c}{\textbf{(b) Event temporal relation extraction (ETRE)}} \\
\midrule
Text encoder & Graph encoder & Hybrid (CoD) & Text only \\
\midrule
\multicolumn{4}{l}{\textit{TDDMan}} \\
BERT     & GCN   & 0.411 & \textbf{0.447} \\
BERT     & RGCN  & 0.384 & \textbf{0.447} \\
BERT     & RGAT  & \textbf{0.481} & 0.447 \\
RoBERTa  & GCN   & 0.435 & \textbf{0.445} \\
RoBERTa  & RGCN  & \textbf{0.452} & 0.445 \\
RoBERTa  & RGAT  & \textbf{0.551} & 0.445 \\
\midrule
\multicolumn{4}{l}{\textit{TDDAuto}} \\
BERT     & GCN   & \textbf{0.631} & 0.624 \\
BERT     & RGCN  & \textbf{0.647} & 0.624 \\
BERT     & RGAT  & \textbf{0.683} & 0.624 \\
RoBERTa  & GCN   & \textbf{0.748} & 0.689 \\
RoBERTa  & RGCN  & 0.665 & \textbf{0.689} \\
RoBERTa  & RGAT  & \textbf{0.771} & 0.689 \\
\midrule
\multicolumn{4}{l}{\textit{TB-Dense}} \\
BERT     & GCN   & \textbf{0.790} & 0.775 \\
BERT     & RGCN  & \textbf{0.782} & 0.775 \\
BERT     & RGAT  & \textbf{0.810} & 0.775 \\
RoBERTa  & GCN   & \textbf{0.805} & 0.767 \\
RoBERTa  & RGCN  & \textbf{0.847} & 0.767 \\
RoBERTa  & RGAT  & \textbf{0.856} & 0.767 \\
\bottomrule
\end{tabular}

\begin{tablenotes}
\footnotesize
\item Note that we did not record numbers for the graph-only approach because the graph approach for this task yields incredibly poor results without the incorporation of linear transformers~\citep{yao2024distilling}.
\end{tablenotes}
\end{threeparttable}

\caption{
Additional results for (a) Reasoning pattern prediction and (b) ETRE using different text and graph encoder backbones. CoD consistently improves over baselines across all combinations in Reasoning pattern prediction, and improves 78\% of the times across all 18 cases for ETRE. These results demonstrate CoD’s generality across diverse model architecture combinations. 
}
\label{tab:additional_backbones}
\end{table*}

\subsection{MLRE dependency parsing illustration} \label{appendix:dependency}
See Figure~\ref{fig:dependency}.

\subsection{FU example}
\begin{figure*}[t]
    \centering
    \includegraphics[width=\textwidth]{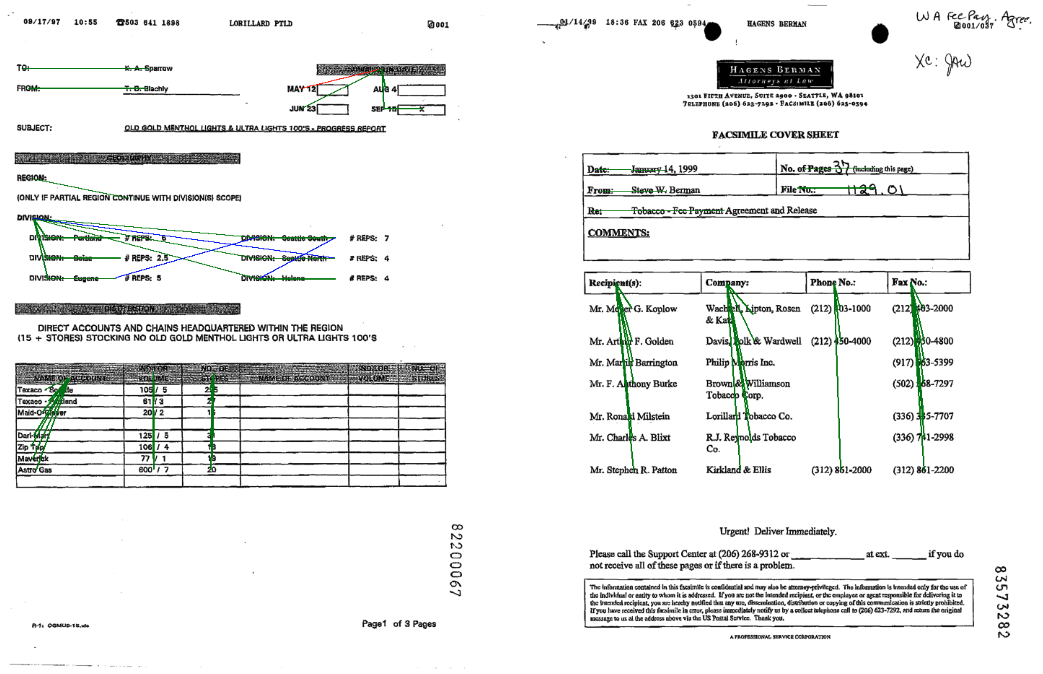}
    \caption{An example of FU task from the FUNSD dataset, adapted from~\citet{nourbakhsh-etal-2024-aligatr}. Green links show correct predictions. Red links show false negatives. Blue links show false positives.}
    \label{fig:fu_example}
\end{figure*}

We adapt an example to showcase the FU task from~\citet{nourbakhsh-etal-2024-aligatr} in Figure~\ref{fig:fu_example}.

\section{Task experiments details} \label{experiment_config}

We present the experimental details for different tasks. In Table~\ref{tab:task_exp_details}, we outline the loss function that we are optimizing, the corresponding evaluation metric, and the backbone architectures used for the primary results reported in Table \ref{tab:task_performance}: the transformer model that encodes the textual information, and the specific GNN architecture that encodes the graph information. In Table~\ref{tab:hyperparams}, we provide hyperparameters values for our experiments. We also present statistics on the task suite datasets and training times in Table~\ref{tab:task_stats}. All datasets we used are publicly available, and we follow the licensing terms and intended use of each.

\footnotetext[1]{\citet{liu2019robertarobustlyoptimizedbert}}
\footnotetext[2]{\citet{devlin2019bertpretrainingdeepbidirectional}}
\footnotetext[3]{\citet{raffel2023exploringlimitstransferlearning}}
\footnotetext[4]{\citet{busbridge2019relationalgraphattentionnetworks}}
\footnotetext[5]{\citet{schlichtkrull2017modelingrelationaldatagraph}}
\footnotetext[6]{K indicates the number of correct answers for an instance.}

\begin{table*}[]
    \centering
    \begin{tabular}{ccccc}
    \toprule
    Language & Text only & Graph only & Hybrid + CoD  & Hybrid + no-CoD \\
    \midrule
    
    % de& 80.66 $\pm$ 0.50&76.74 $\pm$ 0.53& 79.53 $\pm$ 0.63 & 80.00 $\pm$ 1.35 \\ 
    % en& 84.84 $\pm$ 2.47&82.41 $\pm$ 0.39& 83.68 $\pm$ 0.15 & 84.70 $\pm$ 0.39\\ 
    % es& 79.37 $\pm$ 1.40&75.33 $\pm$ 1.23& 78.73 $\pm$ 0.33 & 80.65 $\pm$ 0.60 \\ 
    % fr& 77.19 $\pm$ 1.78&73.66 $\pm$ 1.24& 77.92 $\pm$ 0.04 & 77.74 $\pm$ 0.70\\ 
    % it& 73.17 $\pm$ 0.61&73.50 $\pm$ 0.22& 75.79 $\pm$ 0.71 & 76.89 $\pm$ 0.44\\ 
    % \midrule
    % Avg& 79.05 $\pm$ 4.15& 76.33 $\pm$ 3.37& 79.13 $\pm$ 2.63 & 80.00 $\pm$ 2.84\\

    de&\textbf{80.41} $\pm$ 0.61&47.13 $\pm$ 2.76&80.35 $\pm$ 0.71&79.55 $\pm$ 0.40\\
    en&\textbf{85.94} $\pm$ 1.41&52.21 $\pm$ 0.56&84.57 $\pm$ 2.25&84.74 $\pm$ 1.07\\
    es&\textbf{80.49} $\pm$ 0.61&51.21 $\pm$ 1.47&76.64 $\pm$ 1.09&80.26 $\pm$ 0.44\\
    fr&77.47 $\pm$ 0.73&45.62 $\pm$ 1.60&\textbf{78.80} $\pm$ 0.58&78.31 $\pm$ 0.78\\
    it&74.25 $\pm$ 0.36&46.61 $\pm$ 1.98&72.67 $\pm$ 1.40&\textbf{74.76} $\pm$ 1.02\\
    \midrule
    Avg&\textbf{79.71} $\pm$ 3.95&48.55 $\pm$ 3.21&78.61 $\pm$ 4.17&79.53 $\pm$ 3.32\\

    \bottomrule
    \end{tabular}
    \caption{F1 score results on MLRE task for the RED\textsuperscript{fm} dataset.}
    \label{tab:f1_multiling}
\end{table*}

\section{Extended CoD results}

To further demonstrate the robustness and generality of CoD, we apply it to new model combinations on two representative tasks: reasoning pattern prediction and ETRE (Table~\ref{tab:additional_backbones}). We also demonstrate additional CoD performance across each language data for MLRE in Table~\ref{tab:f1_multiling}.

\section{Extended visualization results across tasks} \label{appendix:all_results}

\subsection{ETRE results}
\begin{figure*}[hbtp]
  \centering
  
  % Row 1: Three PCA plots
  \begin{subfigure}[t]{0.32\textwidth}
    \centering
    \includegraphics[trim=38 25 70 70, clip, width=\linewidth]{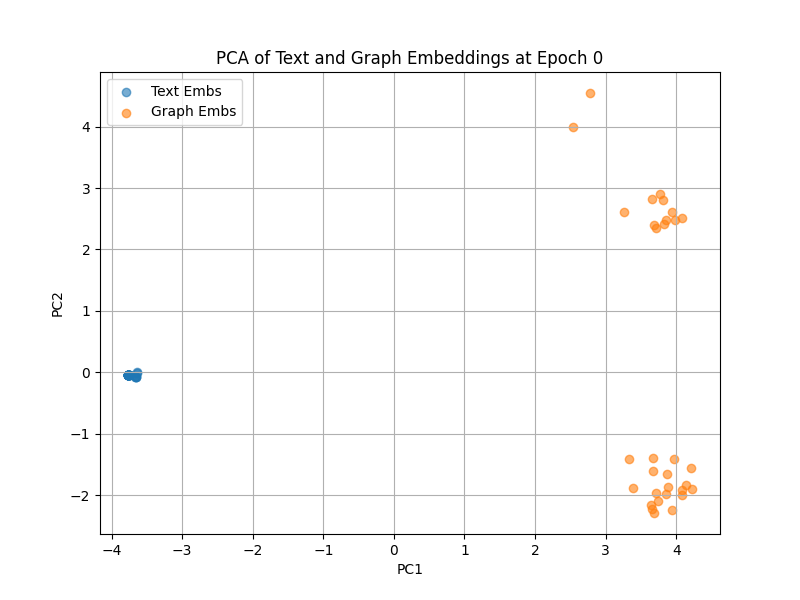}
    \caption{Initial epoch}
  \end{subfigure}
  \hfill
  \begin{subfigure}[t]{0.32\textwidth}
    \centering
    \includegraphics[trim=38 25 70 70, clip, width=\linewidth]{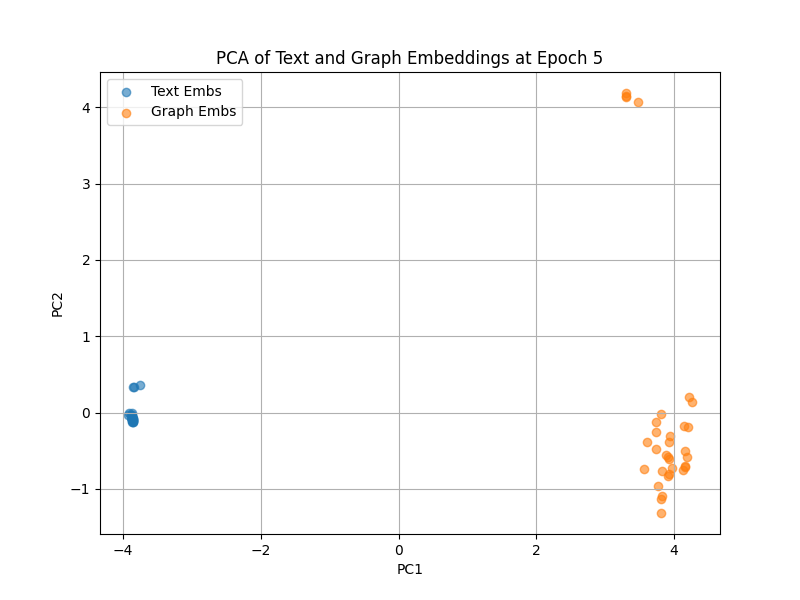}
    \caption{Intermediate epoch}
  \end{subfigure}
  \hfill
  \begin{subfigure}[t]{0.32\textwidth}
    \centering
    \includegraphics[trim=38 25 70 70, clip, width=\linewidth]{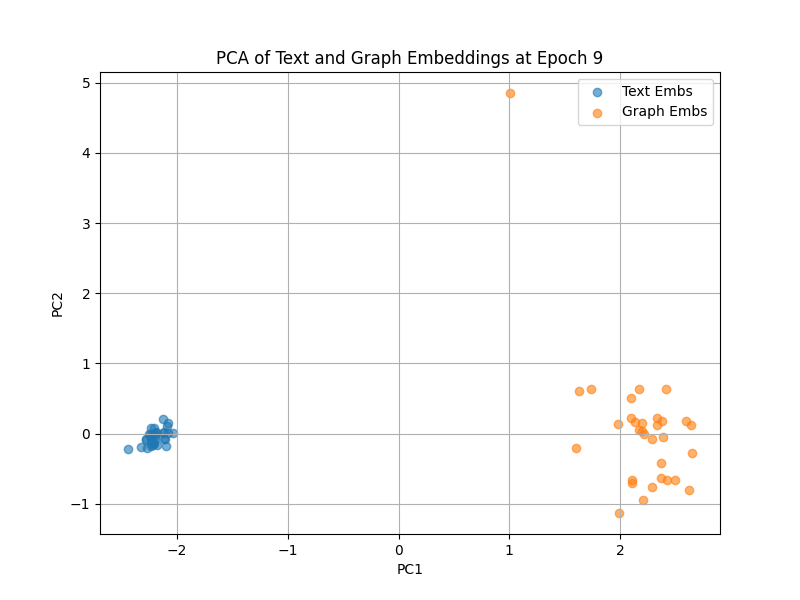}
    \caption{Final epoch}
  \end{subfigure}
  
  % \vspace{1em}
  
% Row 2: Distance-related plots in 2x2 grid, trimmed more aggressively
\begin{subfigure}[t]{0.48\textwidth}
\centering
\includegraphics[trim=0 10 0 160, clip, width=\linewidth]{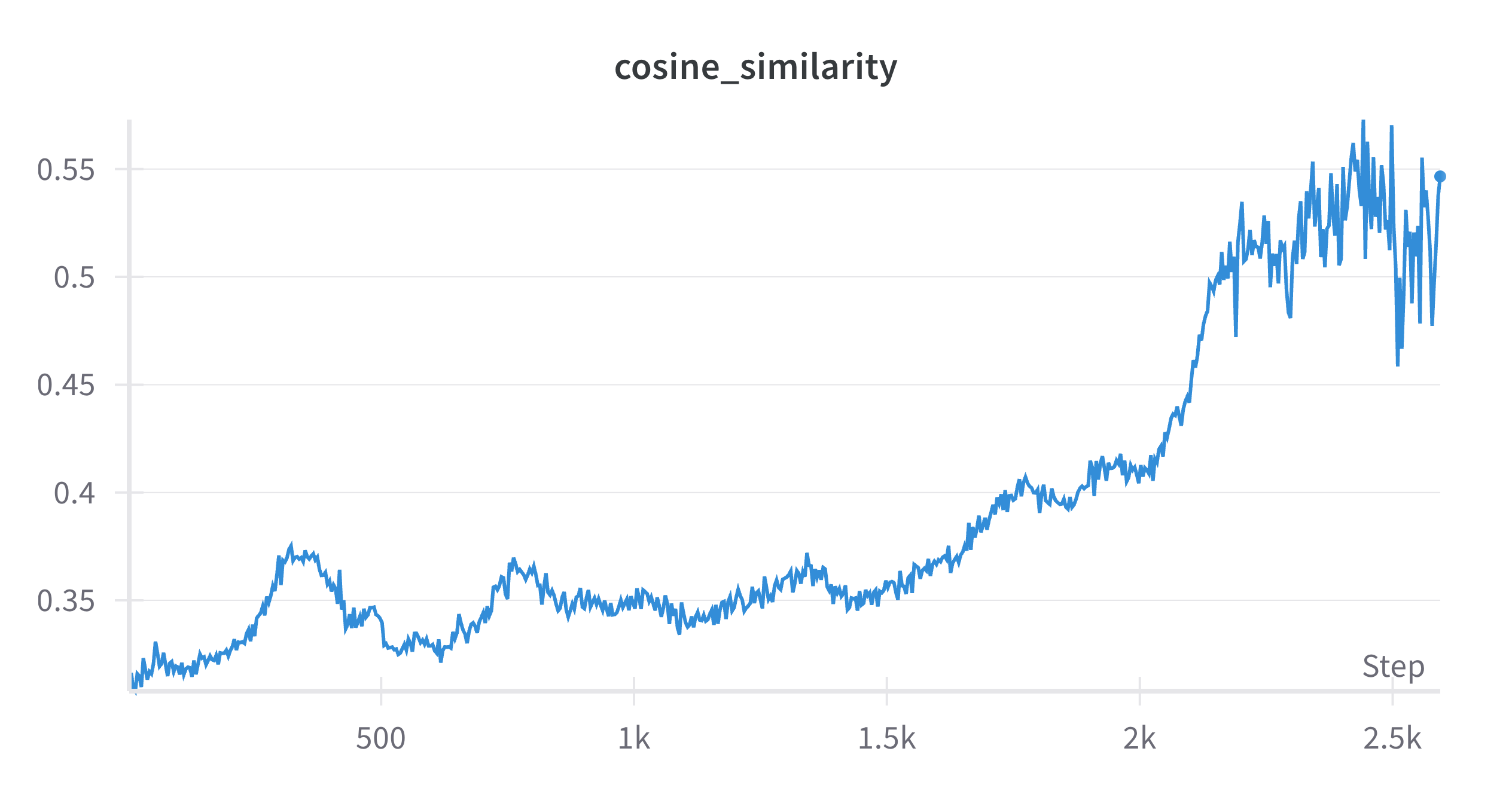}
\caption{Cosine similarity}
\end{subfigure}
\hfill
\begin{subfigure}[t]{0.48\textwidth}
\centering
\includegraphics[trim=0 10 0 160, clip, width=\linewidth]{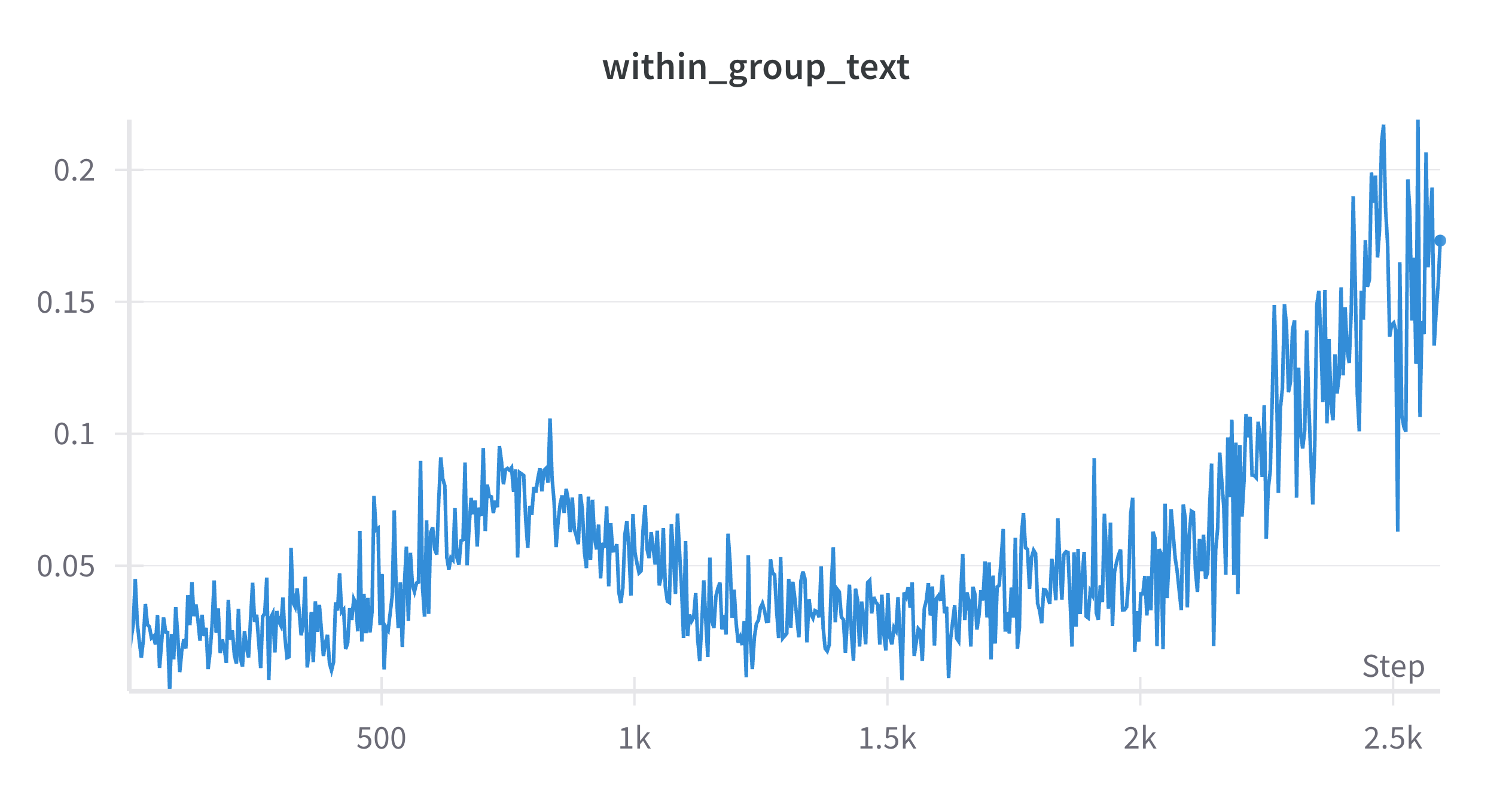}
\caption{Distance within text}
\end{subfigure}
% \vspace{1em}
\begin{subfigure}[t]{0.48\textwidth}
\centering
\includegraphics[trim=0 10 0 160, clip, width=\linewidth]{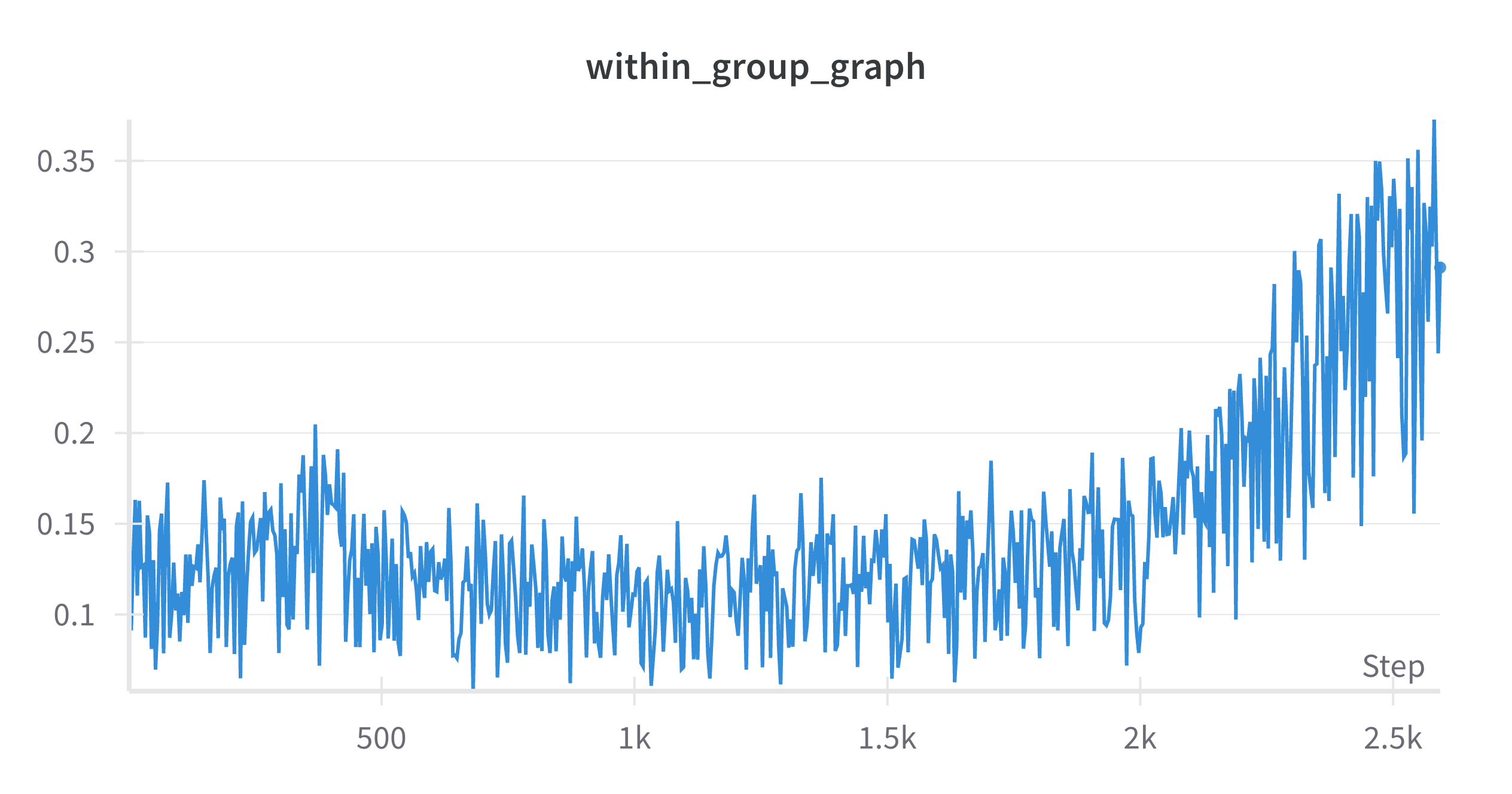}
\caption{Distance within graph}
\end{subfigure}
\hfill
\begin{subfigure}[t]{0.48\textwidth}
\centering
\includegraphics[trim=0 10 0 160, clip, width=\linewidth]{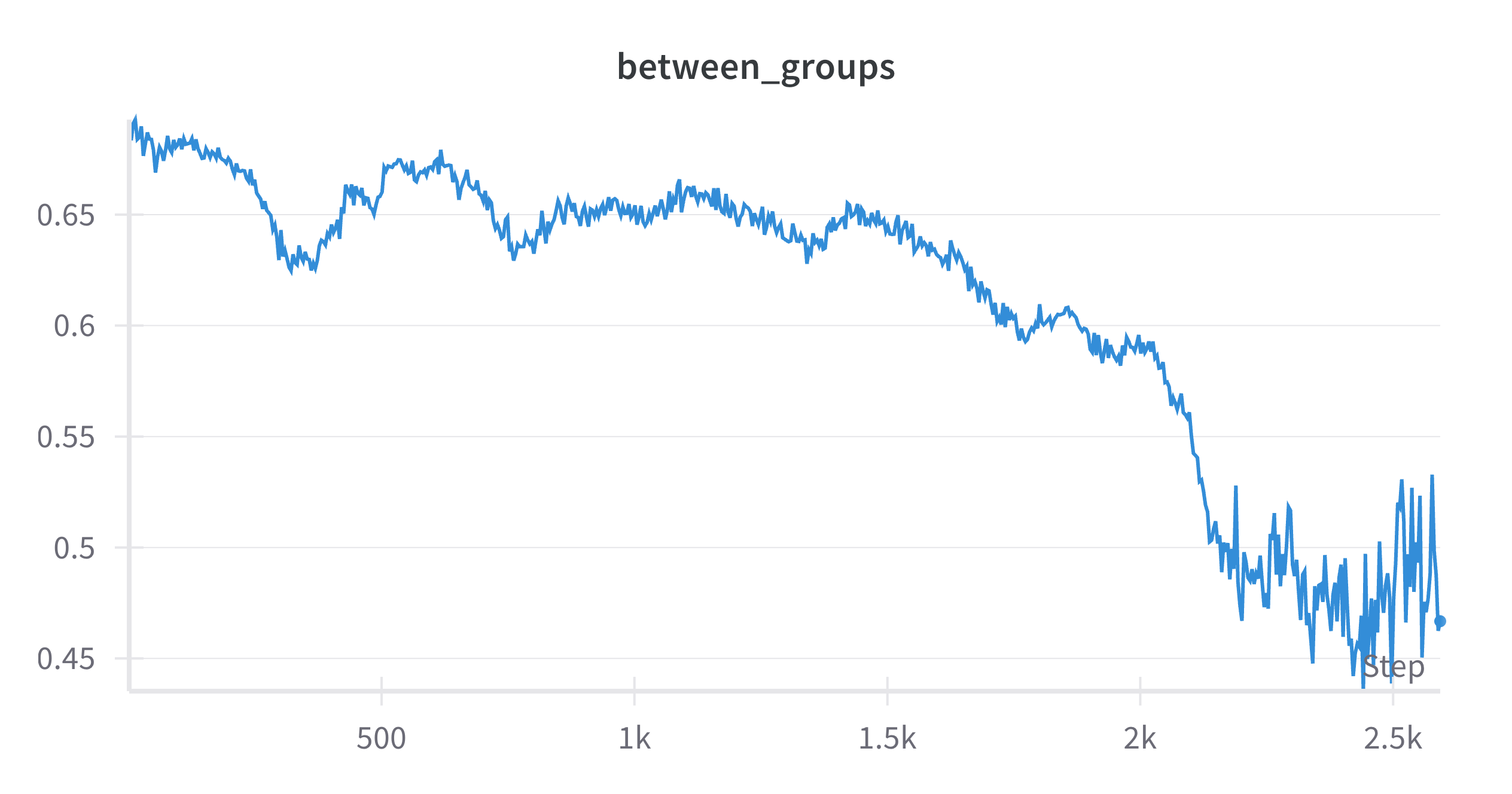}
\caption{Distance between text and graph}
\end{subfigure}
  % \vspace{\baselineskip}
  \caption{
    Results for ETRE on the TimeBank-Dense dataset.}
  \label{fig:etre_tbd_results}
\end{figure*}

\begin{figure*}[hbtp]
  \centering
  
  % Row 1: Three PCA plots
  \begin{subfigure}[t]{0.32\textwidth}
    \centering
    \includegraphics[trim=38 25 70 70, clip,width=\linewidth]{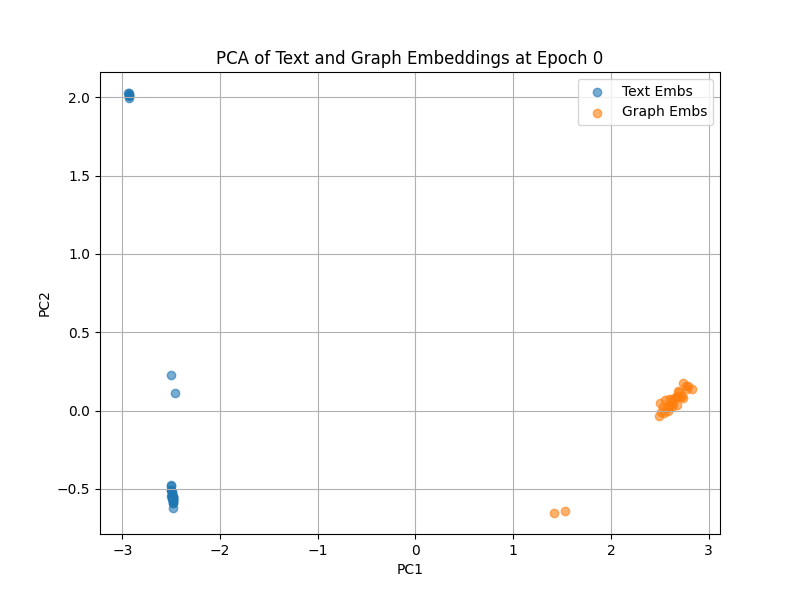}
    \caption{Initial epoch}
  \end{subfigure}
  \hfill
  \begin{subfigure}[t]{0.32\textwidth}
    \centering
    \includegraphics[trim=38 25 70 70, clip,width=\linewidth]{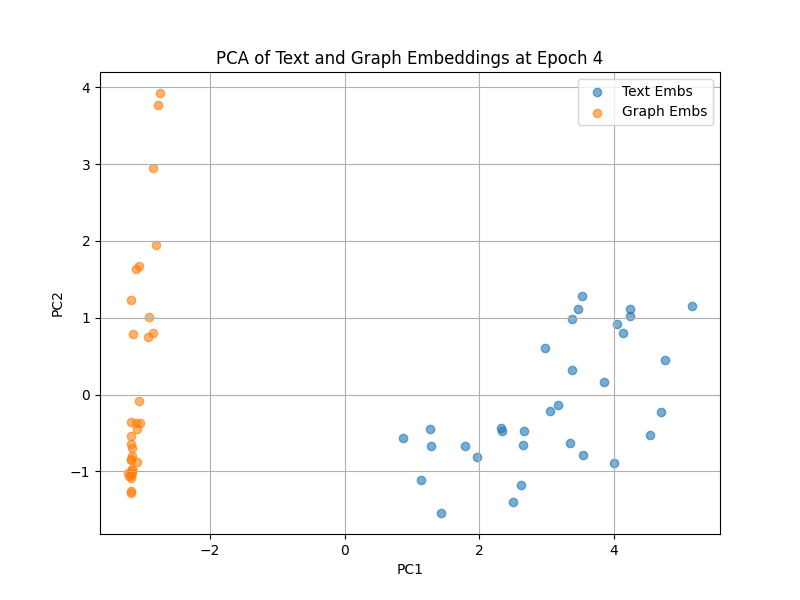}
    \caption{Intermediate epoch}
  \end{subfigure}
  \hfill
  \begin{subfigure}[t]{0.32\textwidth}
    \centering
    \includegraphics[trim=38 25 70 70, clip,width=\linewidth]{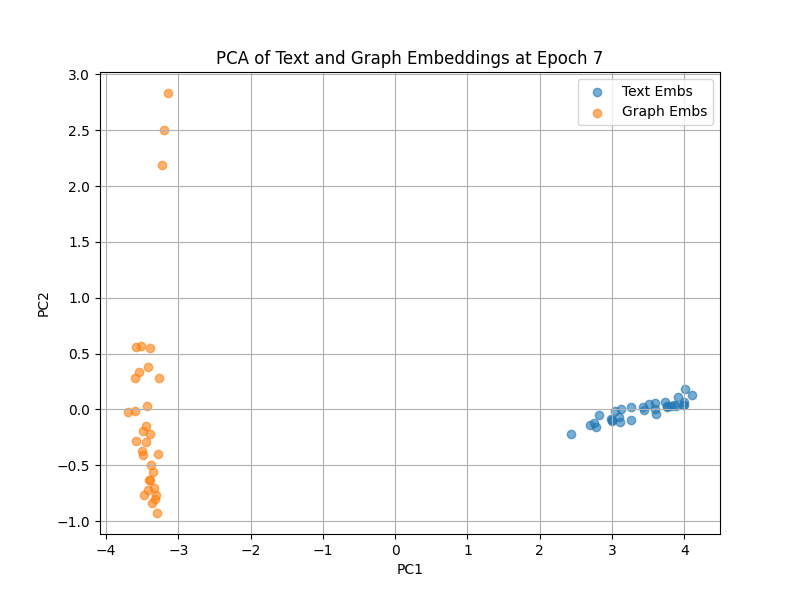}
    \caption{Final epoch}
  \end{subfigure}
  
  % \vspace{1em}
  
  % Row 2: Distance-related plots in 2x2 grid, trimmed more aggressively
\begin{subfigure}[t]{0.48\textwidth}
\centering
\includegraphics[trim=0 10 0 160, clip, width=\linewidth]{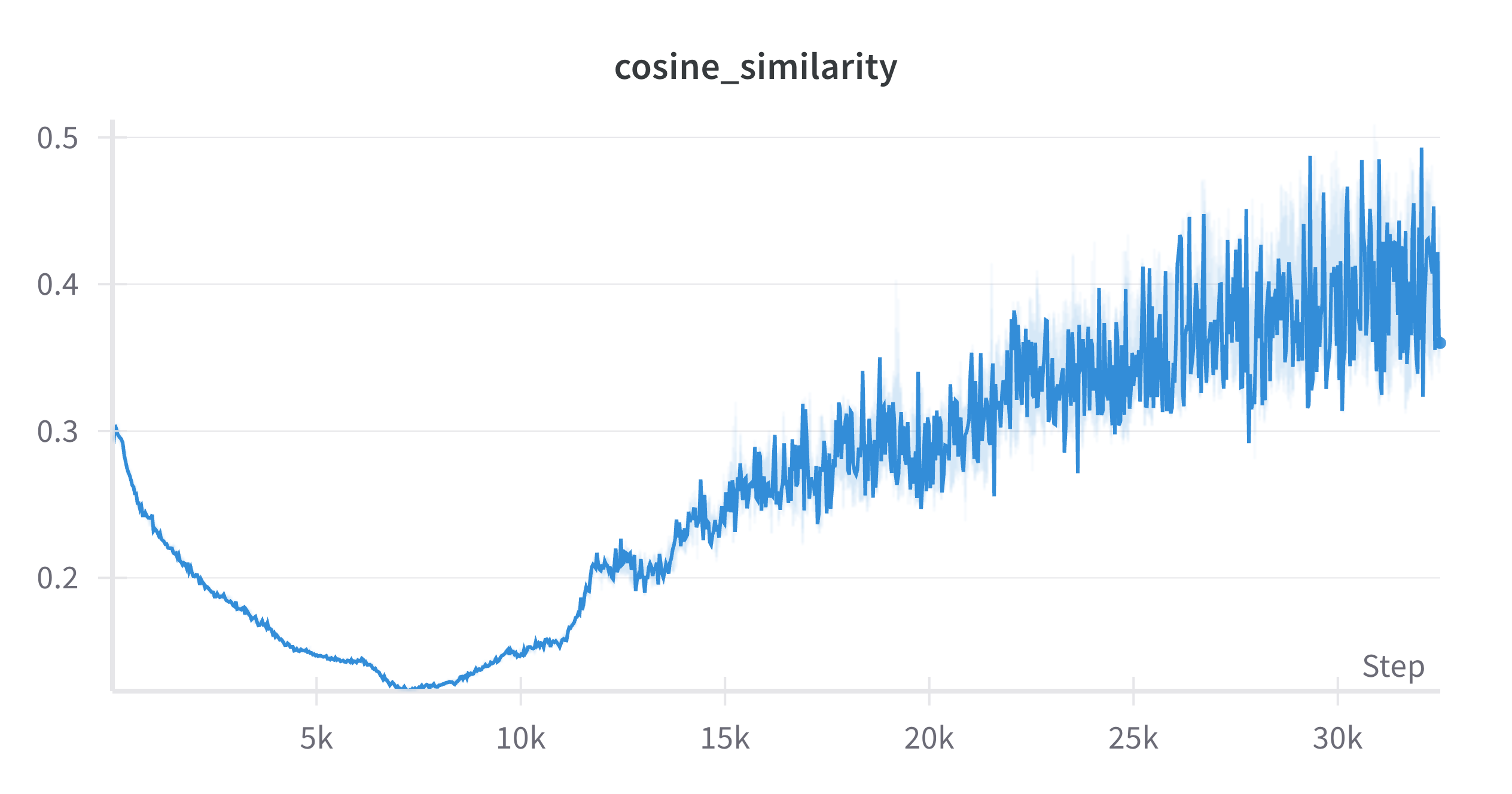}
\caption{Cosine similarity}
\end{subfigure}
\hfill
\begin{subfigure}[t]{0.48\textwidth}
\centering
\includegraphics[trim=0 10 0 160, clip, width=\linewidth]{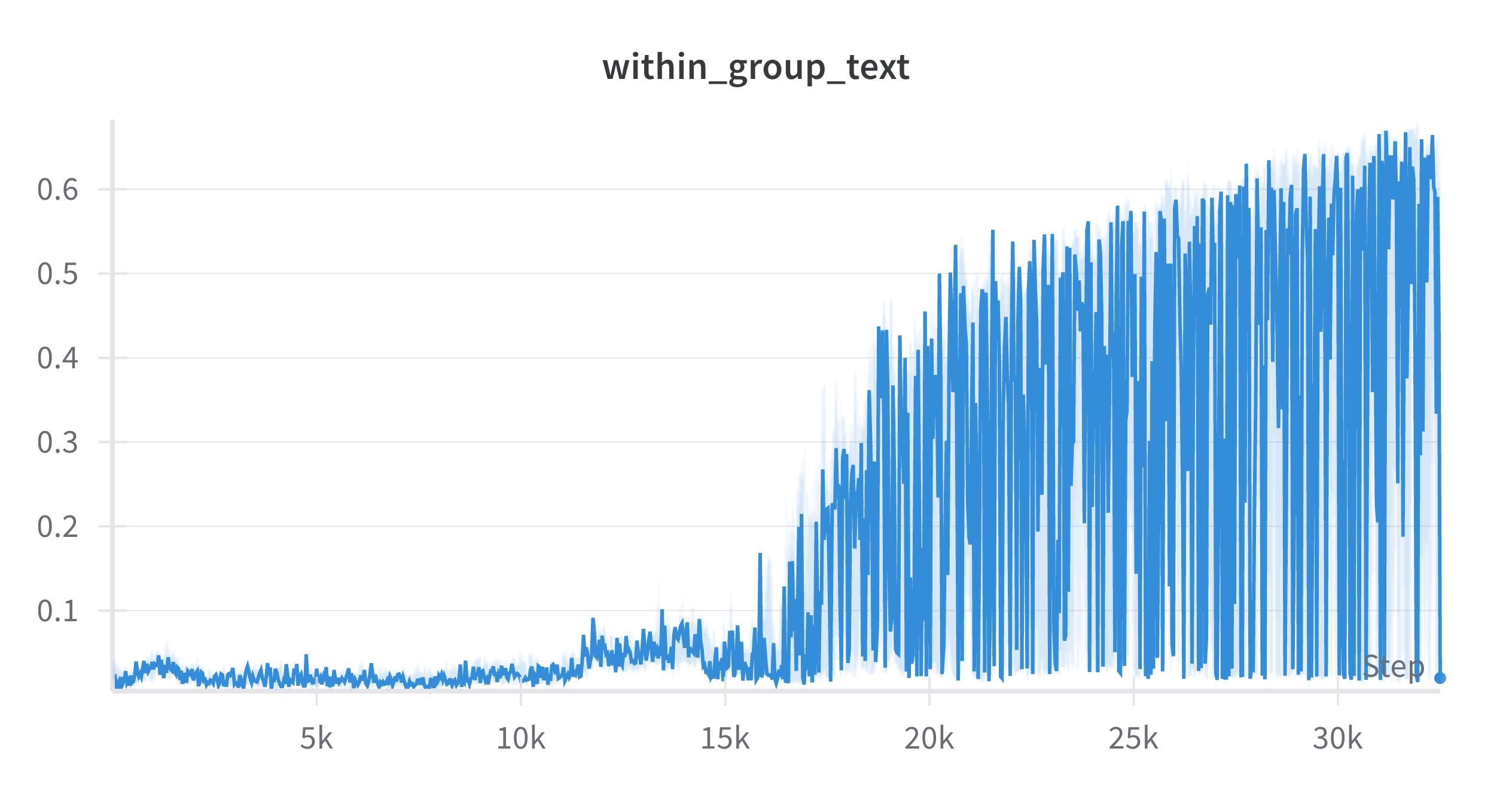}
\caption{Distance within text}
\end{subfigure}
% \vspace{1em}
\begin{subfigure}[t]{0.48\textwidth}
\centering
\includegraphics[trim=0 10 0 160, clip, width=\linewidth]{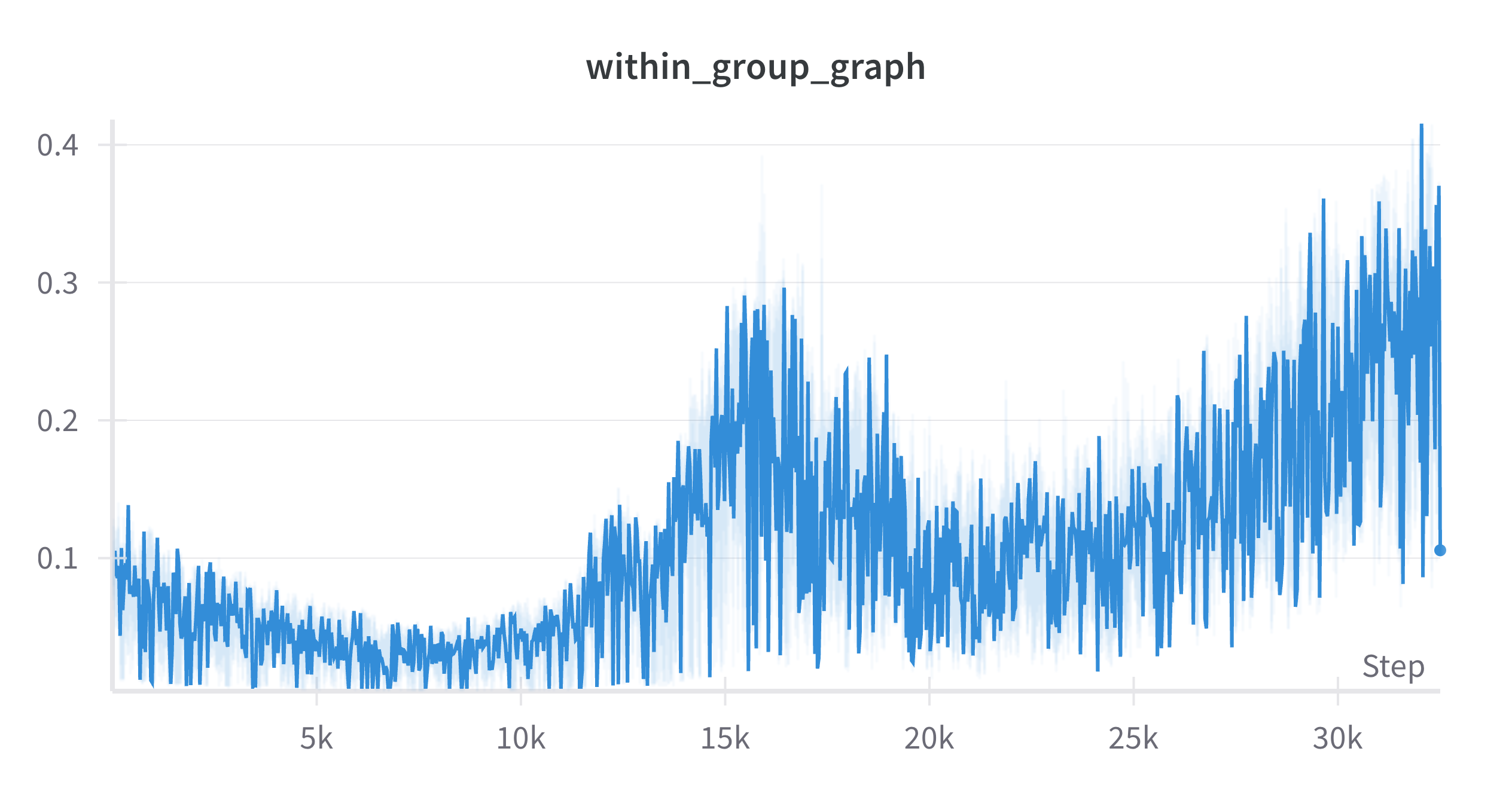}
\caption{Distance within graph}
\end{subfigure}
\hfill
\begin{subfigure}[t]{0.48\textwidth}
\centering
\includegraphics[trim=0 10 0 160, clip, width=\linewidth]{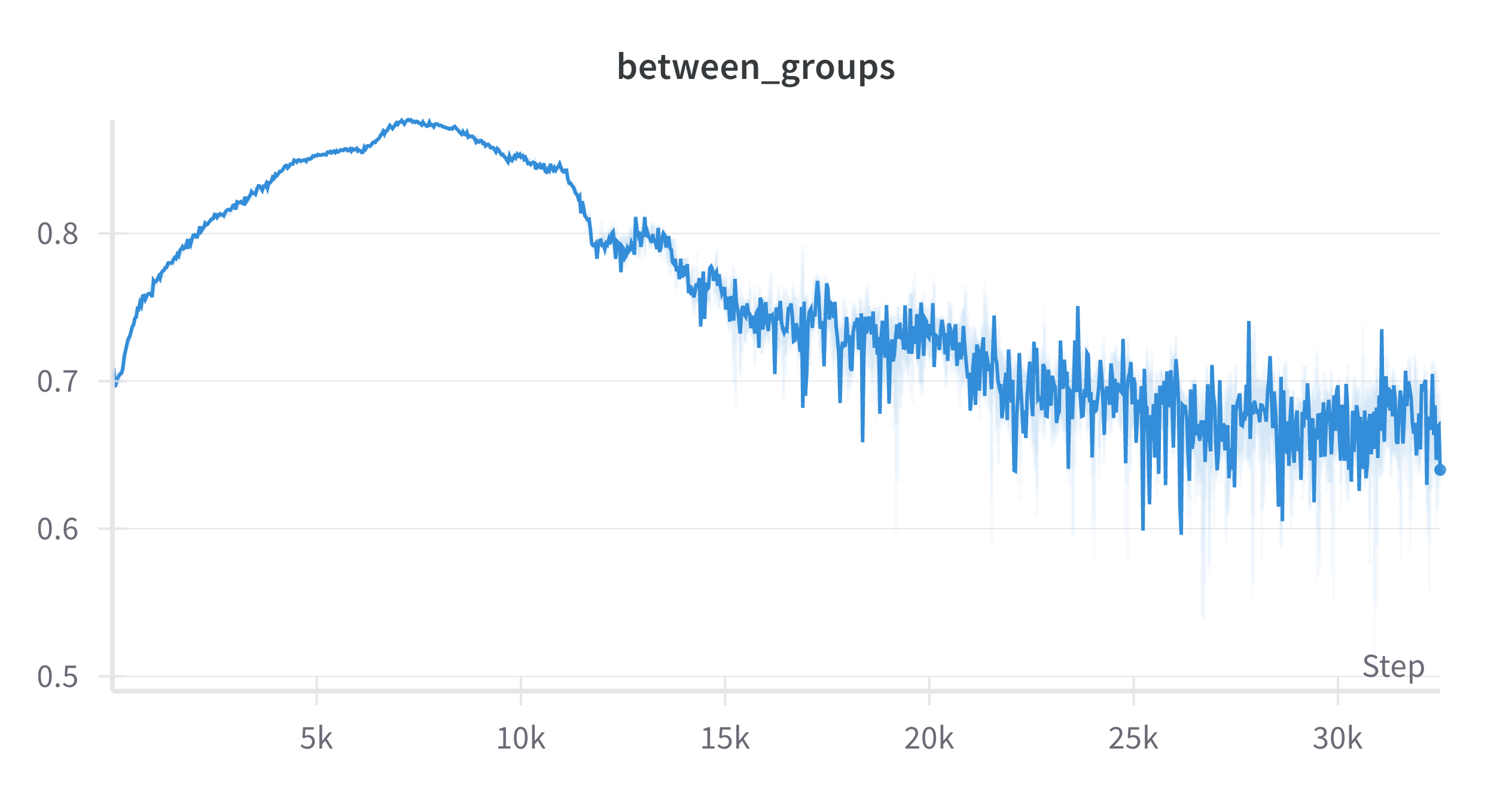}
\caption{Distance between text and graph}
\end{subfigure}
  % \vspace{\baselineskip}
  \caption{
    Results for ETRE on the TDDAuto dataset.}
  \label{fig:etre_tddauto_results}
\end{figure*}

\begin{figure*}[hbtp]
  \centering
  
  % Row 1: Three PCA plots
  \begin{subfigure}[t]{0.32\textwidth}
    \centering
    \includegraphics[trim=38 25 70 70, clip, width=\linewidth]{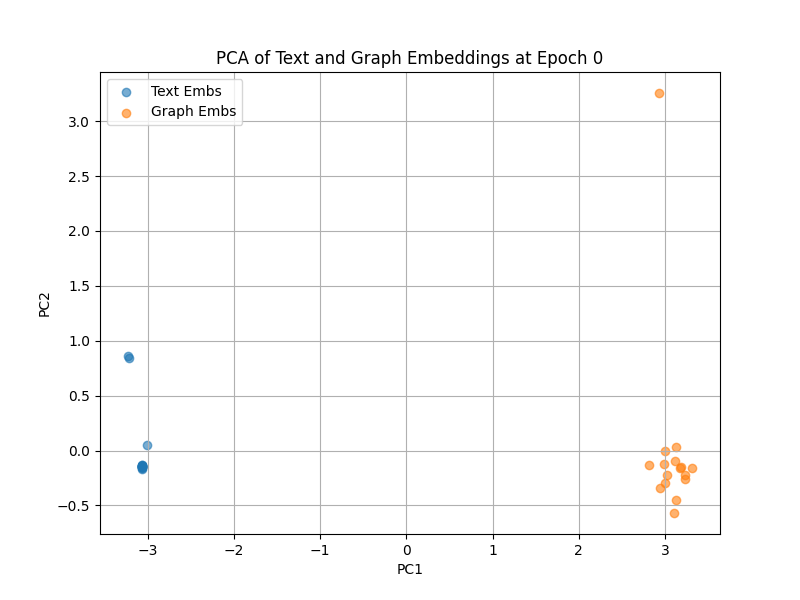}
    \caption{Initial epoch}
  \end{subfigure}
  \hfill
  \begin{subfigure}[t]{0.32\textwidth}
    \centering
    \includegraphics[trim=38 25 70 70, clip, width=\linewidth]{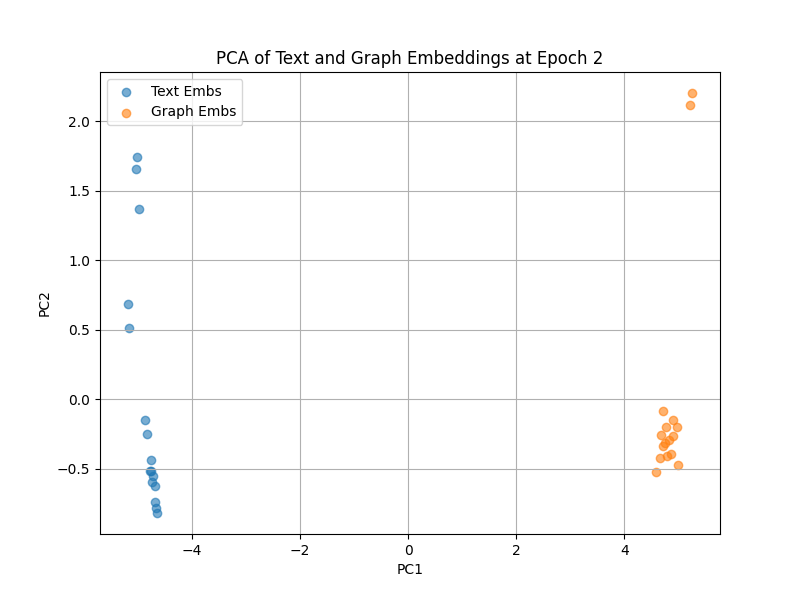}
    \caption{Intermediate epoch}
  \end{subfigure}
  \hfill
  \begin{subfigure}[t]{0.32\textwidth}
    \centering
    \includegraphics[trim=38 25 70 70, clip, width=\linewidth]{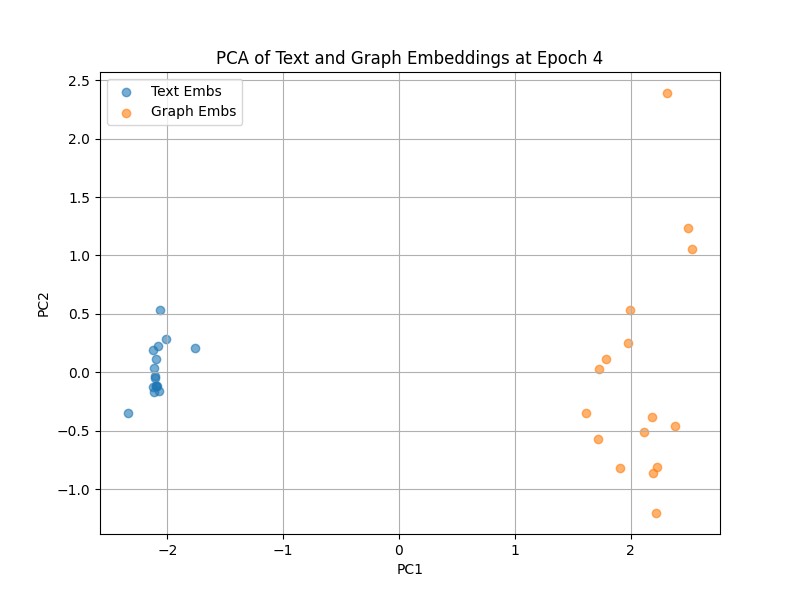}
    \caption{Final epoch}
  \end{subfigure}
  
  % \vspace{1em}
  
% Row 2: Distance-related plots in 2x2 grid, trimmed more aggressively
\begin{subfigure}[t]{0.48\textwidth}
\centering
\includegraphics[trim=0 10 0 160, clip, width=\linewidth]{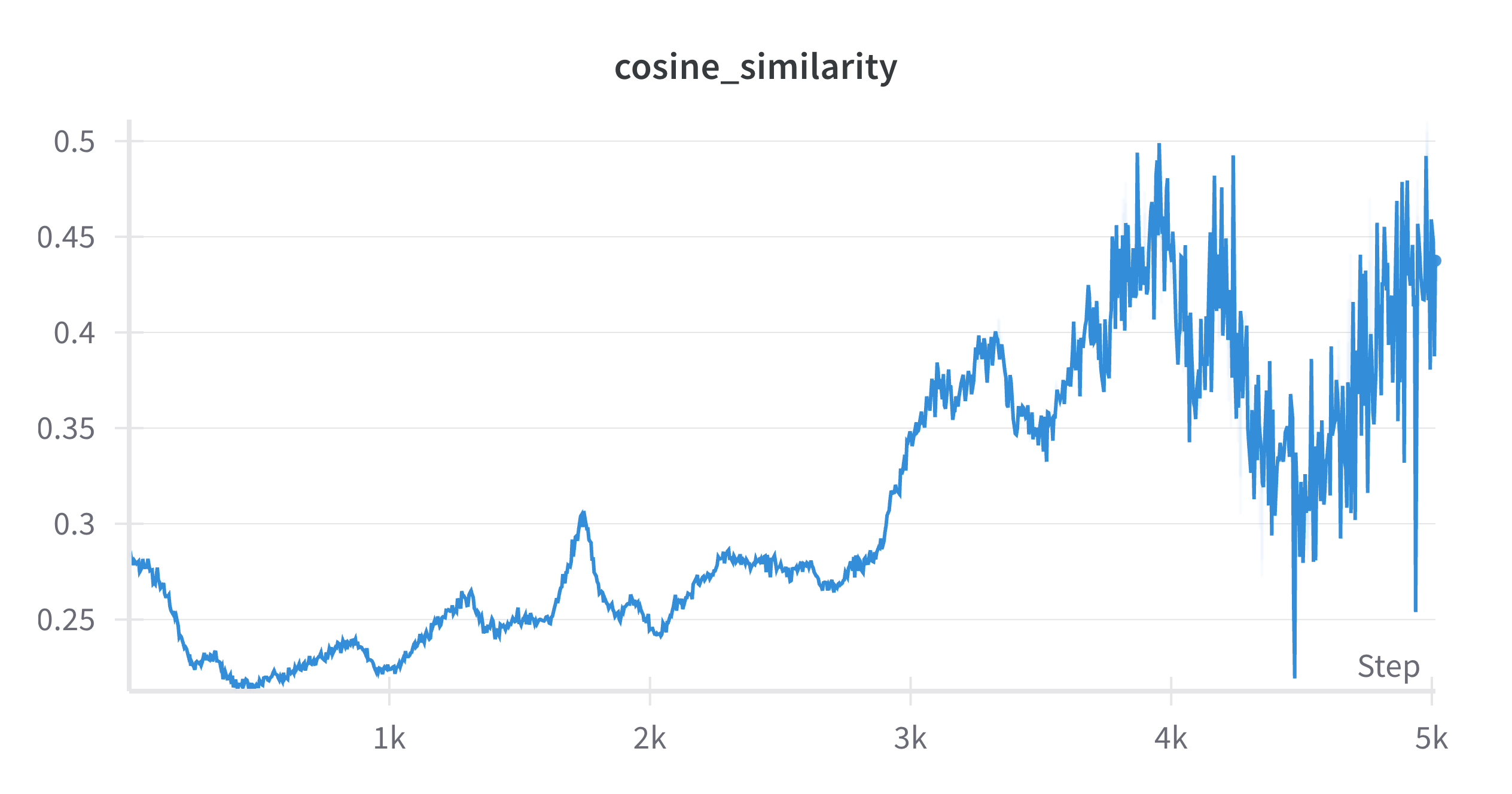}
\caption{Cosine similarity}
\end{subfigure}
\hfill
\begin{subfigure}[t]{0.48\textwidth}
\centering
\includegraphics[trim=0 10 0 160, clip, width=\linewidth]{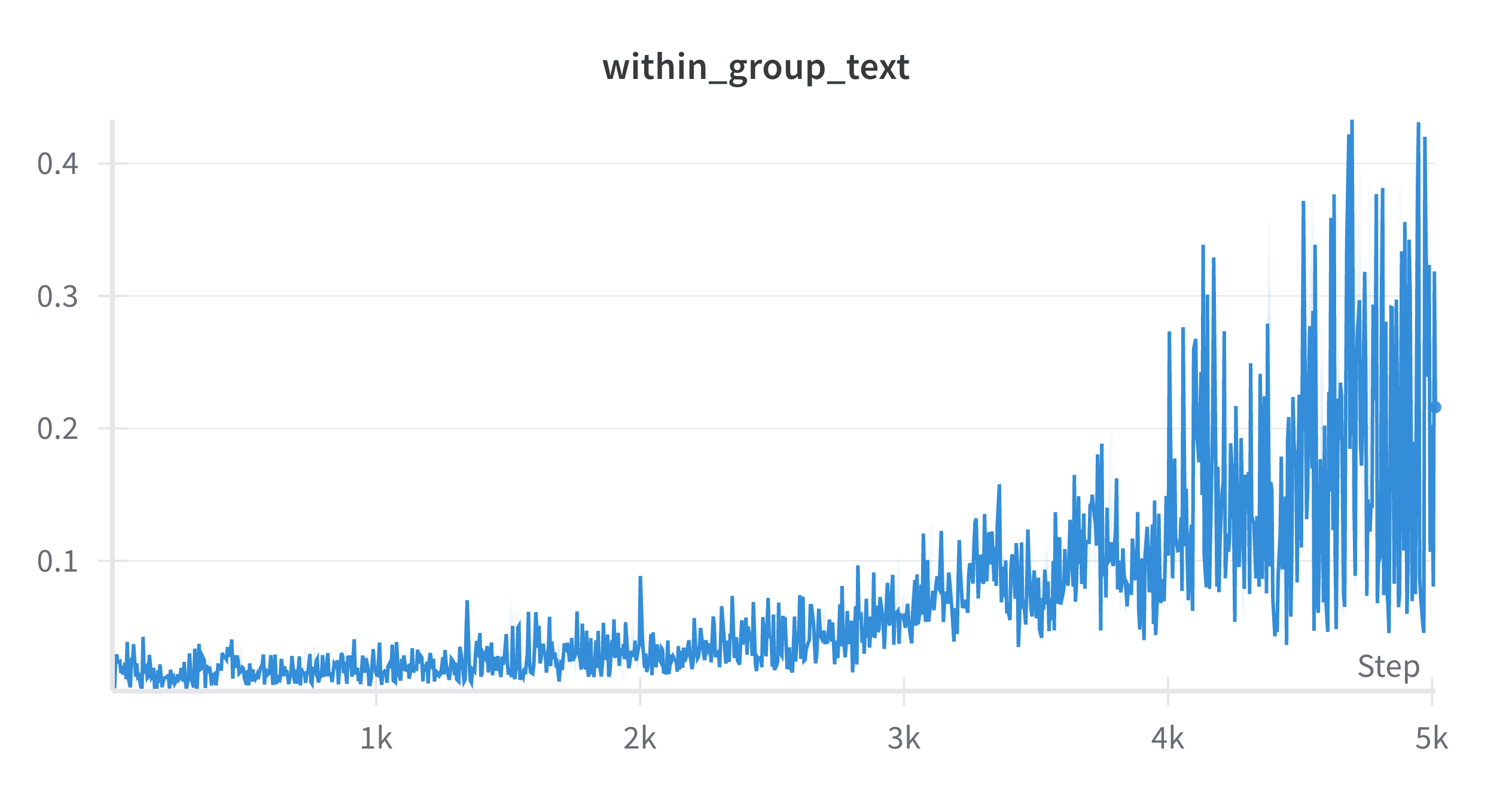}
\caption{Distance within text}
\end{subfigure}
% \vspace{1em}
\begin{subfigure}[t]{0.48\textwidth}
\centering
\includegraphics[trim=0 10 0 160, clip, width=\linewidth]{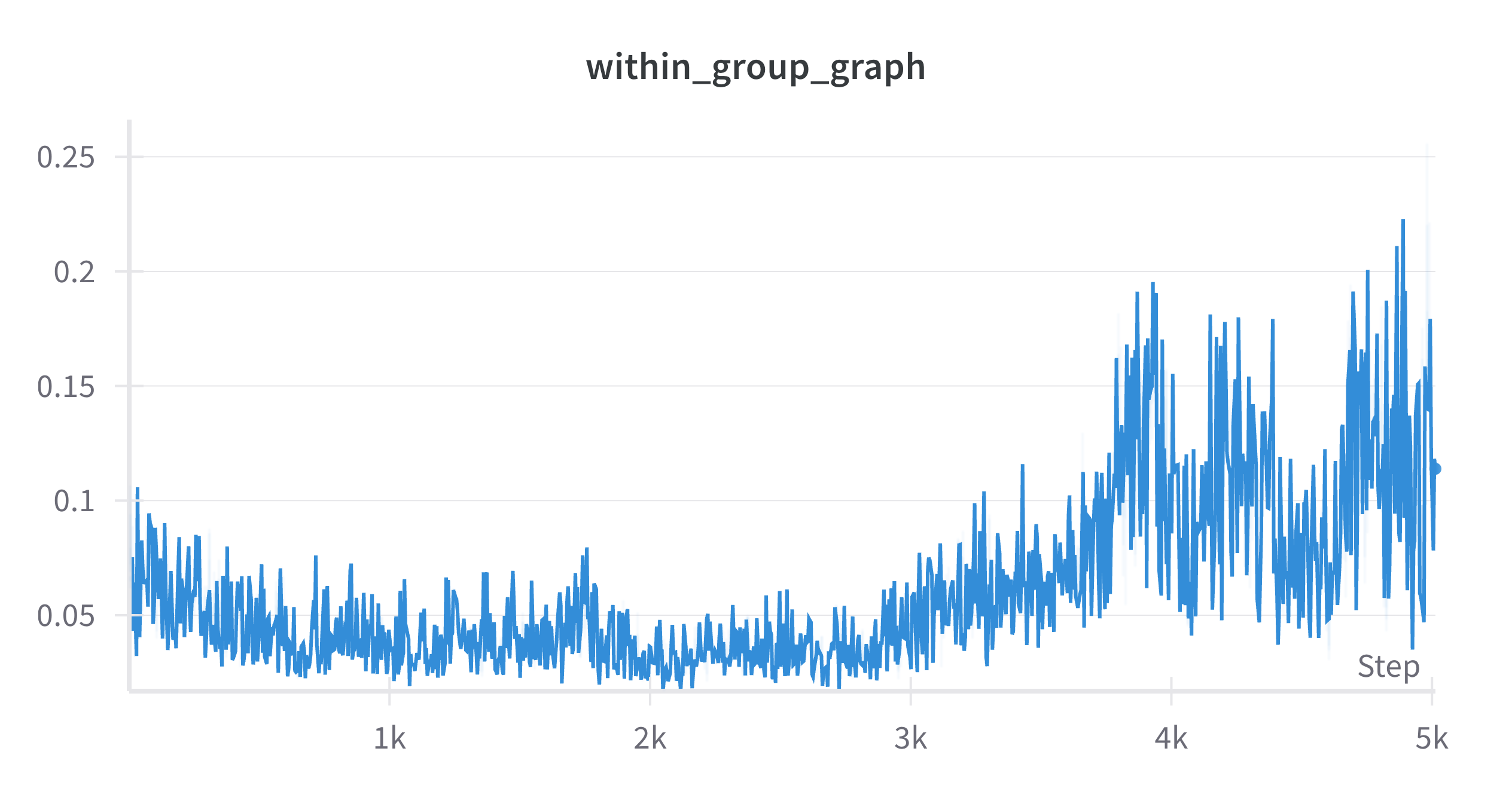}
\caption{Distance within graph}
\end{subfigure}
\hfill
\begin{subfigure}[t]{0.48\textwidth}
\centering
\includegraphics[trim=0 10 0 160, clip, width=\linewidth]{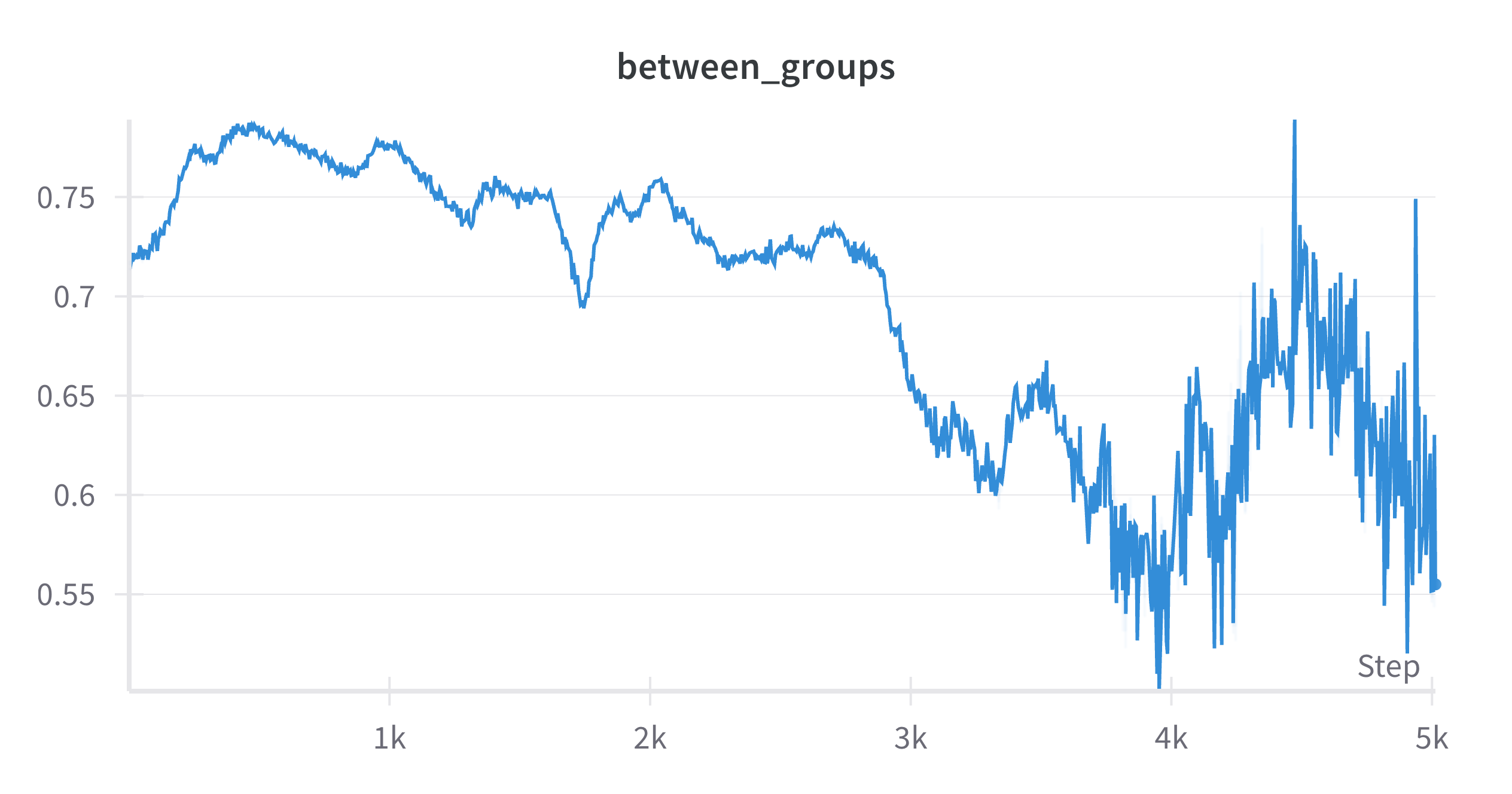}
\caption{Distance between text and graph}
\end{subfigure}
  % \vspace{\baselineskip}
  \caption{
    Results for ETRE on the TDDMan dataset when no CoD is applied.}
  \label{fig:etre_tddman-no-cod_results}
\end{figure*}

See Figure~\ref{fig:etre_tbd_results} and Figure~\ref{fig:etre_tddauto_results} for results on TimeBank-Dense and TDDAuto datasets, respectively. See Figure~\ref{fig:etre_tddman-no-cod_results} for results on TDDMan dataset when no CoD is applied.

\subsection{MLRE results}
% \paragraph{Detailed task performance across languages}
% See Table~\ref{tab:f1_multiling}.

\begin{figure*}[hbtp]
  \centering
  
  % Row 1: Three PCA plots
  \begin{subfigure}[t]{0.325\textwidth}
    \centering
    \includegraphics[trim=38 18 45 50, clip, width=\linewidth]{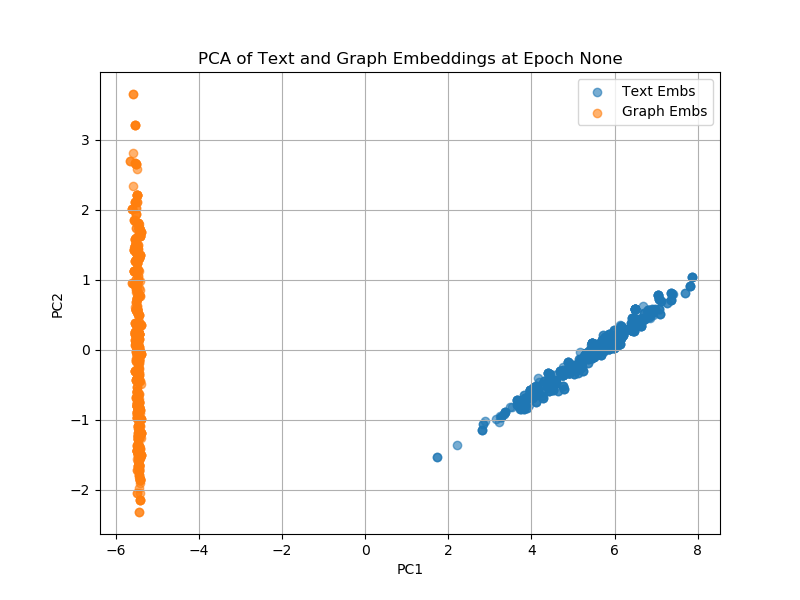}
    \caption{Initial epoch (de)}
  \end{subfigure}
  \hfill
  \begin{subfigure}[t]{0.325\textwidth}
    \centering
    \includegraphics[trim=38 18 45 50, clip, width=\linewidth]{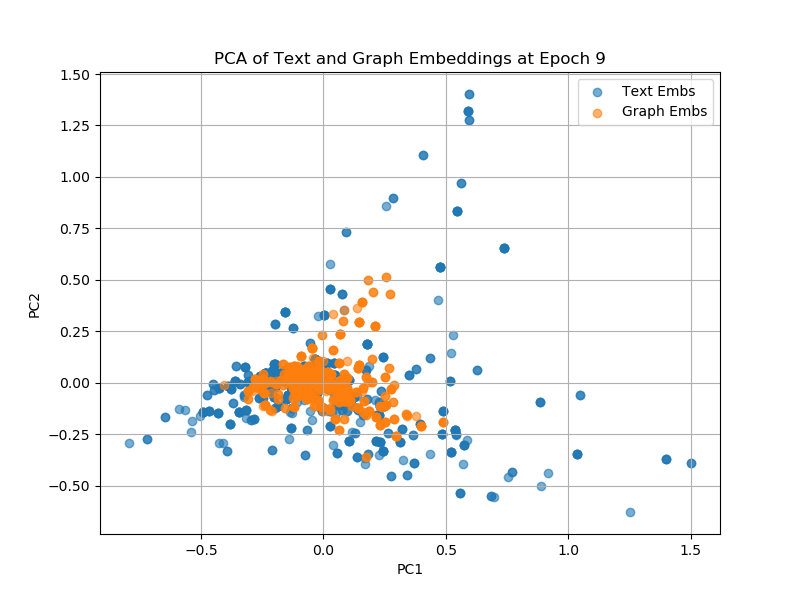}
    \caption{Intermediate epoch (de)}
  \end{subfigure}
  \hfill
  \begin{subfigure}[t]{0.325\textwidth}
    \centering
    \includegraphics[trim=38 18 45 50, clip, width=\linewidth]{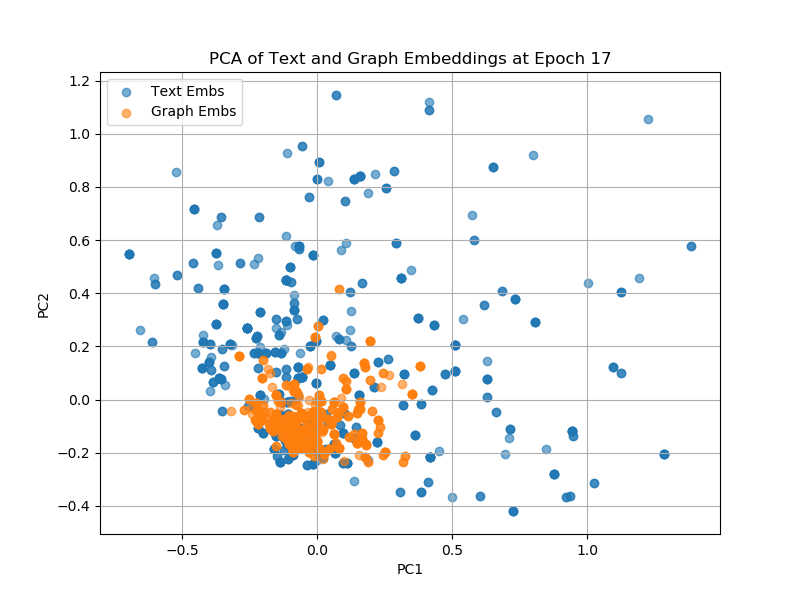}
    \caption{Final epoch (de)}
  \end{subfigure}

  \begin{subfigure}[t]{0.325\textwidth}
    \centering
    \includegraphics[trim=38 18 45 50, clip, width=\linewidth]{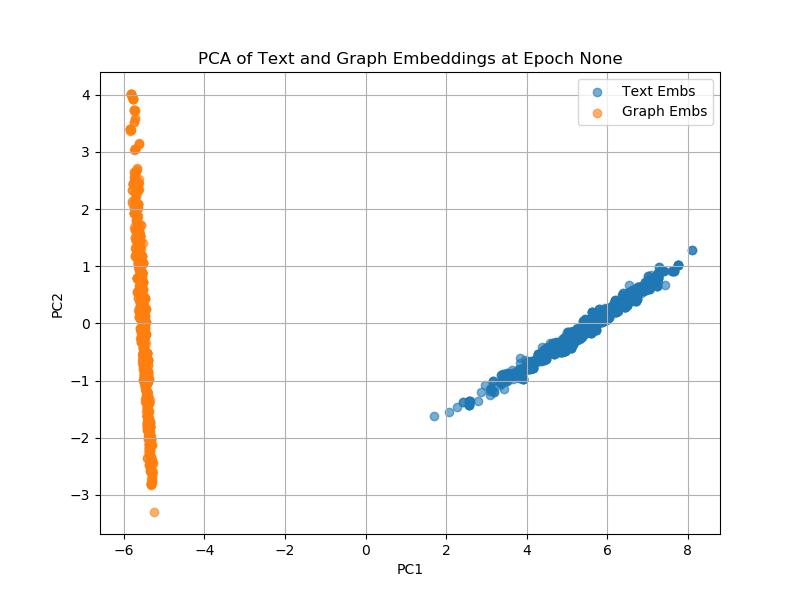}
    \caption{Initial epoch (en)}
  \end{subfigure}
  \hfill
  \begin{subfigure}[t]{0.325\textwidth}
    \centering
    \includegraphics[trim=38 18 45 50, clip, width=\linewidth]{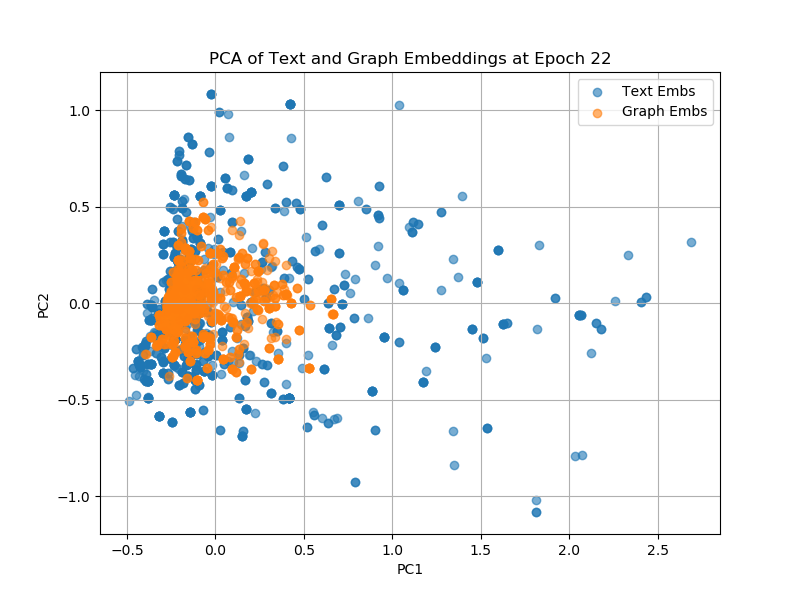}
    \caption{Intermediate epoch (en)}
  \end{subfigure}
  \hfill
  \begin{subfigure}[t]{0.325\textwidth}
    \centering
    \includegraphics[trim=38 18 45 50, clip, width=\linewidth]{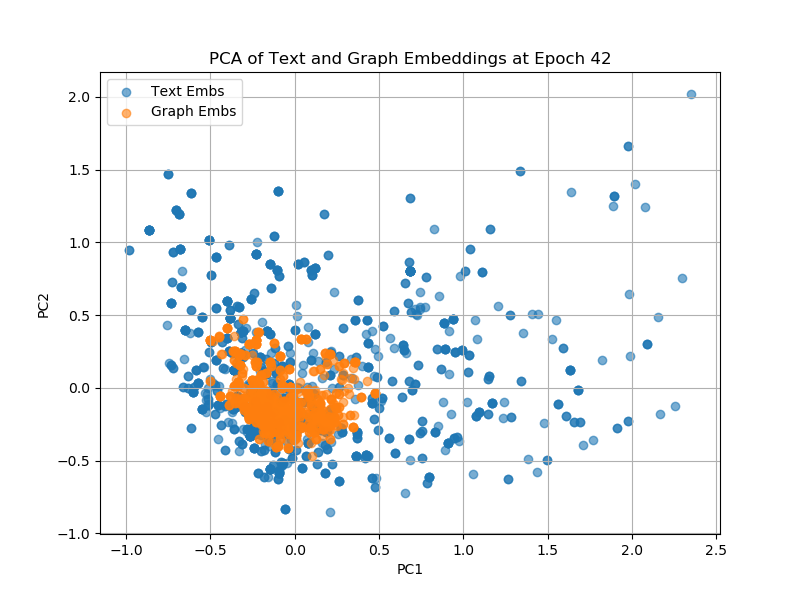}
    \caption{Final epoch (en)}
  \end{subfigure}

  \begin{subfigure}[t]{0.325\textwidth}
    \centering
    \includegraphics[trim=38 18 45 50, clip, width=\linewidth]{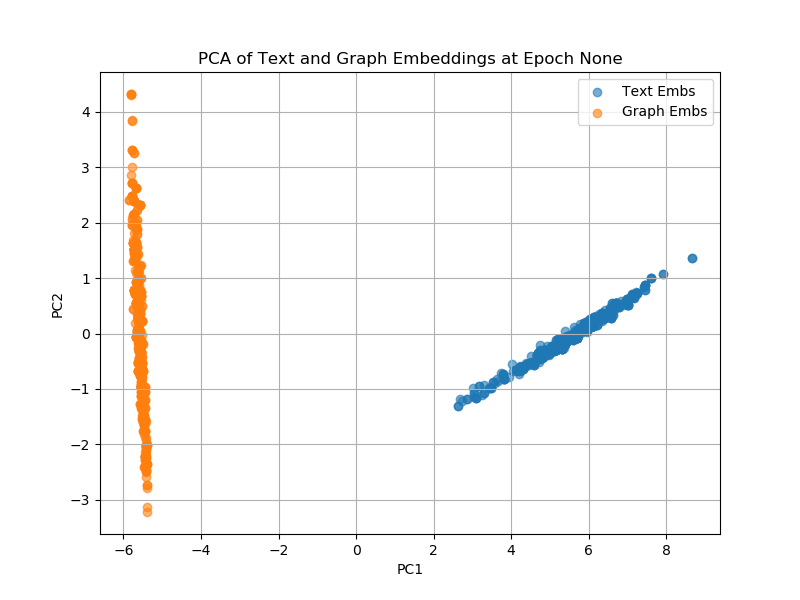}
    \caption{Initial epoch (es)}
  \end{subfigure}
  \hfill
  \begin{subfigure}[t]{0.325\textwidth}
    \centering
    \includegraphics[trim=38 18 45 50, clip, width=\linewidth]{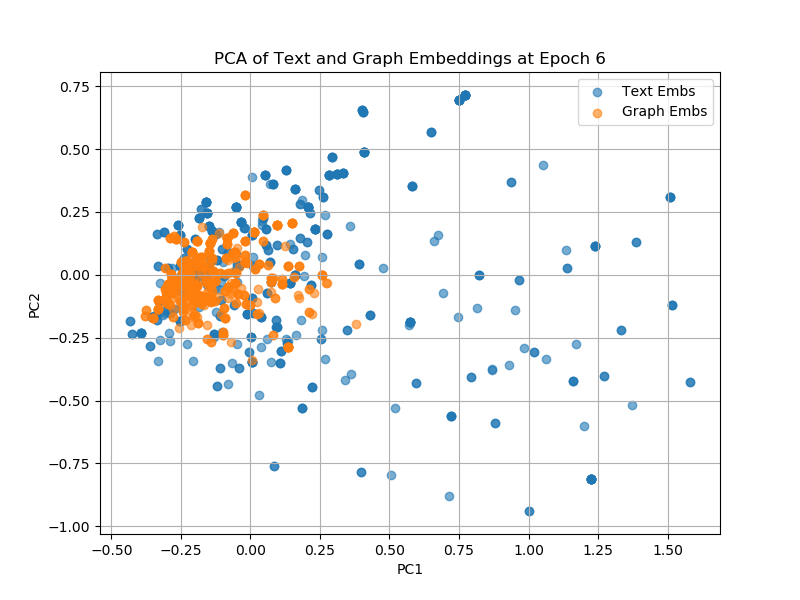}
    \caption{Intermediate epoch (es)}
  \end{subfigure}
  \hfill
  \begin{subfigure}[t]{0.325\textwidth}
    \centering
    \includegraphics[trim=38 18 45 50, clip, width=\linewidth]{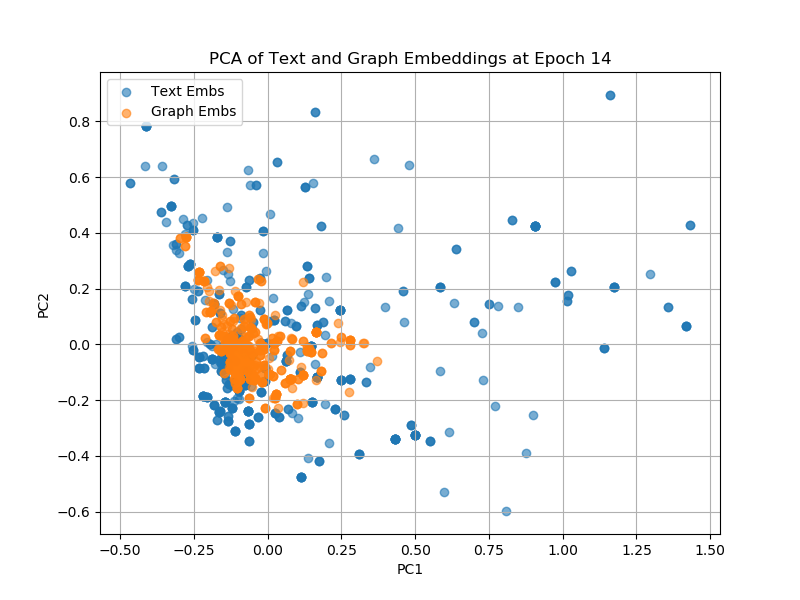}
    \caption{Final epoch (es)}
  \end{subfigure}

  \begin{subfigure}[t]{0.325\textwidth}
    \centering
    \includegraphics[trim=38 18 45 50, clip, width=\linewidth]{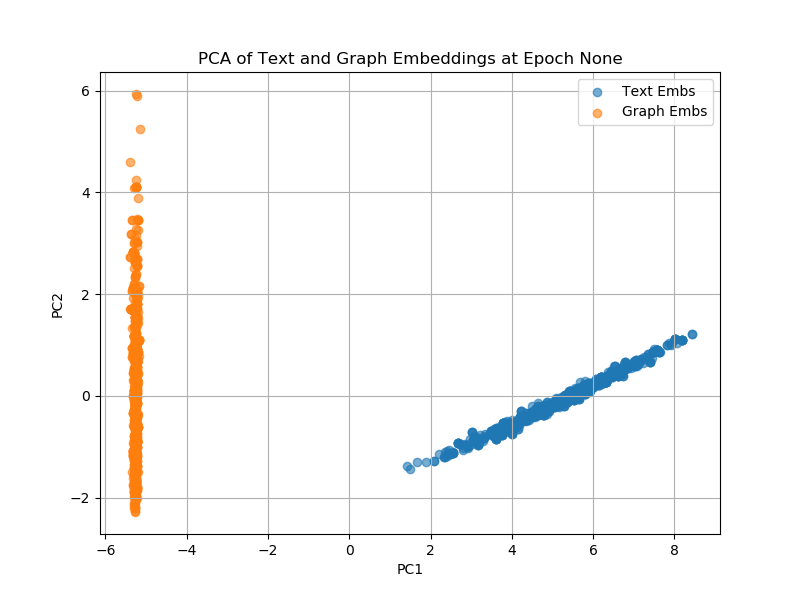}
    \caption{Initial epoch (fr)}
  \end{subfigure}
  \hfill
  \begin{subfigure}[t]{0.325\textwidth}
    \centering
    \includegraphics[trim=38 18 45 50, clip, width=\linewidth]{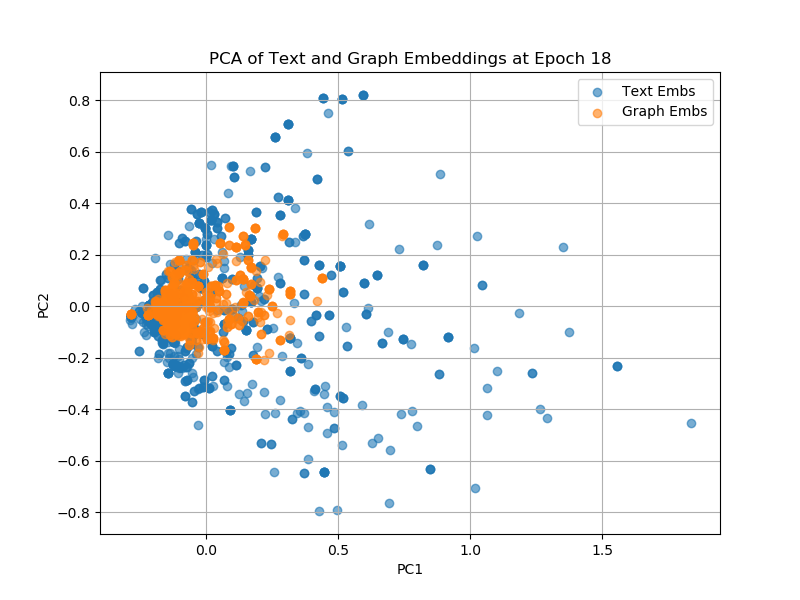}
    \caption{Intermediate epoch (fr)}
  \end{subfigure}
  \hfill
  \begin{subfigure}[t]{0.325\textwidth}
    \centering
    \includegraphics[trim=38 18 45 50, clip, width=\linewidth]{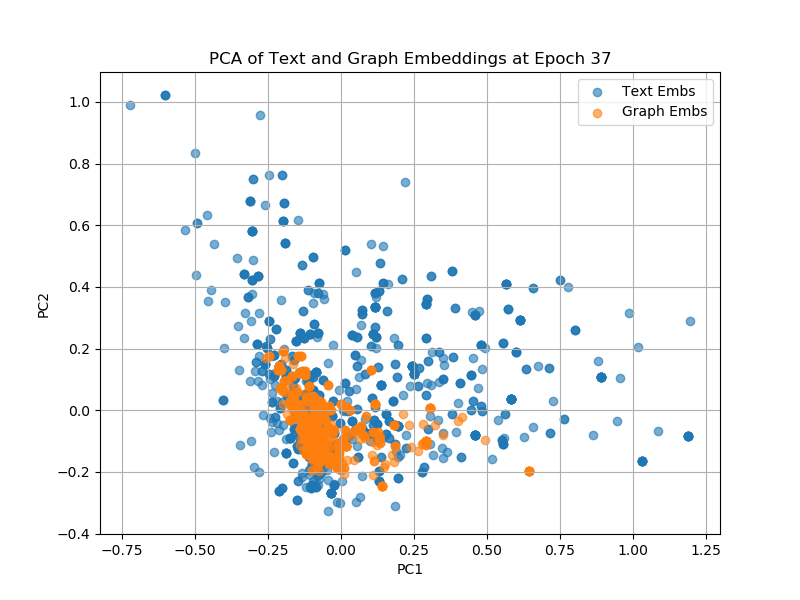}
    \caption{Final epoch (fr)}
  \end{subfigure}

  \begin{subfigure}[t]{0.325\textwidth}
    \centering
    \includegraphics[trim=38 18 45 50, clip, width=\linewidth]{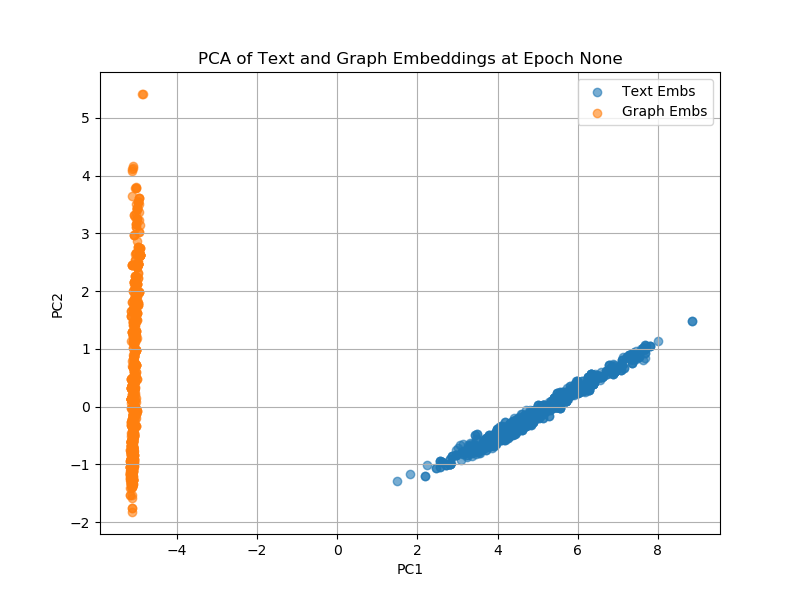}
    \caption{Initial epoch (it)}
  \end{subfigure}
  \hfill
  \begin{subfigure}[t]{0.325\textwidth}
    \centering
    \includegraphics[trim=38 18 45 50, clip, width=\linewidth]{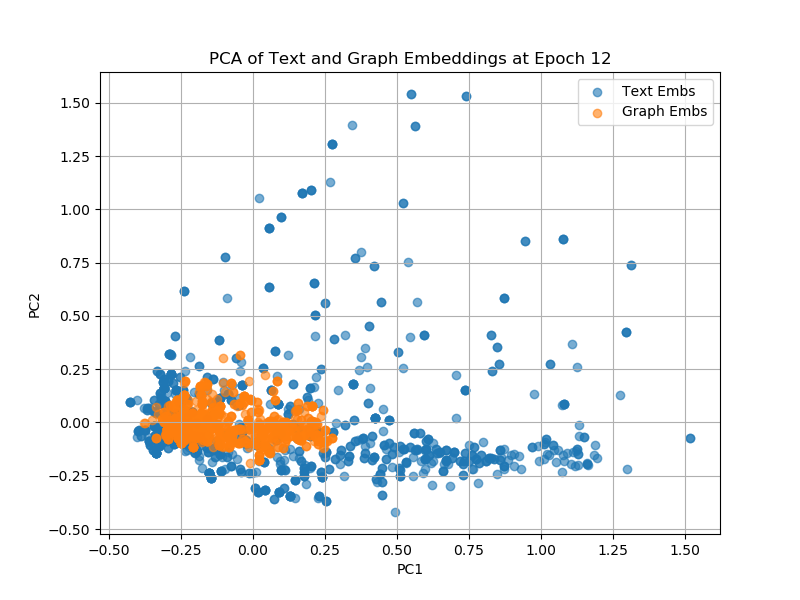}
    \caption{Intermediate epoch (it)}
  \end{subfigure}
  \hfill
  \begin{subfigure}[t]{0.325\textwidth}
    \centering
    \includegraphics[trim=38 18 45 50, clip, width=\linewidth]{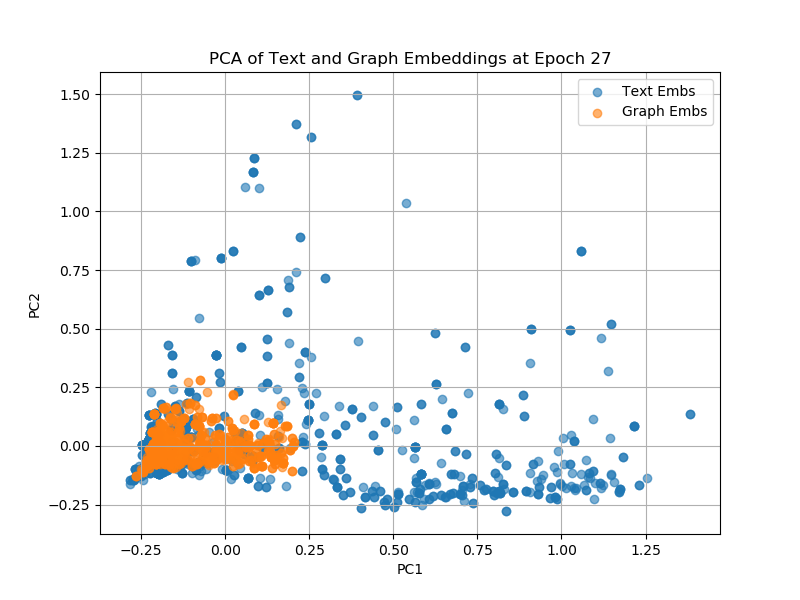}
    \caption{Final epoch (it)}
  \end{subfigure}

  \caption{
    PCA plots for MLRE across the different languages in the RED\textsuperscript{fm} dataset. }
  \label{fig:REDFM_PCA_results}
\end{figure*}

\begin{figure*}[hbtp]
  \centering
  
  % Row 1: Three PCA plots
  % \begin{subfigure}[t]{0.32\textwidth}
  %   \centering
  %   \includegraphics[trim=38 25 70 70, clip, width=\linewidth]{figures/multiling/text_dist.png}
  %   \caption{Initial epoch}
  % \end{subfigure}
  % \hfill
  % \begin{subfigure}[t]{0.32\textwidth}
  %   \centering
  %   \includegraphics[trim=38 25 70 70, clip, width=\linewidth]{figures/etre/tbd/pca_5.png}
  %   \caption{Intermediate epoch}
  % \end{subfigure}
  % \hfill
  % \begin{subfigure}[t]{0.32\textwidth}
  %   \centering
  %   \includegraphics[trim=38 25 70 70, clip, width=\linewidth]{figures/etre/tbd/pca_9.png}
  %   \caption{Final epoch}
  % \end{subfigure}
  
  % \vspace{1em}
  
  % Row 2: Distance-related plots in 2x2 grid, trimmed more aggressively
\begin{subfigure}[t]{0.48\textwidth}
\centering
\includegraphics[width=\linewidth]{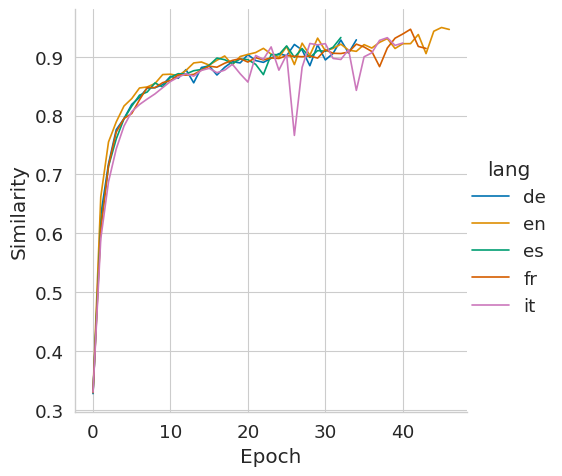}
\caption{Cosine similarity}
\end{subfigure}
\hfill
\begin{subfigure}[t]{0.48\textwidth}
\centering
\includegraphics[width=\linewidth]{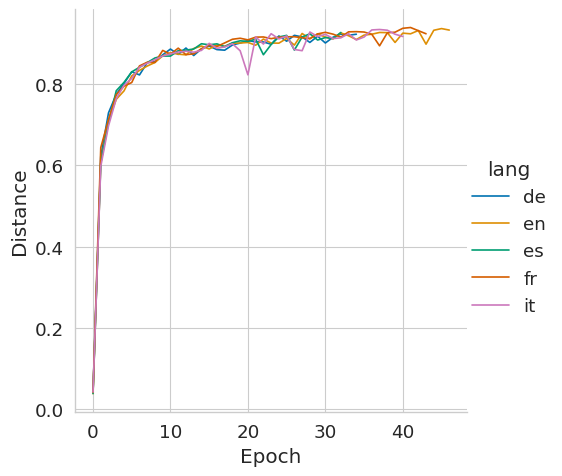}
\caption{Distance within text}
\end{subfigure}
\vspace{1em}
\begin{subfigure}[t]{0.48\textwidth}
\centering
\includegraphics[width=\linewidth]{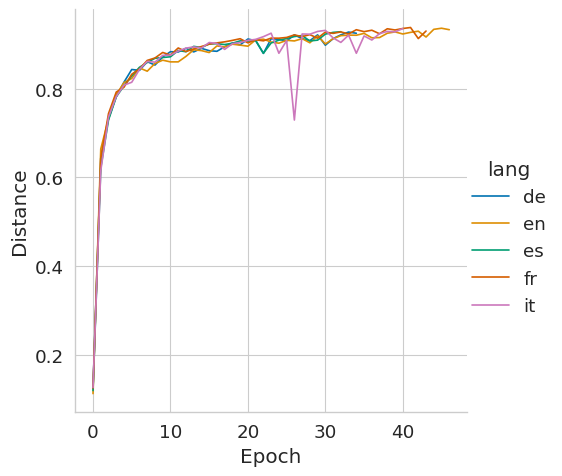}
\caption{Distance within graph}
\end{subfigure}
\hfill
\begin{subfigure}[t]{0.48\textwidth}
\centering
\includegraphics[width=\linewidth]{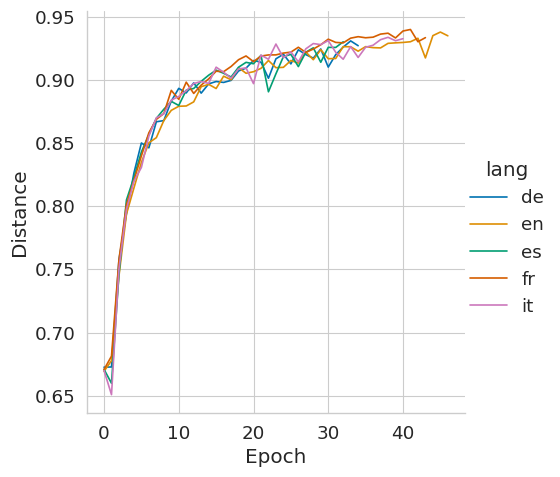}
\caption{Distance between text and graph}
\end{subfigure}
  \caption{
    Cosine similarity and distance results for MLRE on the RED\textsuperscript{fm} dataset.}
  \label{fig:multiling_results}
\end{figure*}

% \paragraph{Visualizations}
See Figure~\ref{fig:REDFM_PCA_results} for PCA plots, and Figure~\ref{fig:multiling_results} for cosine similarity and distance metrics results.

% PLOTTING THE RESULTS FOR THE DIFFERENT LANGUAGES IN THE MLRE task

\subsection{FU results}
\begin{figure*}[hbtp]
  \centering
  
  % Row 1: Three PCA plots
  \begin{subfigure}[t]{0.32\textwidth}
    \centering
    \includegraphics[trim=38 25 70 70, clip, width=\linewidth]{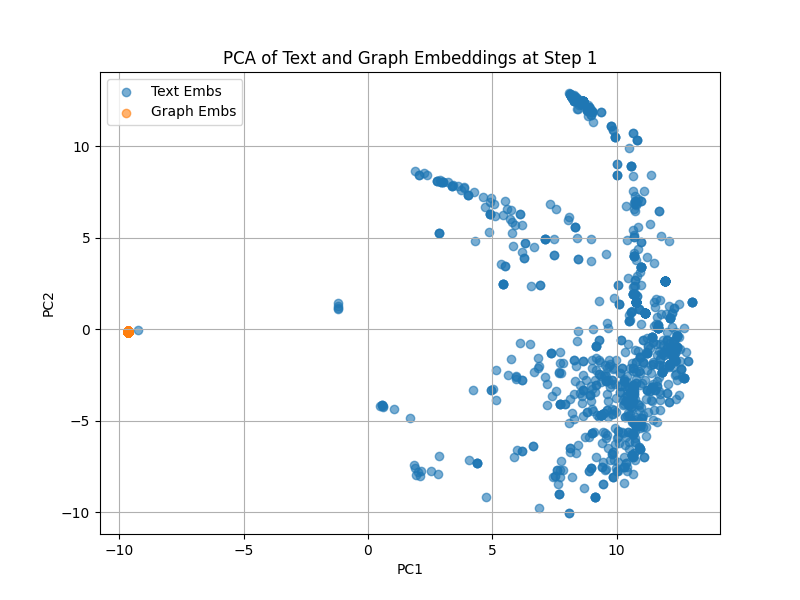}
    \caption{Initial epoch}
  \end{subfigure}
  \hfill
  \begin{subfigure}[t]{0.32\textwidth}
    \centering
    \includegraphics[trim=38 25 70 70, clip, width=\linewidth]{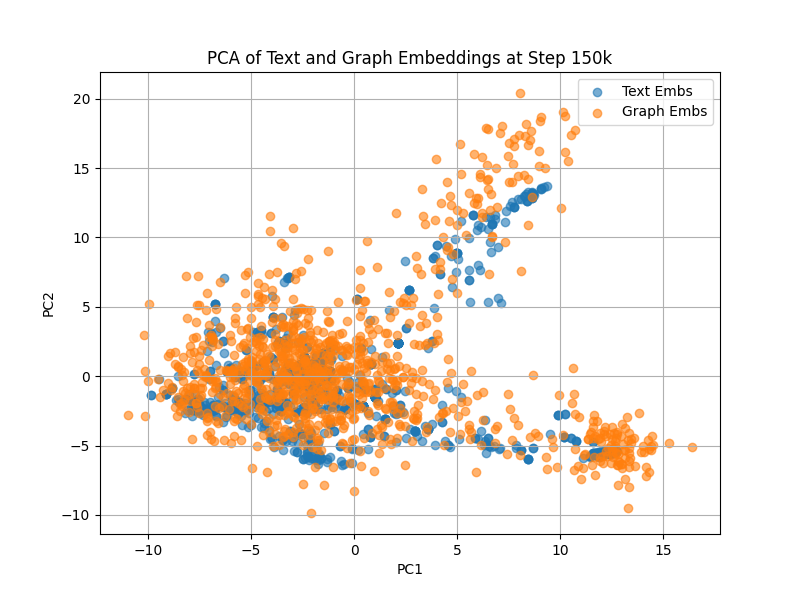}
    \caption{Intermediate epoch}
  \end{subfigure}
  \hfill
  \begin{subfigure}[t]{0.32\textwidth}
    \centering
    \includegraphics[trim=38 25 70 70, clip, width=\linewidth]{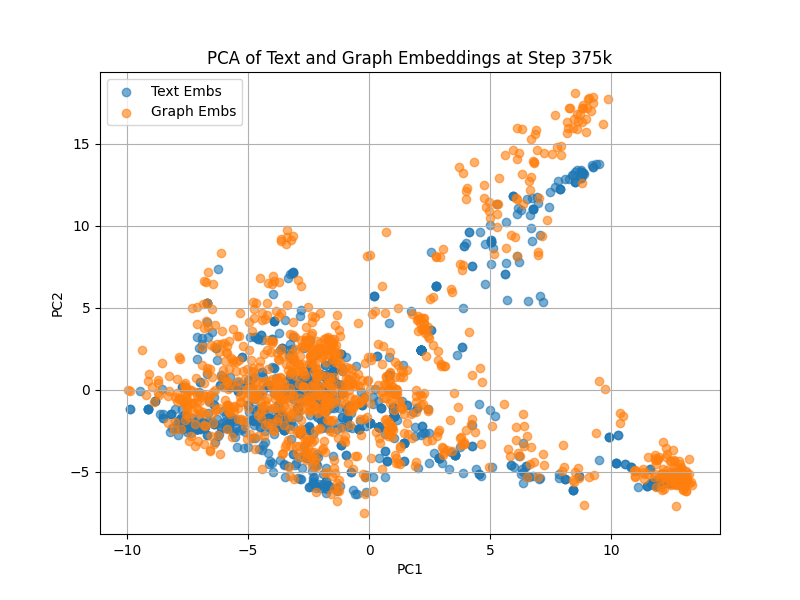}
    \caption{Final epoch}
  \end{subfigure}
  
  % \vspace{1em}
  
  % Row 2: Four distance-related plots
  % \begin{subfigure}[t]{0.22\textwidth}
  %   \centering
  %   \includegraphics[width=\linewidth]{figures/for}
  %   \caption{Cosine similarity}
  % \end{subfigure}
  % \hfill
  \begin{subfigure}[t]{0.32\textwidth}
    \centering
    \includegraphics[trim=0 20 0 20, clip, width=\linewidth]{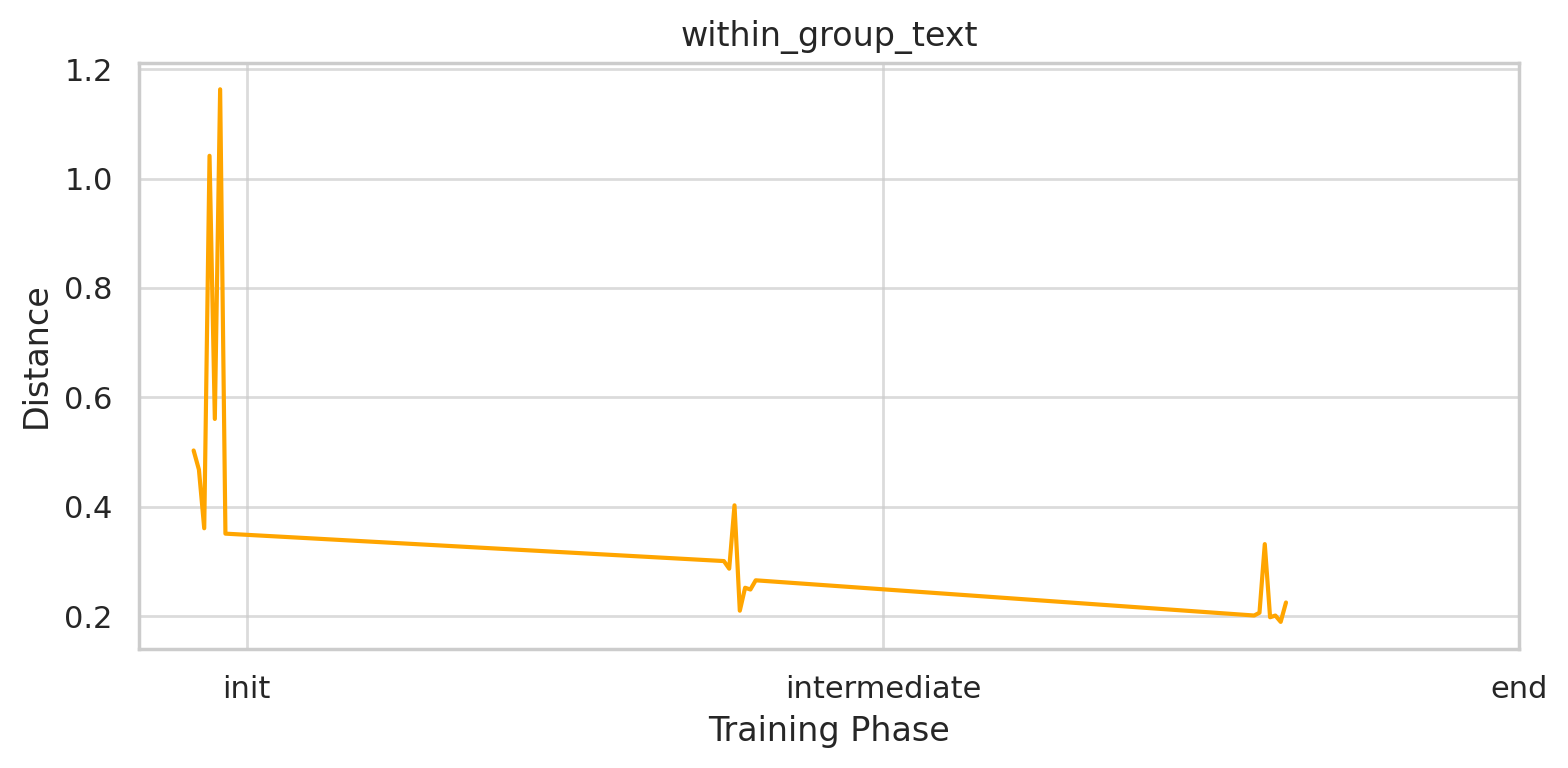}
    \caption{Distance within text}
  \end{subfigure}
  \hfill
  \begin{subfigure}[t]{0.32\textwidth}
    \centering
    \includegraphics[trim=0 20 0 20, clip, width=\linewidth]{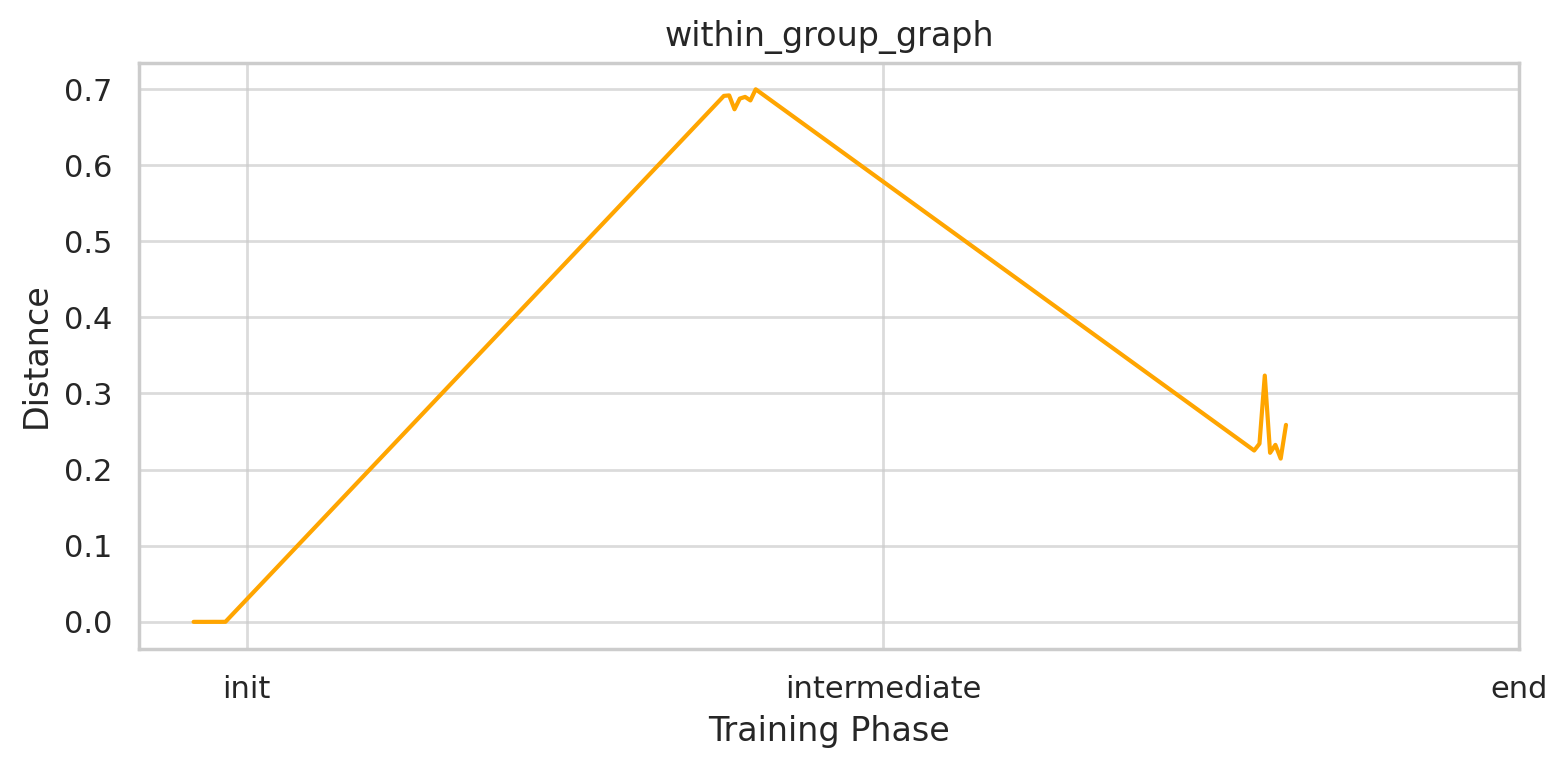}
    \caption{Distance within graph}
  \end{subfigure}
  \hfill
  \begin{subfigure}[t]{0.32\textwidth}
    \centering
    \includegraphics[trim=0 20 0 20, clip, width=\linewidth]{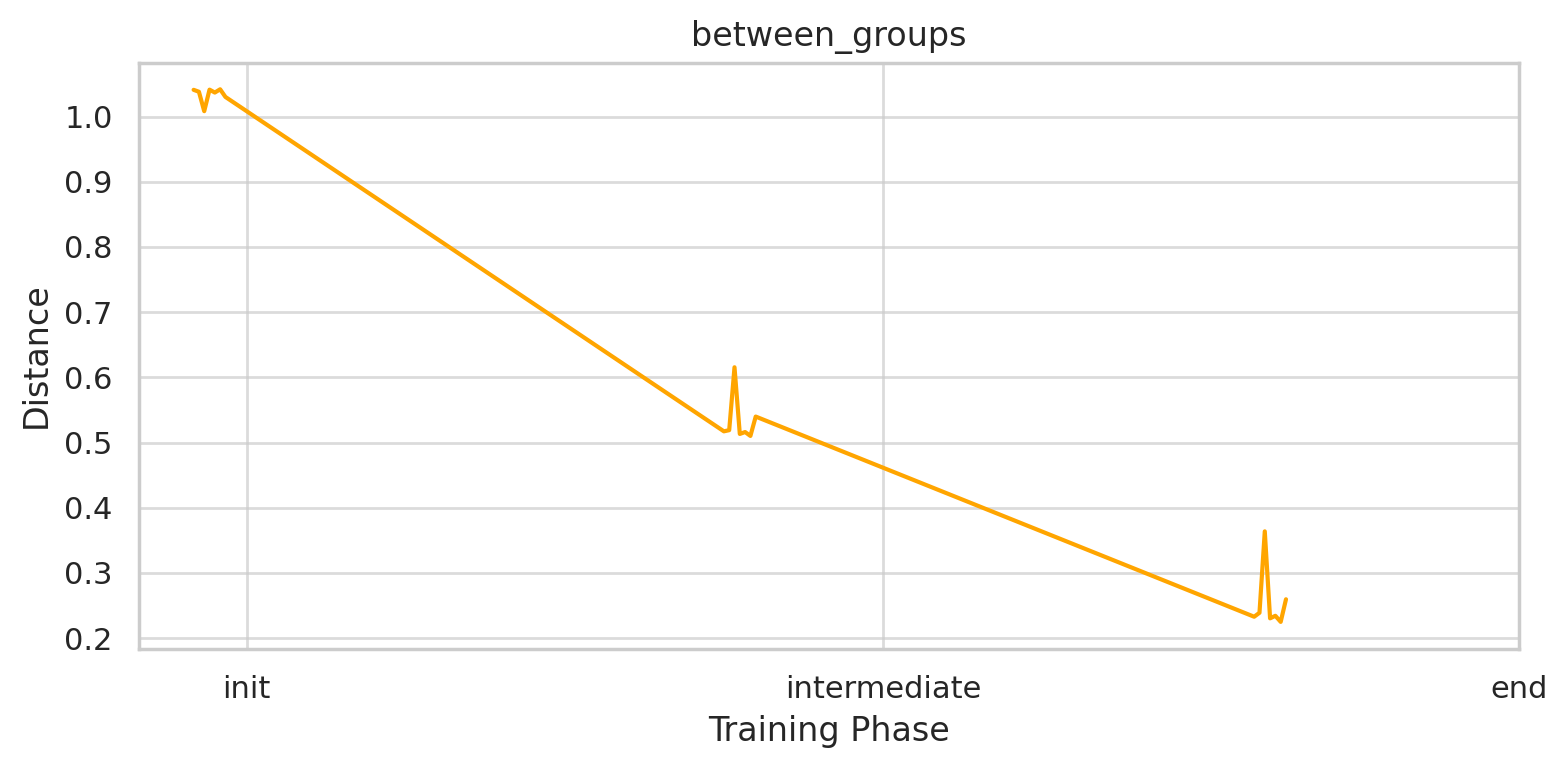}
    \caption{Distance between text and graph}
  \end{subfigure}
  \caption{
    Results for form understanding on the SROIE dataset.}
  \label{fig:form_sroie_results}
\end{figure*}

\begin{figure*}[hbtp]
  \centering
  
   % Row 1: Three PCA plots
  \begin{subfigure}[t]{0.32\textwidth}
    \centering
    \includegraphics[trim=38 25 70 70, clip, width=\linewidth]{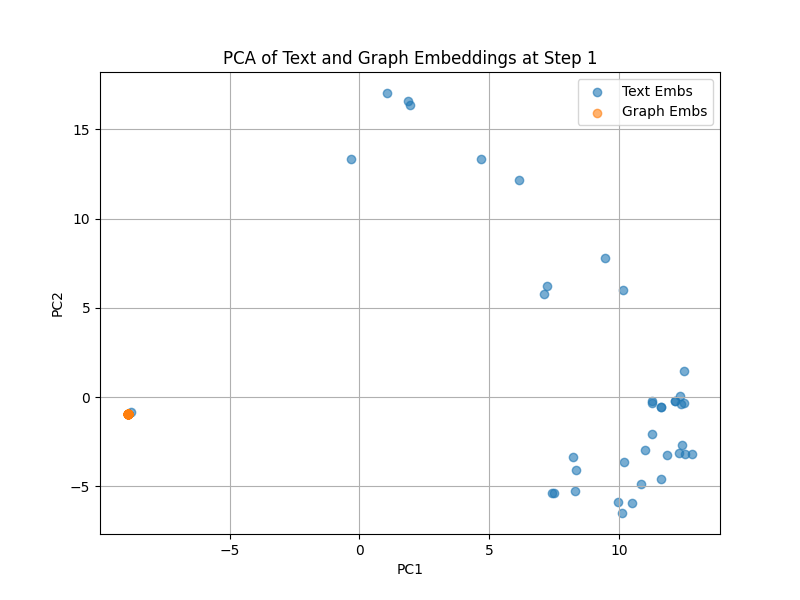}
    \caption{Initial epoch}
  \end{subfigure}
  \hfill
  \begin{subfigure}[t]{0.32\textwidth}
    \centering
    \includegraphics[trim=38 25 70 70, clip, width=\linewidth]{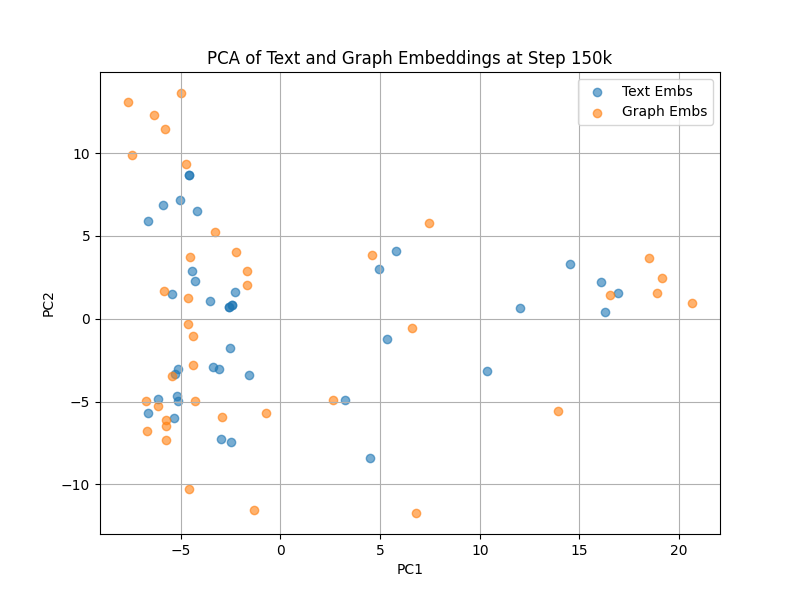}
    \caption{Intermediate epoch}
  \end{subfigure}
  \hfill
  \begin{subfigure}[t]{0.32\textwidth}
    \centering
    \includegraphics[trim=38 25 70 70, clip, width=\linewidth]{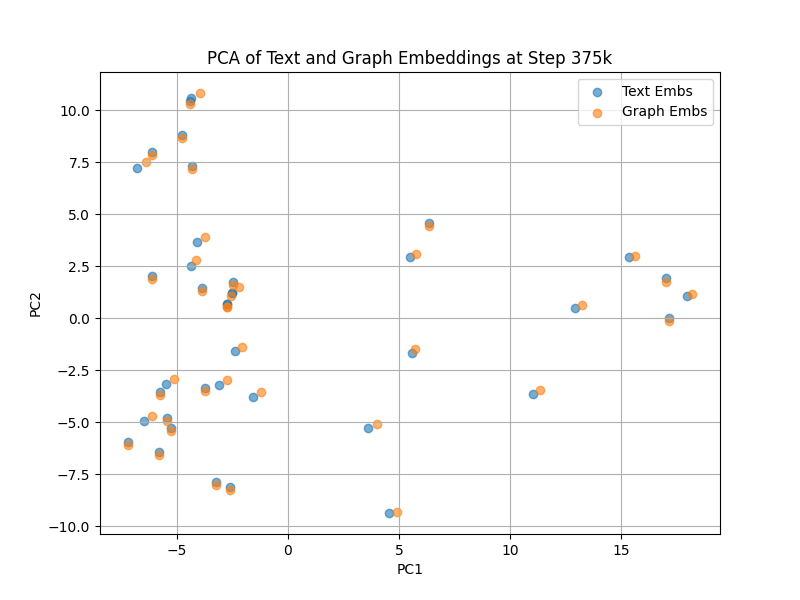}
    \caption{Final epoch}
  \end{subfigure}
  
  % \vspace{1em}
  
  % Row 2: Four distance-related plots
  % \begin{subfigure}[t]{0.22\textwidth}
  %   \centering
  %   \includegraphics[width=\linewidth]{figures/for}
  %   \caption{Cosine similarity}
  % \end{subfigure}
  % \hfill
  \begin{subfigure}[t]{0.32\textwidth}
    \centering
    \includegraphics[trim=0 20 0 20, clip, width=\linewidth]{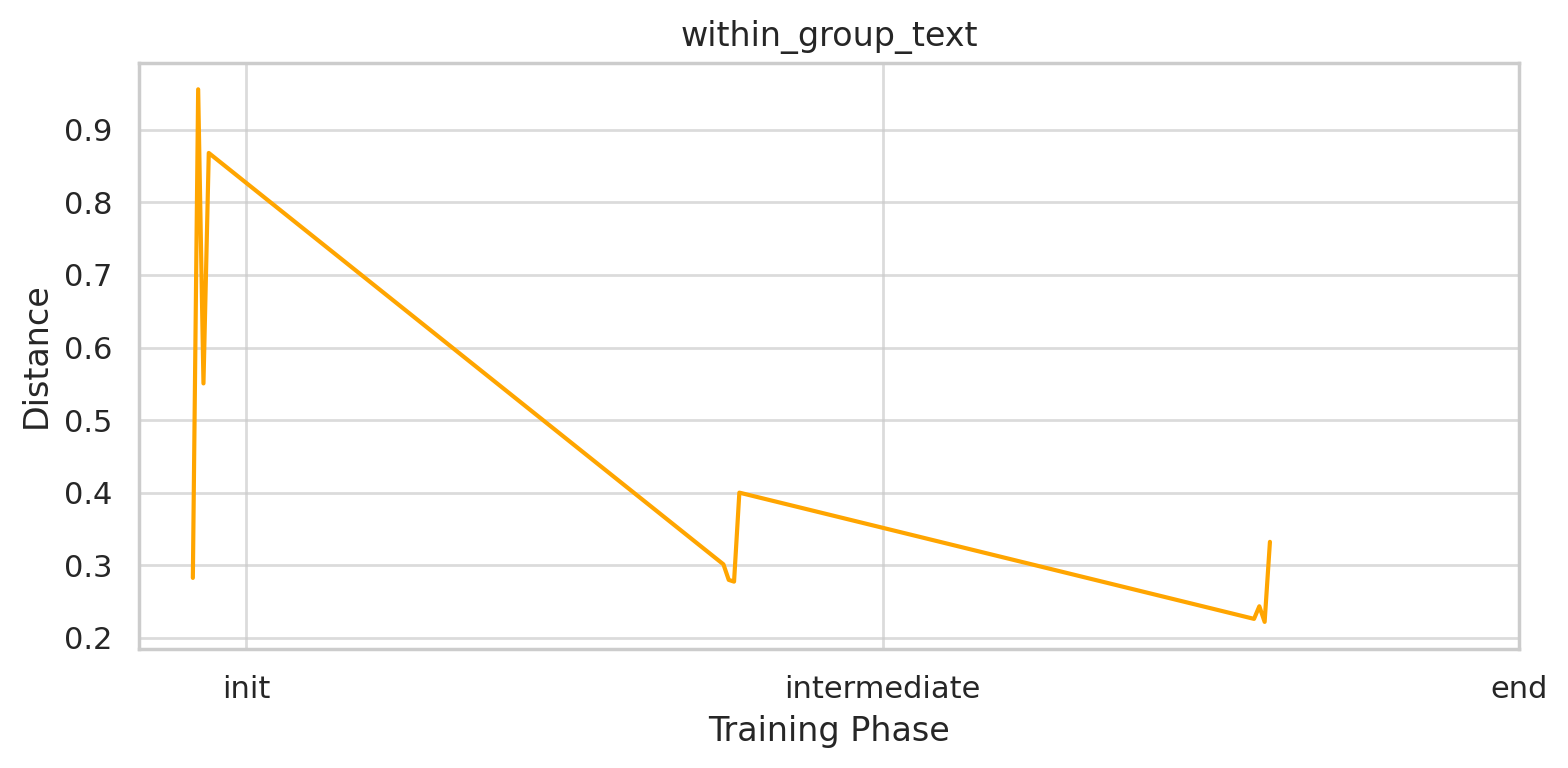}
    \caption{Distance within text}
  \end{subfigure}
  \hfill
  \begin{subfigure}[t]{0.32\textwidth}
    \centering
    \includegraphics[trim=0 20 0 20, clip, width=\linewidth]{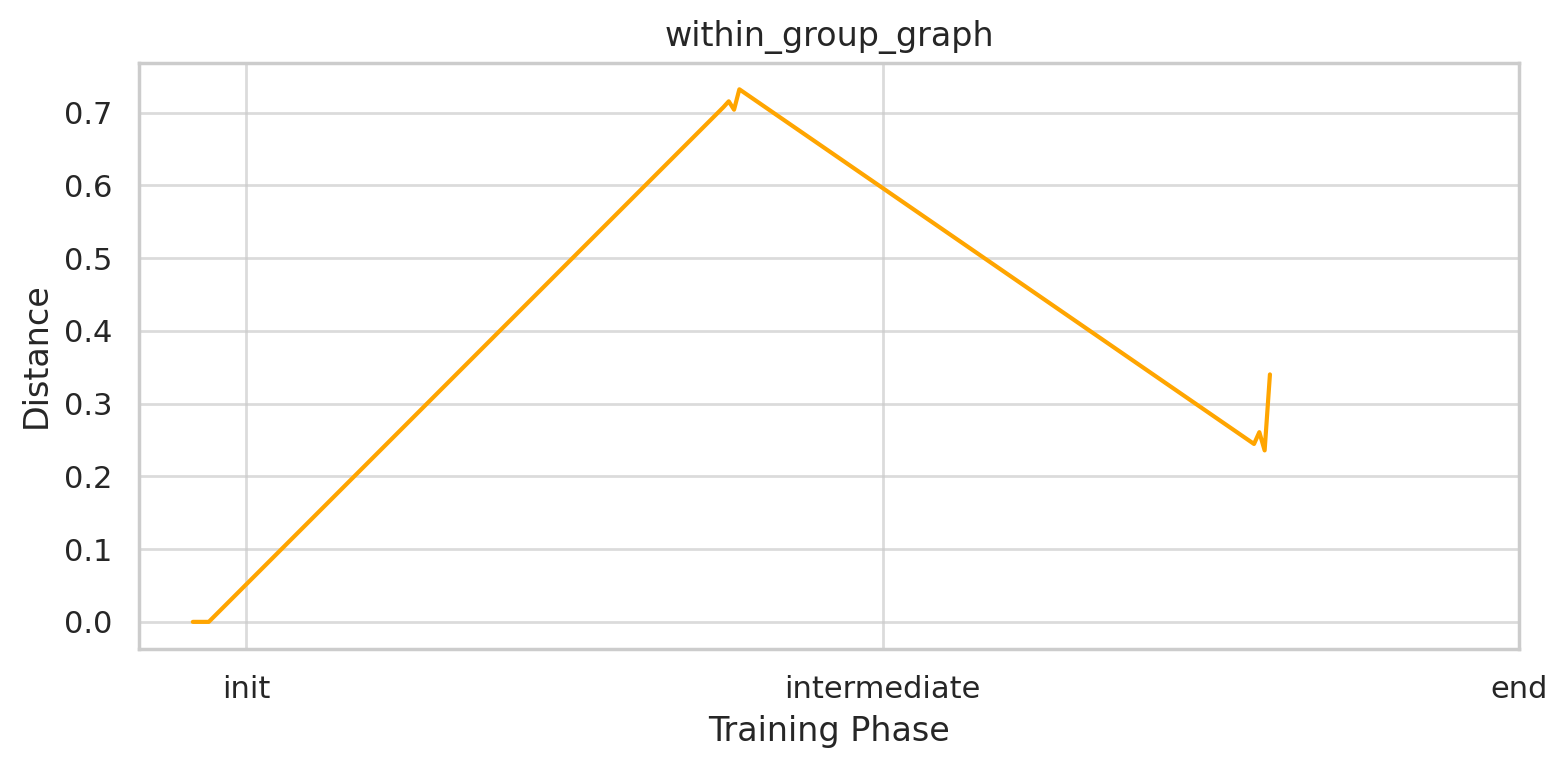}
    \caption{Distance within graph}
  \end{subfigure}
  \hfill
  \begin{subfigure}[t]{0.32\textwidth}
    \centering
    \includegraphics[trim=0 20 0 20, clip, width=\linewidth]{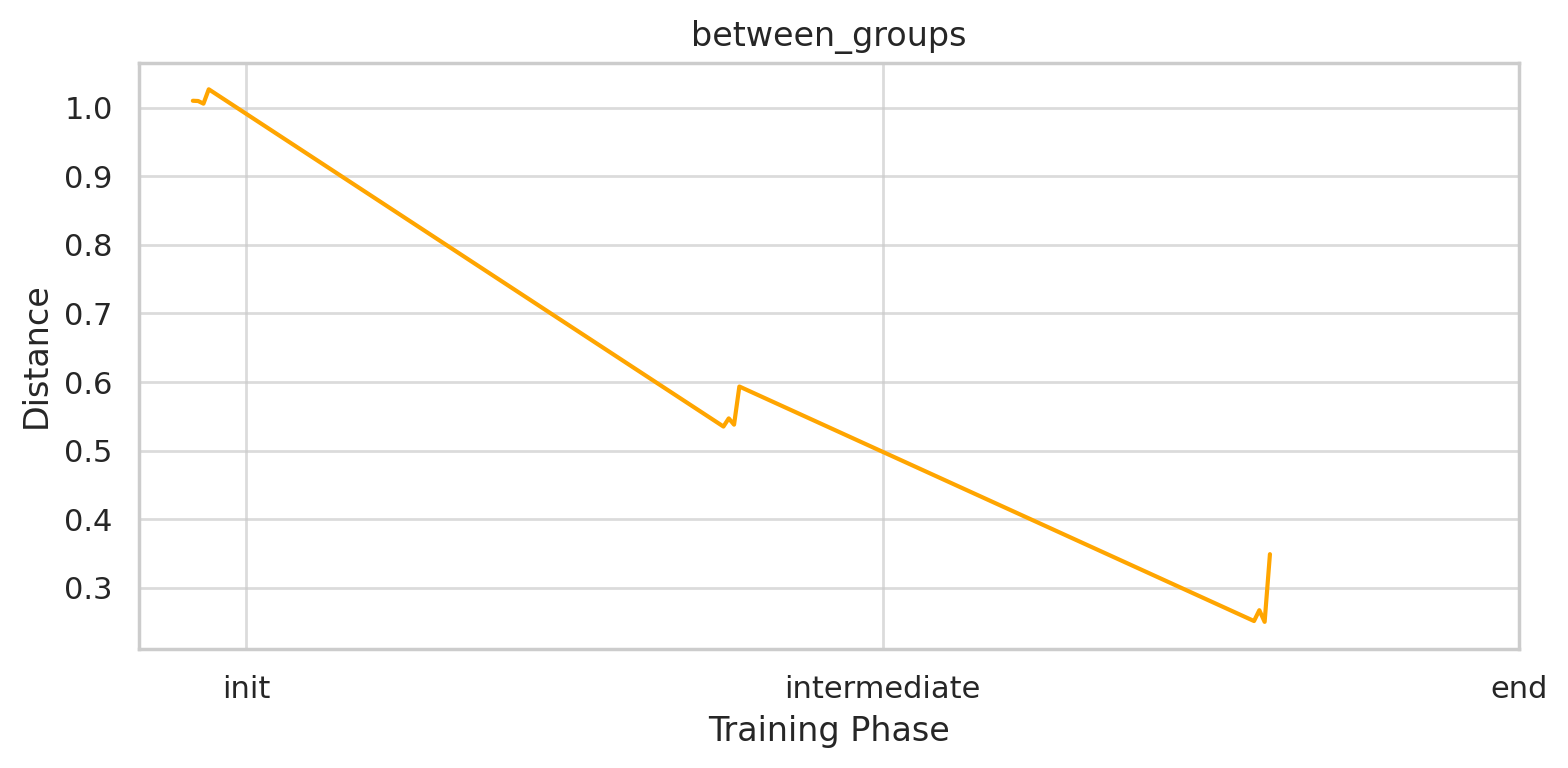}
    \caption{Distance between text and graph}
  \end{subfigure}
  \caption{
    Results for form understanding on the FUNSD dataset.}
  \label{fig:form_funsd_results}
\end{figure*}

See Figure~\ref{fig:form_sroie_results} and Figure~\ref{fig:form_funsd_results} for results on SROIE and FUNSD datasets, respectively.

\subsection{RPP results}

\begin{figure*}[hbtp]
  \centering
  
  % Row 1: Three PCA plots
  \begin{subfigure}[t]{0.32\textwidth}
    \centering
    \includegraphics[trim=38 25 70 70, clip,width=\linewidth]{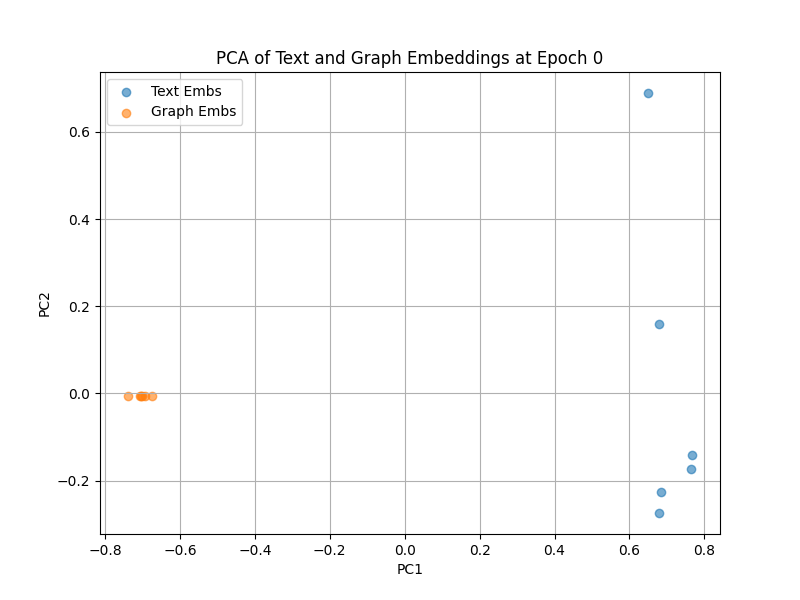}
    \caption{Initial epoch}
  \end{subfigure}
  \hfill
  \begin{subfigure}[t]{0.32\textwidth}
    \centering
    \includegraphics[trim=38 25 70 70, clip,width=\linewidth]{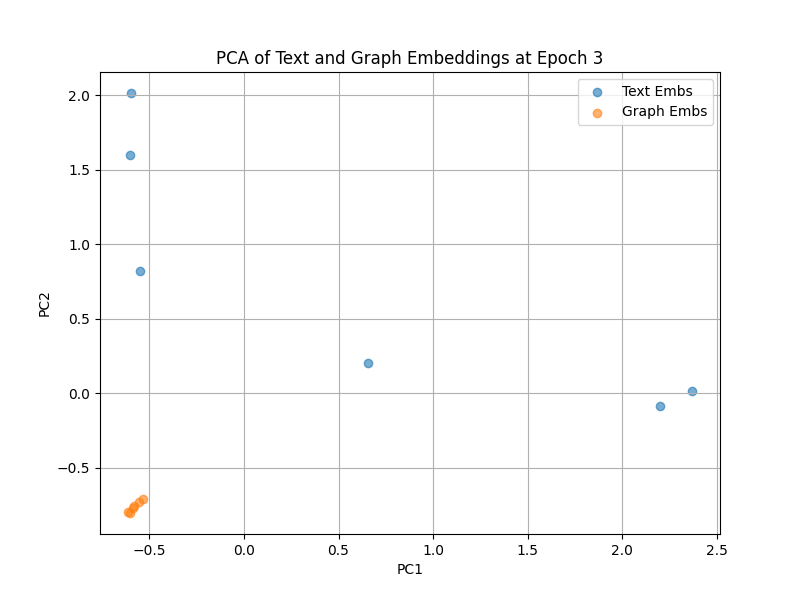}
    \caption{Intermediate epoch}
  \end{subfigure}
  \hfill
  \begin{subfigure}[t]{0.32\textwidth}
    \centering
    \includegraphics[trim=38 25 70 70, clip,width=\linewidth]{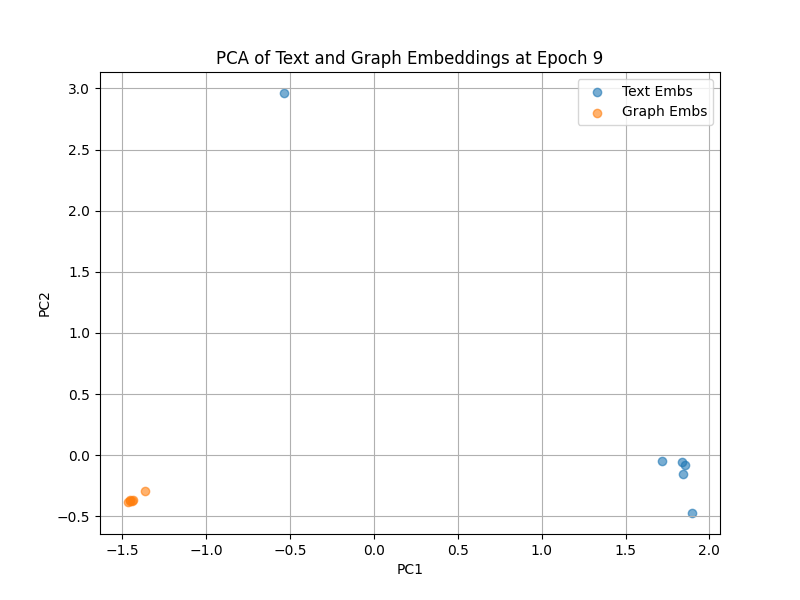}
    \caption{Final epoch}
  \end{subfigure}
  
  % \vspace{1em}
  
  % Row 2: Distance-related plots in 2x2 grid, trimmed more aggressively
\begin{subfigure}[t]{0.48\textwidth}
\centering
\includegraphics[trim=0 10 0 220, clip, width=\linewidth]{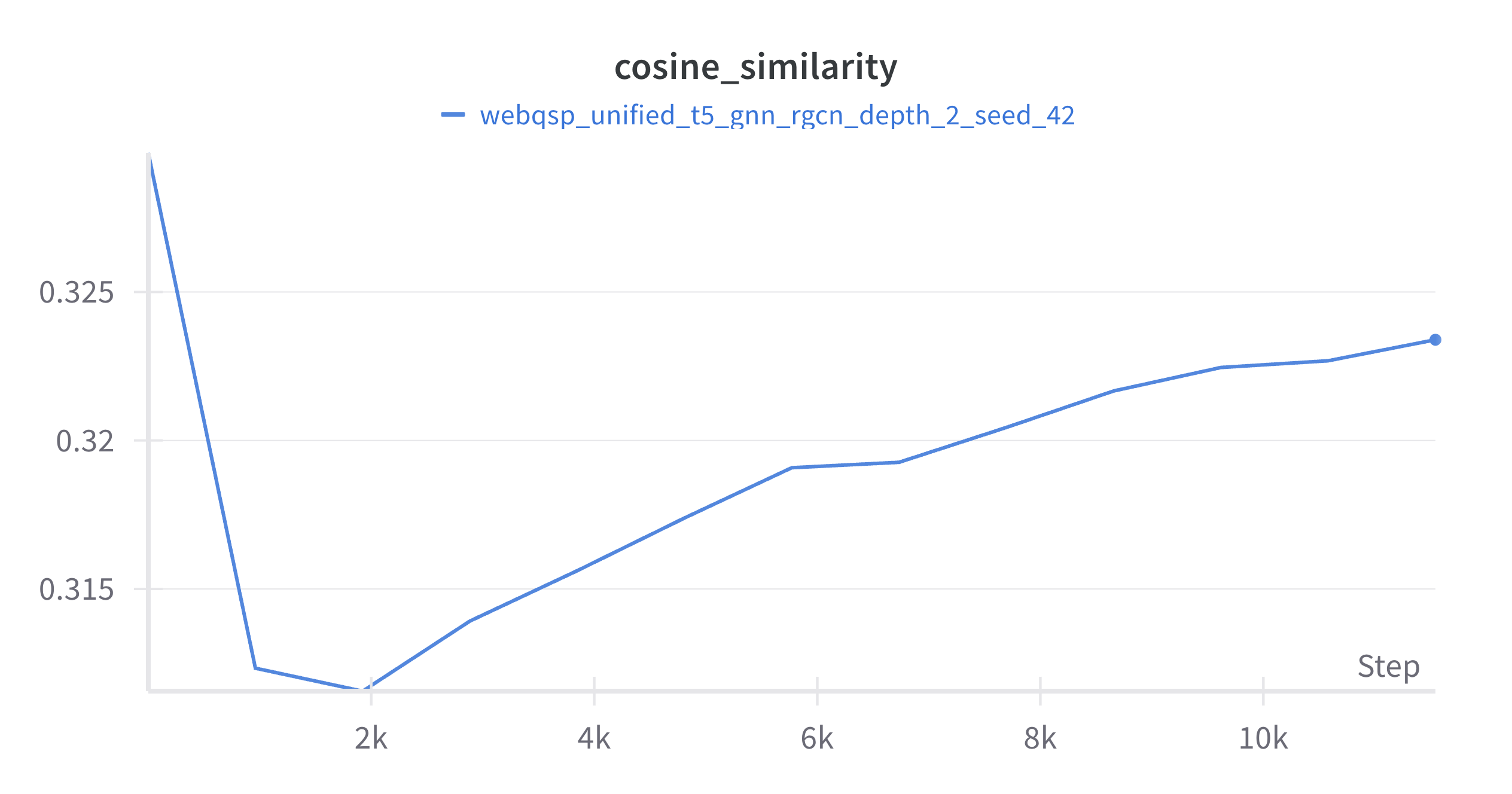}
\caption{Cosine similarity}
\end{subfigure}
\hfill
\begin{subfigure}[t]{0.48\textwidth}
\centering
\includegraphics[trim=0 10 0 220, clip, width=\linewidth]{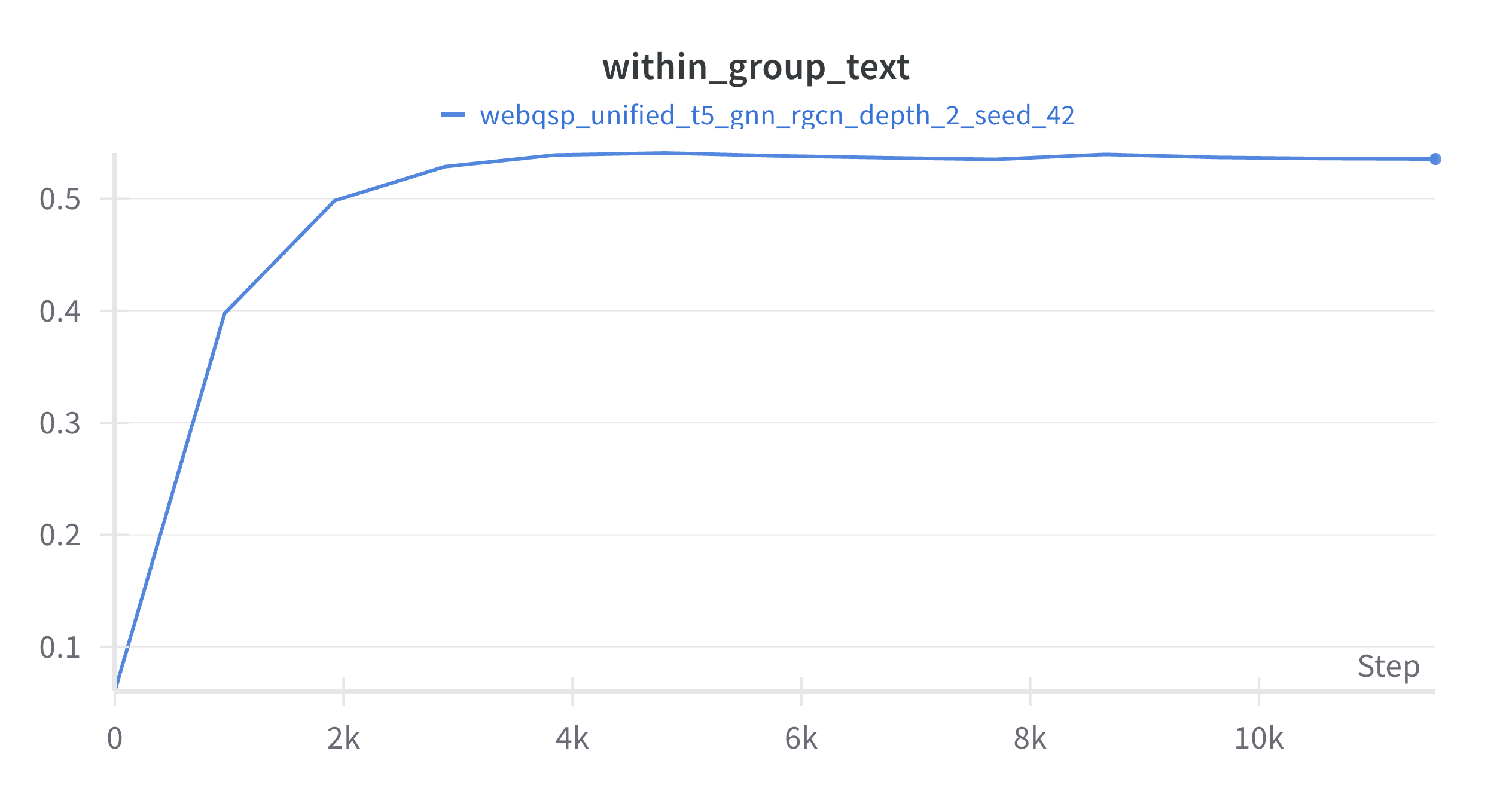}
\caption{Distance within text}
\end{subfigure}
% \vspace{1em}
\begin{subfigure}[t]{0.48\textwidth}
\centering
\includegraphics[trim=0 10 0 220, clip, width=\linewidth]{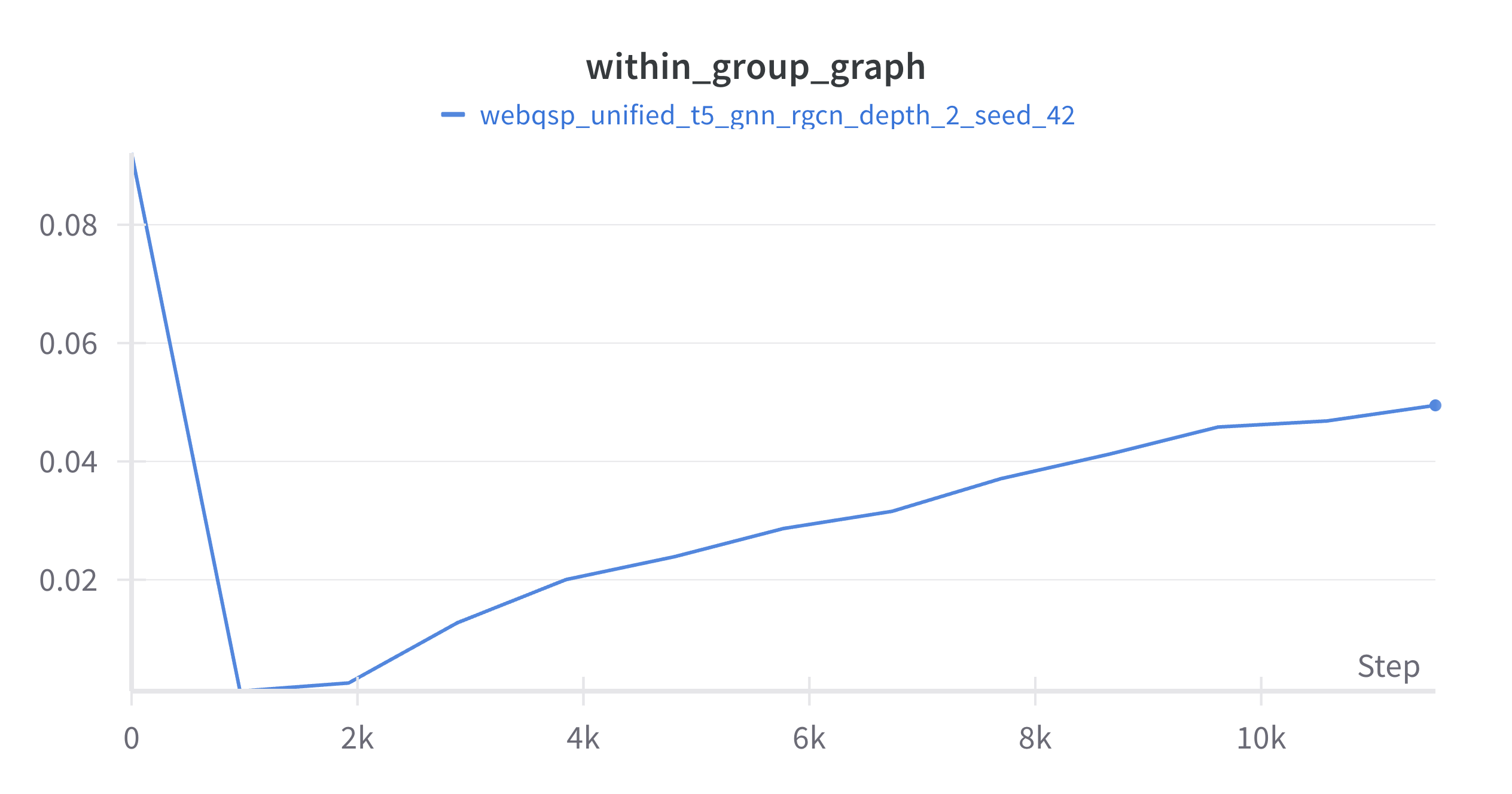}
\caption{Distance within graph}
\end{subfigure}
\hfill
\begin{subfigure}[t]{0.48\textwidth}
\centering
\includegraphics[trim=0 10 0 220, clip, width=\linewidth]{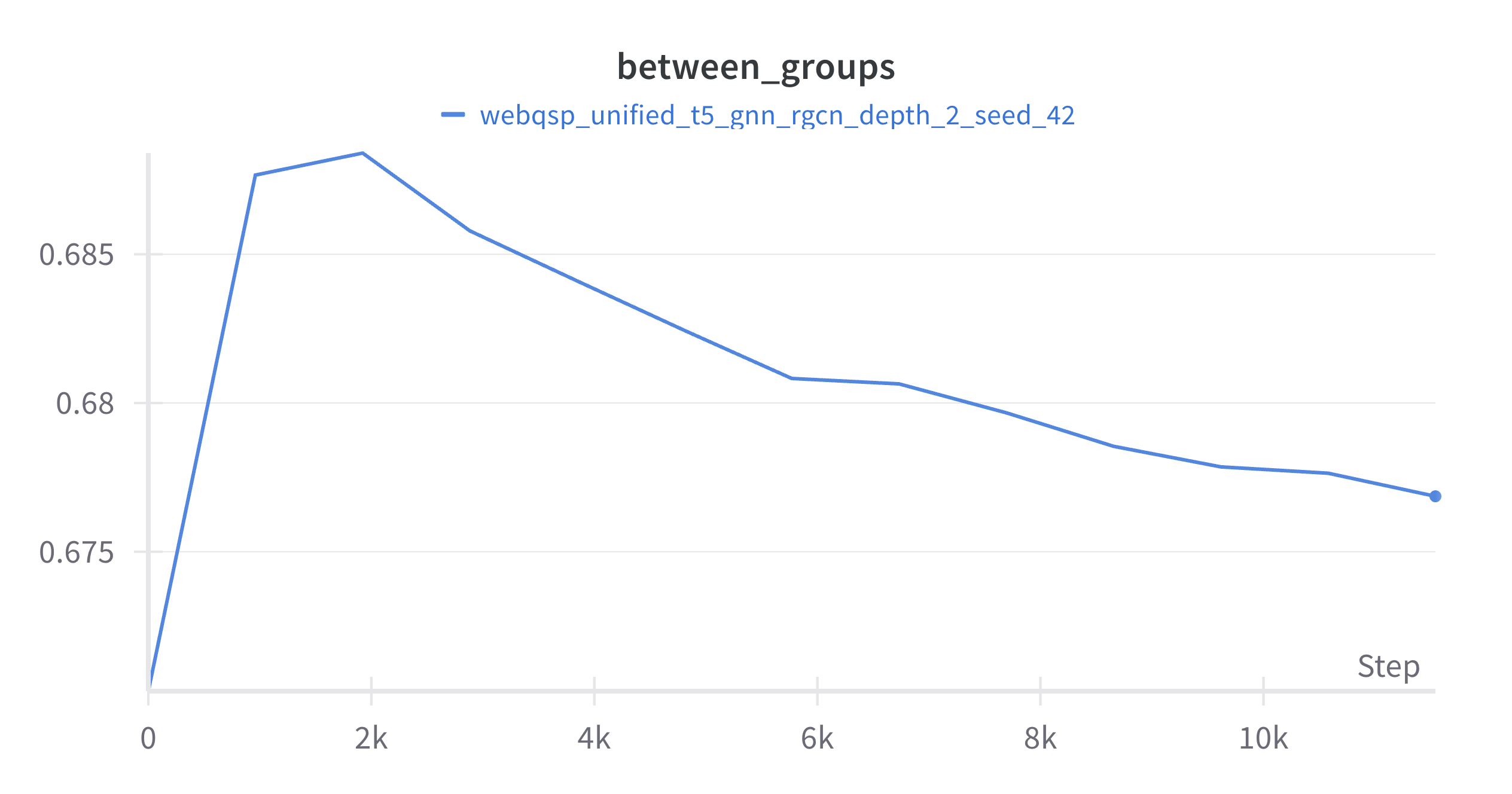}
\caption{Distance between text and graph}
\end{subfigure}
  % \vspace{\baselineskip}
  \caption{
        Results for reasoning pattern prediction on the WebQSP dataset when no CoD is applied.}
  \label{fig:iso_no-cod_results}
\end{figure*}

See Figure~\ref{fig:iso_no-cod_results} for Reasoning Pattern Prediction task without CoD applied.

\subsection{KBQA entity-ranking results}
\begin{figure*}[hbtp]
  \centering
  
  % Row 1: Three PCA plots
  \begin{subfigure}[t]{0.32\textwidth}
    \centering
    \includegraphics[trim=38 25 70 70, clip,width=\linewidth]{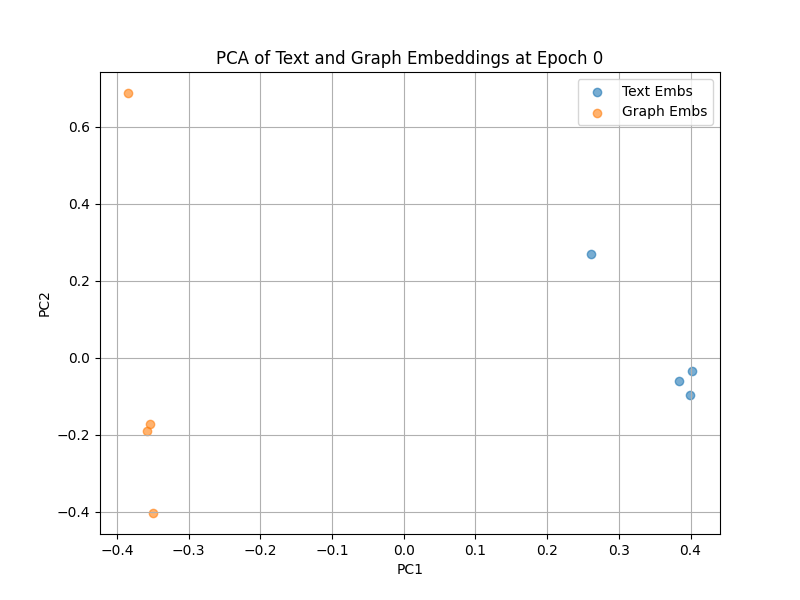}
    \caption{Initial epoch}
  \end{subfigure}
  \hfill
  \begin{subfigure}[t]{0.32\textwidth}
    \centering
    \includegraphics[trim=38 25 70 70, clip,width=\linewidth]{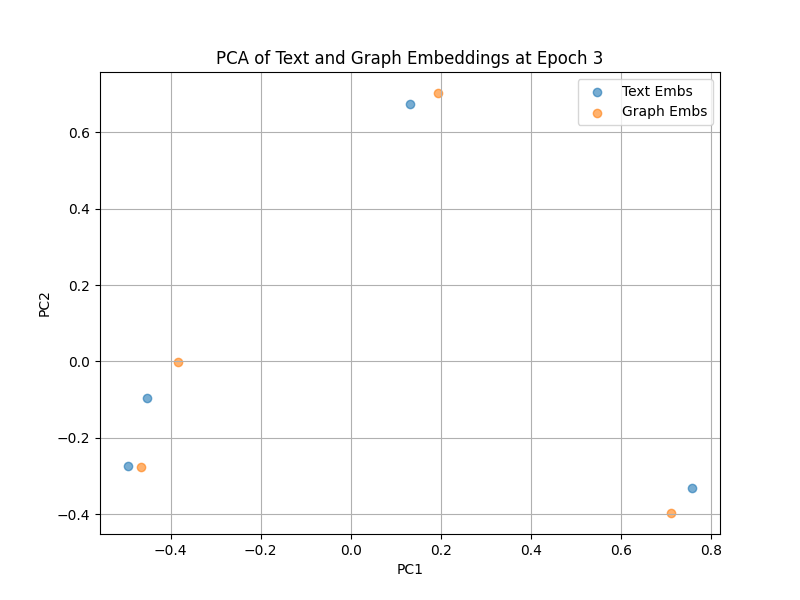}
    \caption{Intermediate epoch}
  \end{subfigure}
  \hfill
  \begin{subfigure}[t]{0.32\textwidth}
    \centering
    \includegraphics[trim=38 25 70 70, clip,width=\linewidth]{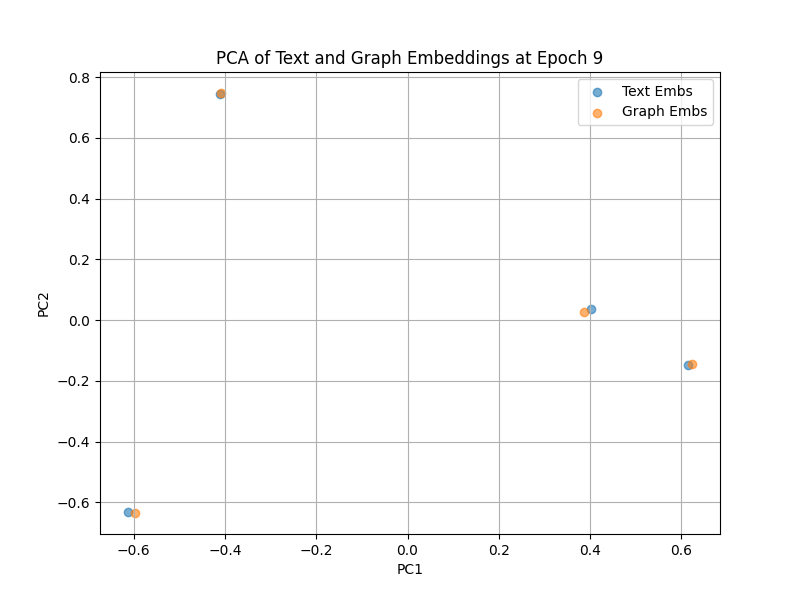}
    \caption{Final epoch}
  \end{subfigure}
  
  % \vspace{1em}
  
  % Row 2: Distance-related plots in 2x2 grid, trimmed more aggressively
\begin{subfigure}[t]{0.48\textwidth}
\centering
\includegraphics[trim=0 10 0 175, clip, width=\linewidth]{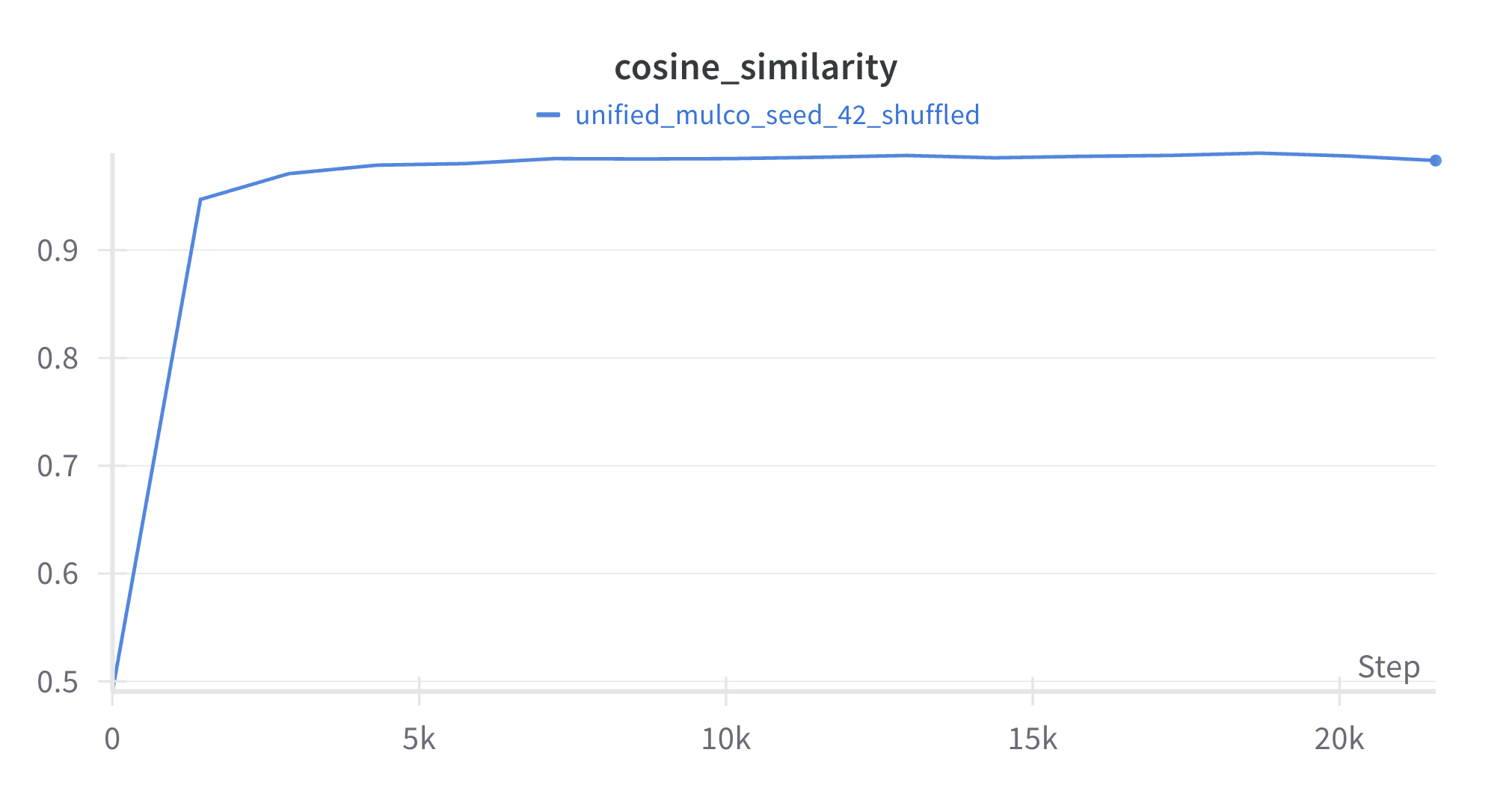}
\caption{Cosine similarity}
\end{subfigure}
\hfill
\begin{subfigure}[t]{0.48\textwidth}
\centering
\includegraphics[trim=0 10 0 175, clip, width=\linewidth]{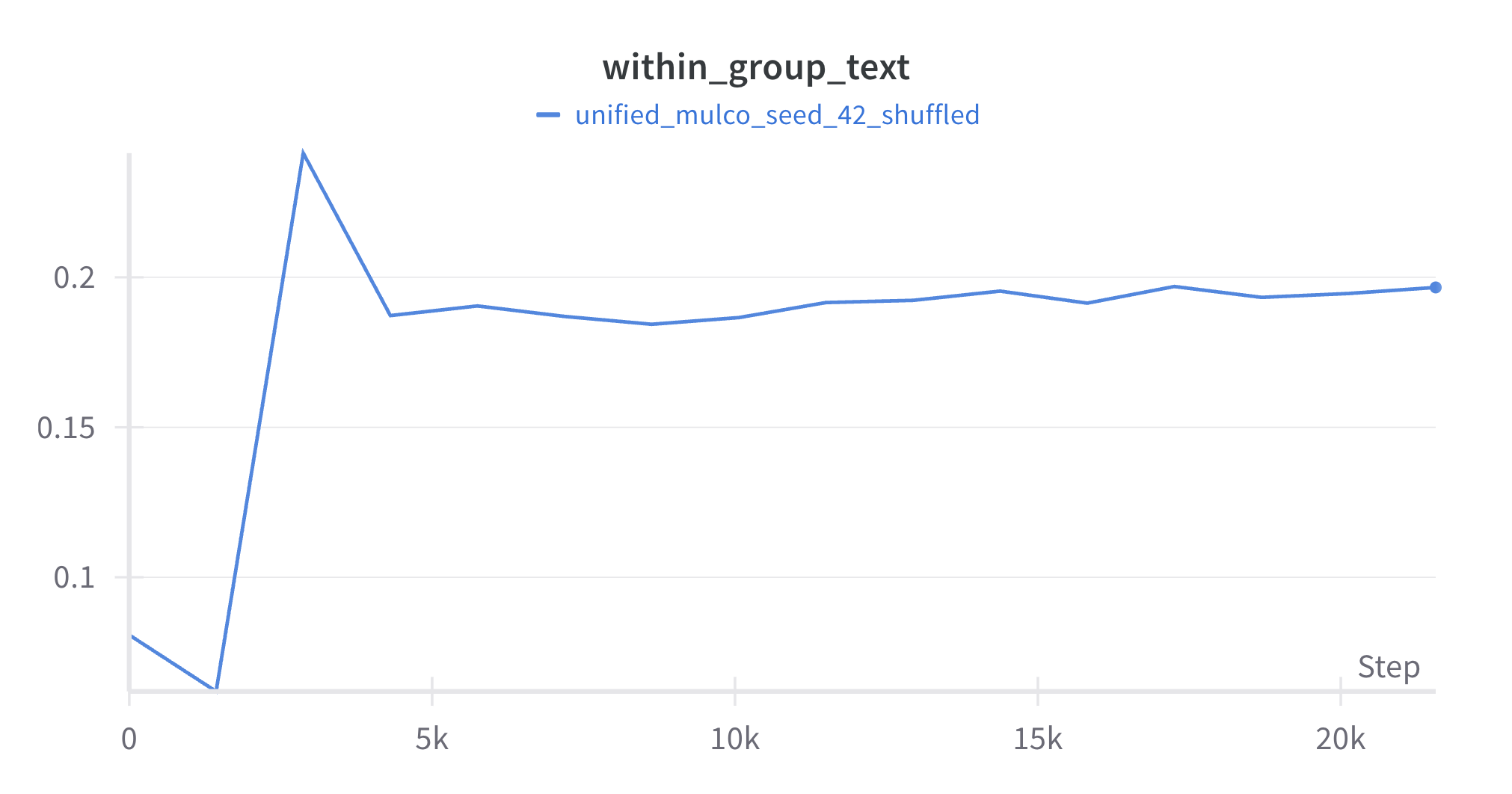}
\caption{Distance within text}
\end{subfigure}
% \vspace{1em}
\begin{subfigure}[t]{0.48\textwidth}
\centering
\includegraphics[trim=0 10 0 175, clip, width=\linewidth]{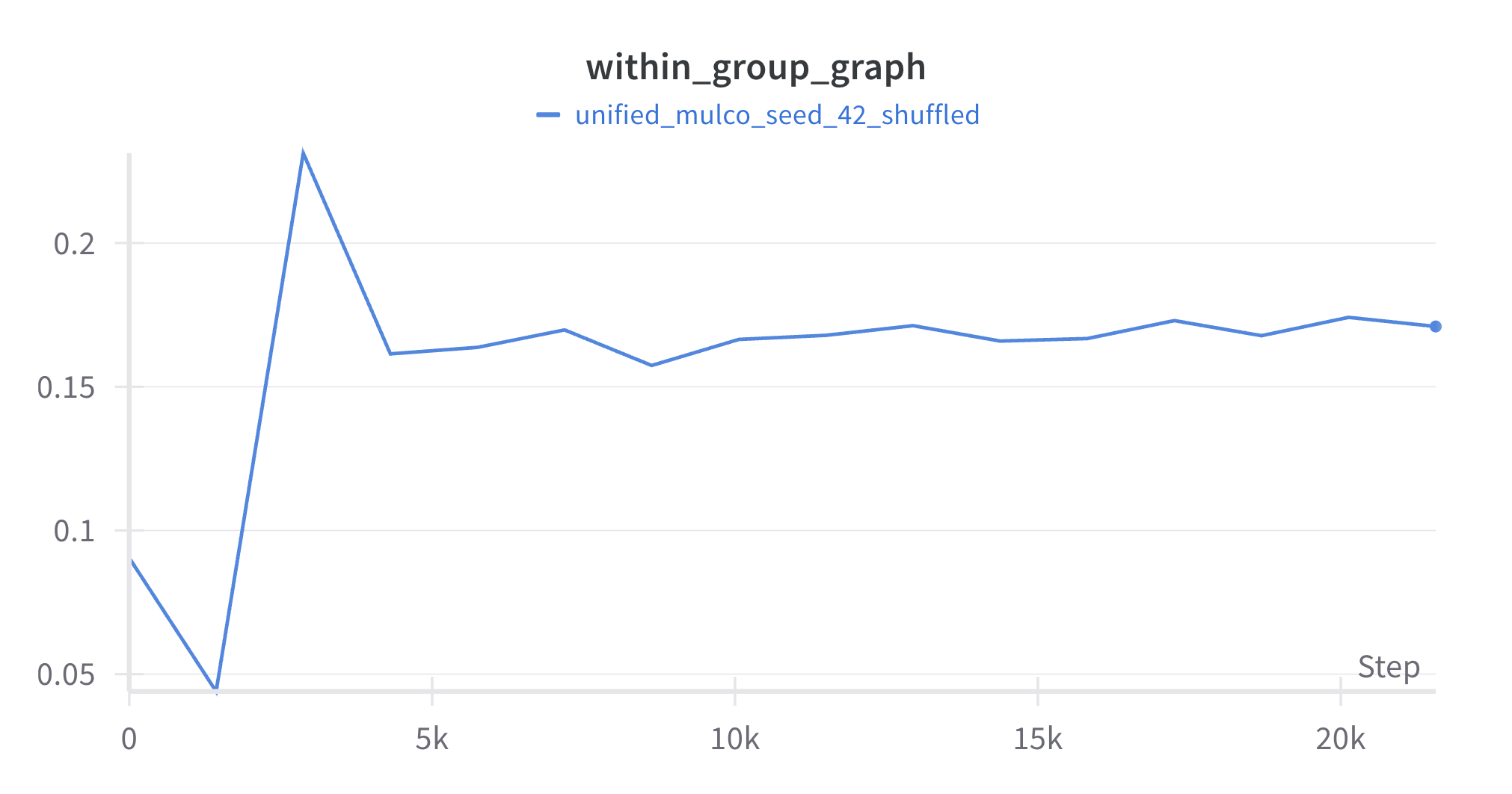}
\caption{Distance within graph}
\end{subfigure}
\hfill
\begin{subfigure}[t]{0.48\textwidth}
\centering
\includegraphics[trim=0 10 0 175, clip, width=\linewidth]{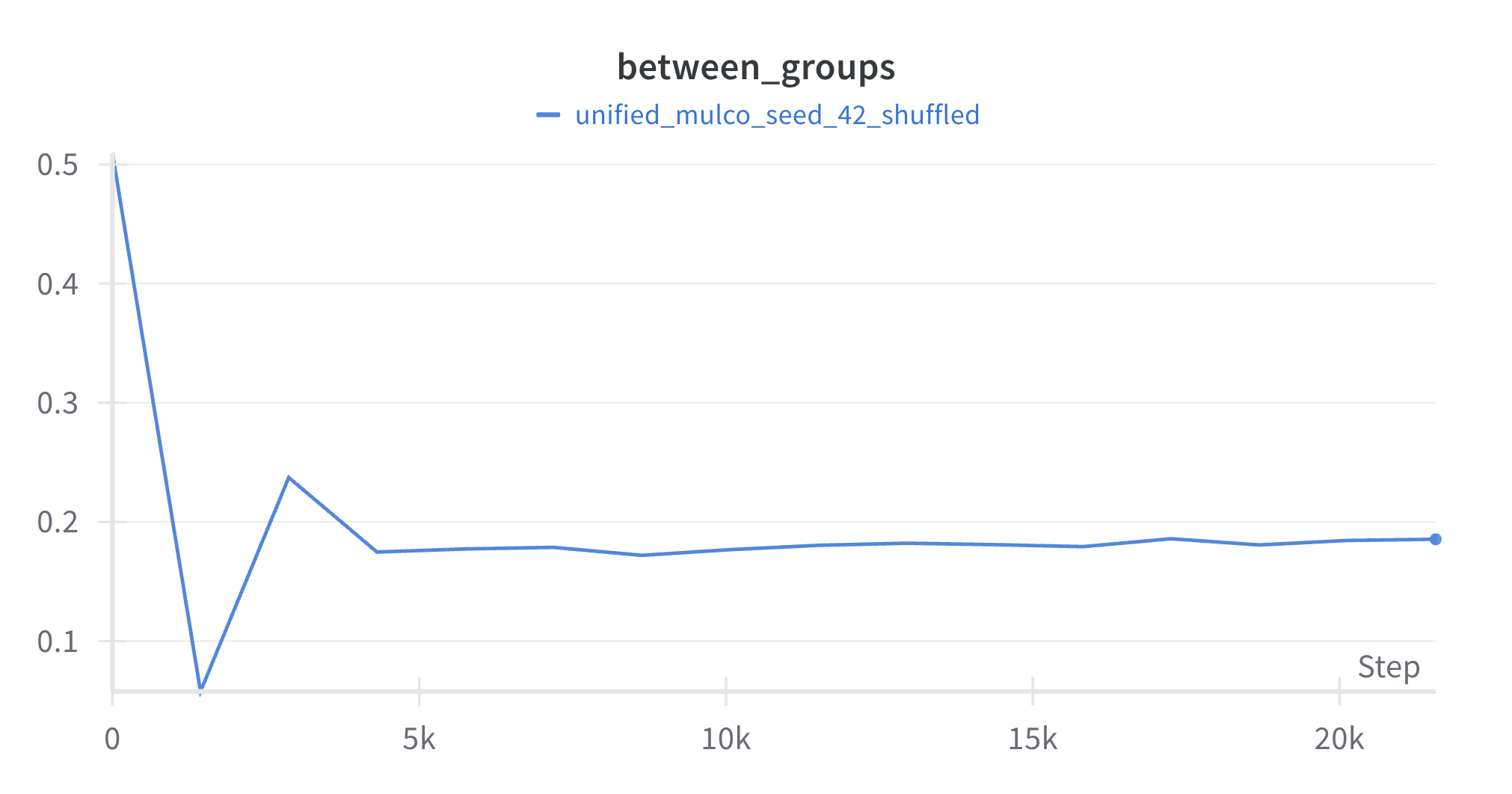}
\caption{Distance between text and graph}
\end{subfigure}
  % \vspace{\baselineskip}
  \caption{
    Results for KBQA entity-ranking on the WebQSP dataset.}
  \label{fig:ent_results}
\end{figure*}

\begin{figure*}[hbtp]
  \centering
  
  % Row 1: Three PCA plots
  \begin{subfigure}[t]{0.32\textwidth}
    \centering
    \includegraphics[trim=38 25 70 70, clip,width=\linewidth]{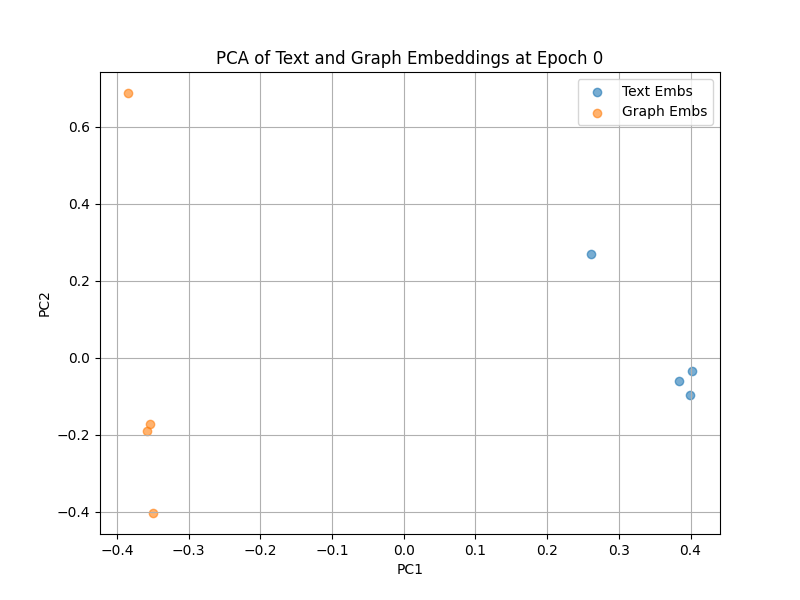}
    \caption{Initial epoch}
  \end{subfigure}
  \hfill
  \begin{subfigure}[t]{0.32\textwidth}
    \centering
    \includegraphics[trim=38 25 70 70, clip,width=\linewidth]{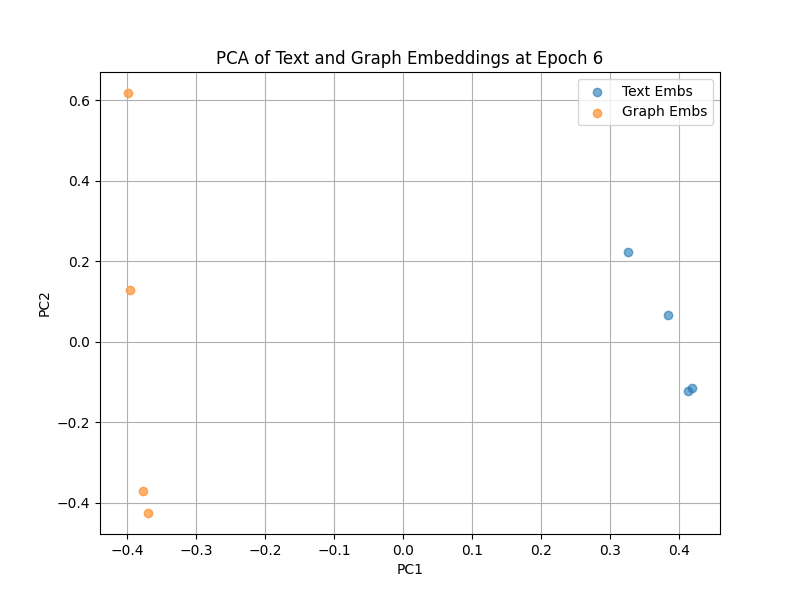}
    \caption{Intermediate epoch}
  \end{subfigure}
  \hfill
  \begin{subfigure}[t]{0.32\textwidth}
    \centering
    \includegraphics[trim=38 25 70 70, clip,width=\linewidth]{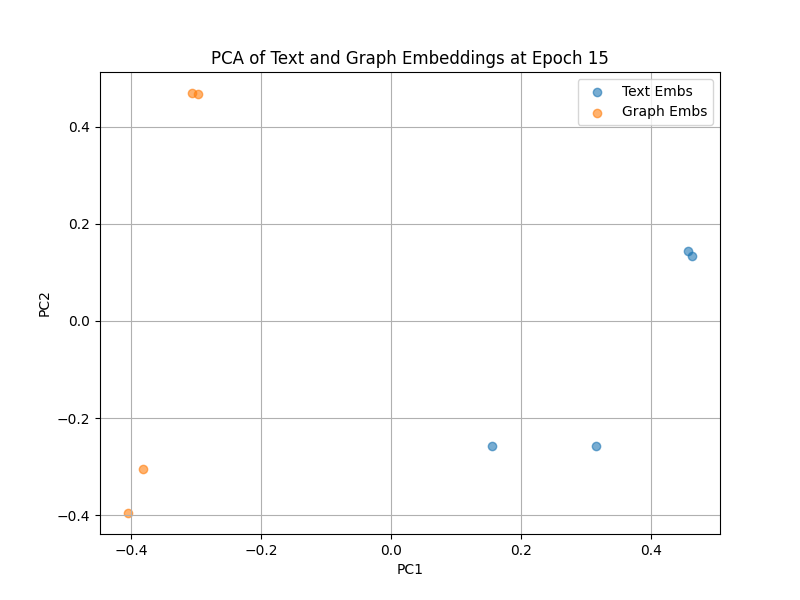}
    \caption{Final epoch}
  \end{subfigure}
  
  % \vspace{1em}
  
  % Row 2: Distance-related plots in 2x2 grid, trimmed more aggressively
\begin{subfigure}[t]{0.48\textwidth}
\centering
\includegraphics[trim=0 10 0 175, clip, width=\linewidth]{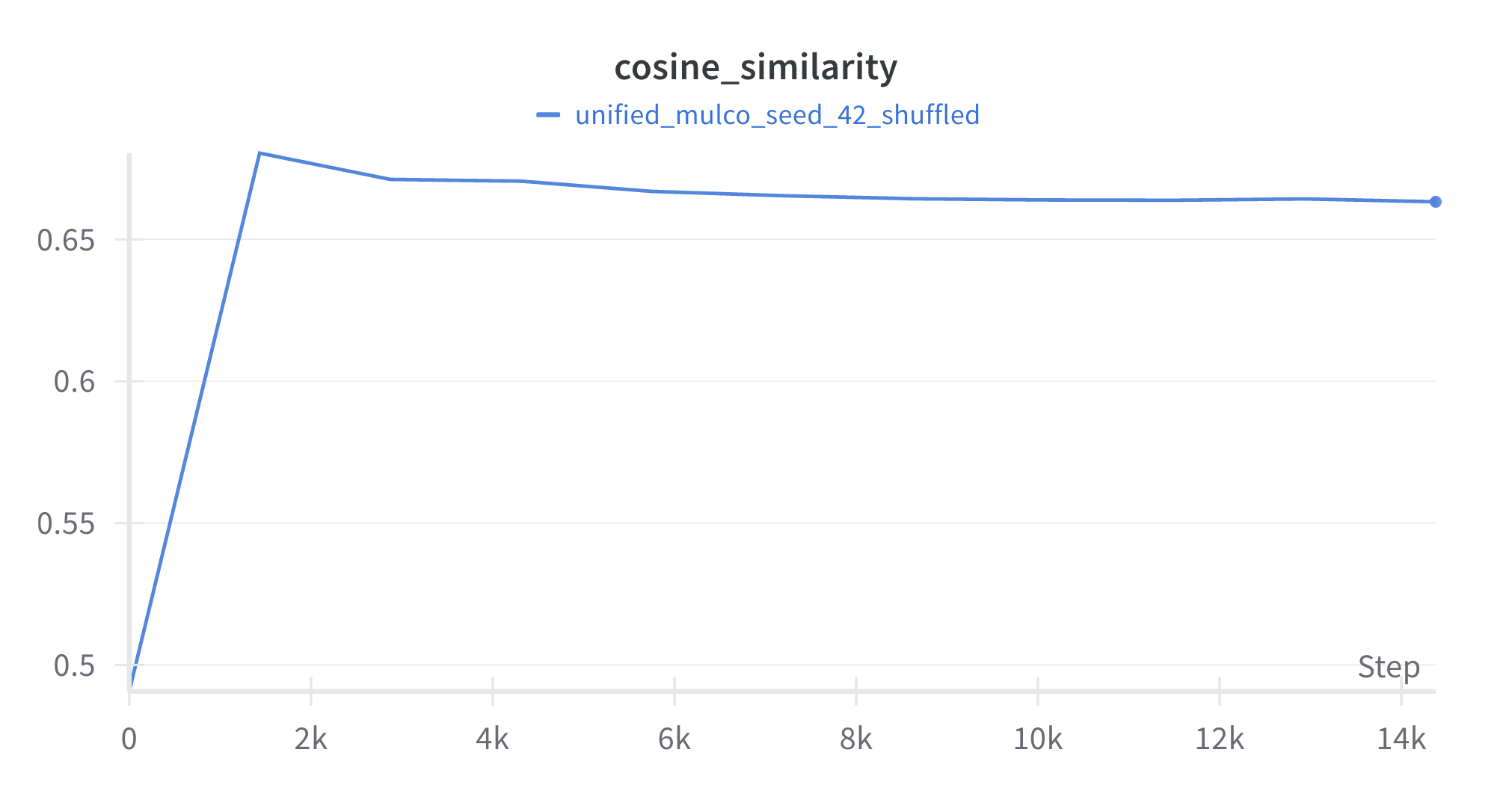}
\caption{Cosine similarity}
\end{subfigure}
\hfill
\begin{subfigure}[t]{0.48\textwidth}
\centering
\includegraphics[trim=0 10 0 175, clip, width=\linewidth]{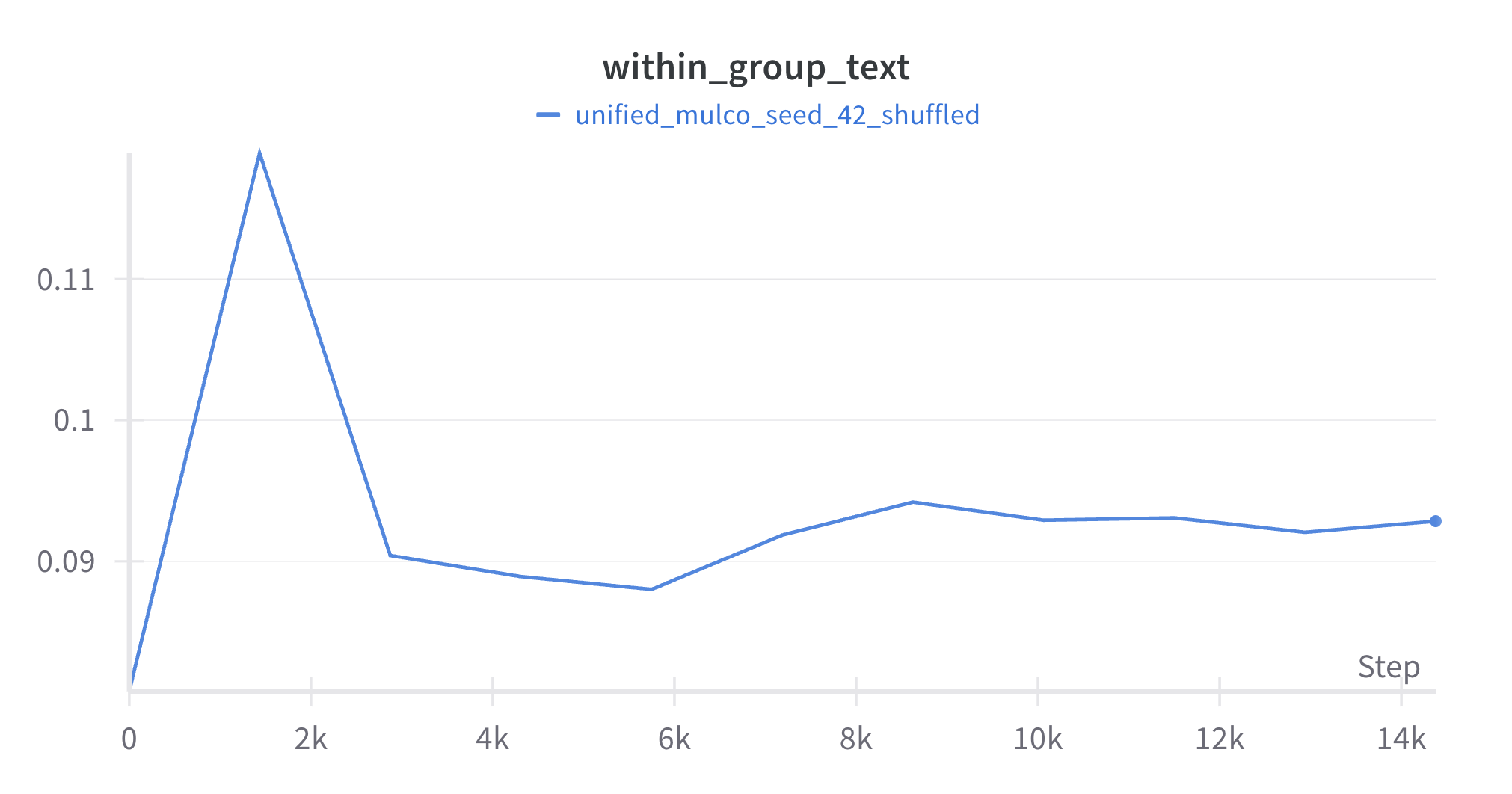}
\caption{Distance within text}
\end{subfigure}
% \vspace{1em}
\begin{subfigure}[t]{0.48\textwidth}
\centering
\includegraphics[trim=0 10 0 175, clip, width=\linewidth]{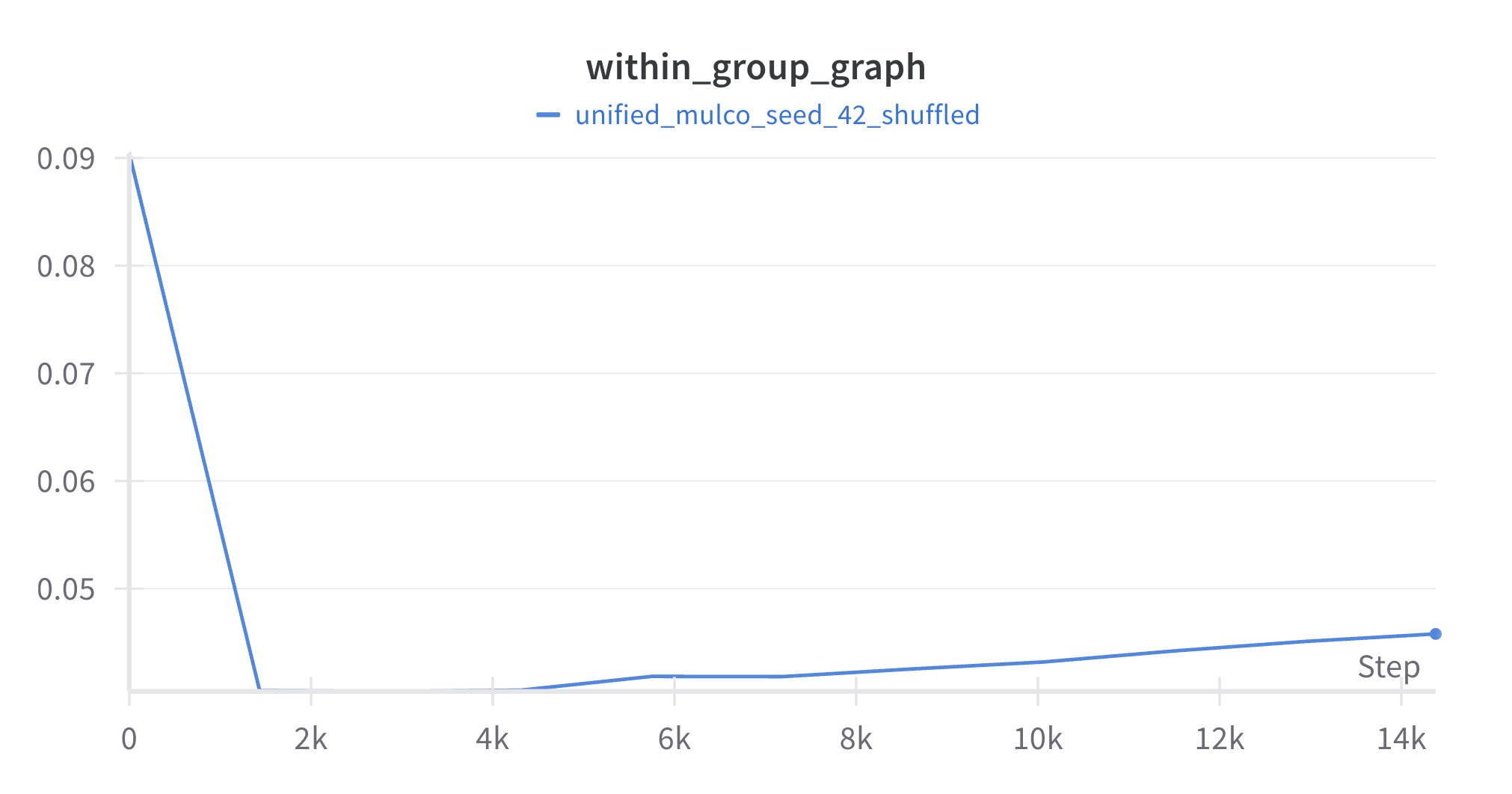}
\caption{Distance within graph}
\end{subfigure}
\hfill
\begin{subfigure}[t]{0.48\textwidth}
\centering
\includegraphics[trim=0 10 0 175, clip, width=\linewidth]{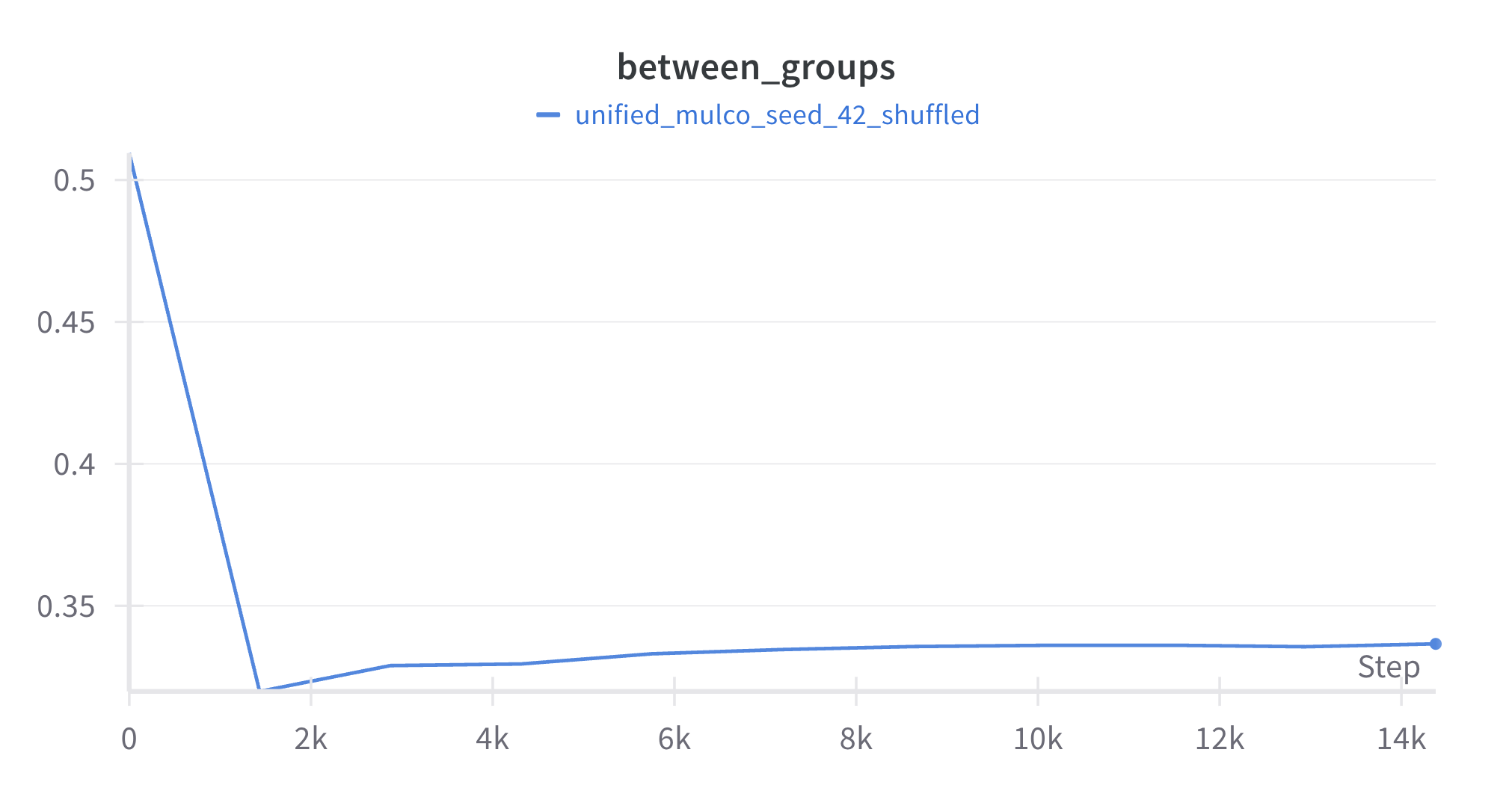}
\caption{Distance between text and graph}
\end{subfigure}
  % \vspace{\baselineskip}
  \caption{
    Results for KBQA entity-ranking on the WebQSP dataset when no CoD is applied.}
  \label{fig:ent_no-cod_results}
\end{figure*}

See Figure~\ref{fig:ent_results} and Figure~\ref{fig:ent_no-cod_results} for results for KBQA entity-ranking with and without CoD applied, respectively.

\end{document}